\documentclass[review]{elsarticle}

\usepackage{lineno,hyperref}
\usepackage{graphicx}
\usepackage{comment}
\usepackage{amssymb}
\usepackage{lineno}
\usepackage{lineno,hyperref}
\usepackage{amsmath,amsfonts,amsthm,amssymb,graphicx,subfigure,mathtools,tikz,hyperref,bm,blindtext,float}
\usepackage{diagbox}
\usepackage{multirow}
\usepackage{geometry}
\usepackage{algorithm}
\usepackage{algpseudocode}
\usepackage{amsmath,amsfonts,bm}

\def\eqref#1{equation~\ref{#1}}
\def\Eqref#1{Equation~\ref{#1}}
\def\1{\bm{1}}

\def\vx{{\bm{x}}}

\DeclareMathAlphabet{\mathsfit}{\encodingdefault}{\sfdefault}{m}{sl}
\SetMathAlphabet{\mathsfit}{bold}{\encodingdefault}{\sfdefault}{bx}{n}

\modulolinenumbers[5]

\journal{Journal of \LaTeX\ Templates}

\begin{document}

\begin{frontmatter}

\title{Learning Functional Priors and Posteriors from Data and Physics}
% \title{Meta-Learning and Uncertainty Quantification for Physical Problems with Data-Driven Physics-Informed Functional Priors}
% \title{Meta-Learning for PDE Problems with Uncertainty Quantification using Functional Priors Learned by Physics-Informed GANs}
% GANs as functional prior: Uncertainty quantification for meta learning, PINNs and DeepONets

%% Group authors per affiliation:
\author[brown]{Xuhui Meng \fnref{1}}
\author[brown]{Liu Yang \fnref{1}}
\author[XMU]{Zhiping Mao}
\author[mit]{Jos\'e del \'Aguila Ferrandis}
\author[brown,PNNL]{George Em Karniadakis \fnref{2}}
% \address{Brown University...}
% \fntext[myfootnote]{Since 1880.}

\address[brown]{Division of Applied Mathematics, Brown University, Providence, RI 02906, USA}
\address[XMU]{School of Mathematical Sciences, Xiamen University, Xiamen, 361005, China}
\address[mit]{Department of Mechanical Engineering,  Massachusetts Institute of                       Technology, Cambridge, MA 02139, USA}
\address[PNNL]{Pacific Northwest National Laboratory, Richland, WA 99354, USA}
\fntext[1]{The first two authors contributed equally to this work.}
\fntext[2]{Corresponding author: george\_karniadakis@brown.edu}
%% or include affiliations in footnotes:
% \author[mymainaddress,mysecondaryaddress]{Elsevier Inc}
% \ead[url]{www.elsevier.com}

% \author[mysecondaryaddress]{Global Customer Service\corref{mycorrespondingauthor}}
% \cortext[mycorrespondingauthor]{Corresponding author}
% \ead{support@elsevier.com}

% \address[mymainaddress]{1600 John F Kennedy Boulevard, Philadelphia}
% \address[mysecondaryaddress]{360 Park Avenue South, New York}

\begin{abstract}
We develop a new Bayesian framework based on deep neural networks to be able to extrapolate in space-time using historical data and to quantify uncertainties arising from both noisy and gappy data in physical problems. Specifically, the proposed approach has two stages: (1) prior learning and (2) posterior estimation. At the first stage, we employ the {\emph{physics-informed Generative Adversarial Networks}} (PI-GAN) to learn a functional prior either from a prescribed function distribution, e.g., Gaussian process, or from historical data and physics. At the second stage, we employ the Hamiltonian Monte Carlo (HMC) method  to estimate the posterior in the latent space of PI-GANs. In addition, we use two different approaches to encode the physics: (1) automatic differentiation, used in the physics-informed neural networks (PINNs) for scenarios with explicitly known partial differential equations (PDEs), and (2) operator regression using the deep operator network (DeepONet) for PDE-agnostic scenarios. We then test the proposed method for (1) meta-learning for one-dimensional regression, and forward/inverse PDE problems (combined with PINNs);  (2) PDE-agnostic physical problems (combined with DeepONet), e.g., fractional diffusion as well as saturated stochastic (100-dimensional)  flows in heterogeneous porous media; and (3) spatial-temporal regression problems, i.e., inference of a marine riser displacement field using experimental data from the Norwegian Deepwater Programme (NDP).  The results demonstrate that the proposed approach can provide accurate predictions as well as  uncertainty quantification given very limited scattered and noisy data, since historical data could be available to provide informative priors. In summary, the proposed method is capable of learning flexible functional priors, e.g.,  both Gaussian and non-Gaussian process, and can be readily extended to big data problems by enabling mini-batch training using stochastic HMC or normalizing flows since the latent space is generally characterized as low dimensional.

\end{abstract}


\end{frontmatter}

% \linenumbers

\section{Introduction}
Deep learning, capable of discovering complex representations from data, has drawn tremendous attention in diverse applications \cite{lecun2015deep,goodfellow2016deep}, such as image classification and generation. Generally, the data in real applications can be noisy and incomplete, resulting in uncertainties in model predictions. Understanding and quantifying uncertainty propagation in machine learning is thus crucial for making better decisions as well as avoiding possible disasters in critical situations  \cite{kendall2017uncertainties}. It can also help greatly in devising efficient active learning/adaptive sampling strategies. 

% As for the Bayesian inference, the former is in the function space, while the latter is in the parameter space, i.e., weight/bias.
Gaussian processes regression (GPR) \cite{rasmussen2003gaussian} and Bayesian neural networks (BNNs) \cite{neal2012bayesian} are two Bayesian-based probabilistic machine learning approaches that are widely used to quantify uncertainties in predictive science. These two approaches have also been extended to scientific machine learning, which is our particular interest in this study, e.g., solving forward and inverse PDE problems \cite{raissi2017machine,yang2021b}. We refer to the approaches in \cite{raissi2017machine} and \cite{yang2021b} as physics-informed GP (PI-GP) and Bayesian physics-informed neural networks (B-PINNs) in the present study, respectively.  

% characterize the functional output for a given prior, or 
It is challenging to apply PI-GP for nonlinear PDEs \cite{raissi2019physics} and problems with big data \cite{meng2020composite,meng2021multi}. B-PINNs are suitable for both linear and nonlinear PDEs, and capable of handling big data with specially designed inference methods (e.g., variational inference \cite{blundell2015weight}, stochastic Hamiltonian Monte Carlo \cite{chen2014stochastic}) by enabling mini-batch training.  In BNNs, a prior distribution needs to be prescribed on the hyperparameters, i.e., weights and biases. It is well known that BNNs with an infinite width are equivalent with a Gaussian process under certain assumptions \cite{neal2012bayesian,lee2017deep}. However, for BNNs with a finite width in real-world applications, it is still challenging to determine the prior for the parameters. Recent studies \cite{flam2017mapping,tran2020all} have proposed to tune the prior distribution in the parameter space to match a prescribed Gaussian process prior by minimizing the Kullback-Leibler (KL) divergence \cite{flam2017mapping} or the Wasserstein distance \cite{tran2020all} in the functional space. However, the results in \cite{tran2020all} showed that BNNs with some commonly used prior distributions such as Gaussian distribution, hierarchical prior, etc., are not quite expressive in functional space. To increase the flexibility of prior for BNNs in the functional space, normalizing flows (NFs) are then utilized to parameterize the prior for the hyperparameters \cite{tran2020all} to achieve better performance. Although great progress has been made in these two studies, we note that the dimensionality of the hyperparameters in the BNNs is generally high, which leads to expensive computational cost as well as difficulties in posterior estimation, and using NFs as priors will significantly increase the computational cost.

In other work, Yang {\sl et al.} proposed to learn stochastic processes from data using the {\emph{Generative Adversarial Networks}} (GANs) in \cite{yang2020physics}. Specifically, in \cite{yang2020physics} a deep neural network (DNN), which takes the spatial coordinate and Gaussian noise as input, is employed to approximate the target stochastic process. We note that a wide range of stochastic processes, including both Gaussian and non-Gaussian ones, can be successfully approximated due to the expressive power of DNNs. This inspired us to use similar techniques to learn functional priors from either a prescribed function distribution, e.g., Gaussian process, as in \cite{flam2017mapping,tran2020all} or historical data for Bayesian learning. In particular, such historical data can be collected from a series of experiments in labs or field experiments \cite{callaham2019robust}. The sensors in experiments may be sufficient and accurate to capture the typical features of the quantity of interest, but in real applications the number and fidelity of sensors would be limited, and thus we need to utilize these typical features so that they serve as our prior knowledge. For example, one can perform experiments in a towing tank with many sensors for the velocity, pressure, etc., in the flow field. From the collected data and physical laws, we will learn the prior distribution for these fields so that in the stage of posterior estimation fewer or cheaper sensors are required to infer the hydrodynamic information for marine systems in the ocean. Apart from a series of experiments, the historical data can also come from simulations with different parameters, boundary/initial conditions, etc. Often, due to the limitation in resolution, the historical data could be of low fidelity, e.g., simulations with coarse grids, measurements with large noise, etc. For such cases we can still learn informative priors with minor modifications to the learning method. Note that the historical data are widely used in scenarios of meta-learning applications \cite{finn2017model,hoffman2014no}. We present a brief introduction of meta-learning in  \ref{sec:maml}.

% Modern experimental methods and the increasing scale and resolution of numerical simulations have led to an abundance of fluid-flow data. Although we are able to achieve unprecedented fidelity in measurement and simulation in laboratory settings, in applications we are typically limited to a few noisy sensors. The challenge in flow-field estimation is thus to synthesize the profusion of offline data and limited, unreliable online information. This synthesis relies on learning and representing the essential structure of the flow field by leveraging the physical behavior observed in past data \cite{callaham2019robust}.

Leveraging the physical knowledge in the form of stochastic differential equations, the proposed physics-informed GANs (PI-GANs) in \cite{yang2020physics} can also integrate data from multiple deterministic or random fields, for example, source terms, boundary conditions, conductivity fields etc., that are intrinsically connected by the physics. This also matches our goal since in engineering problems both historical data (for prior) and new data (for posterior) could be collected from multiple random fields. The stochastic differential equation is assumed to be known in \cite{yang2020physics}, and thus the physical knowledge is encoded with automatic differentiation. In this paper, we make an extension of PI-GANs by introducing a PDE-agnostic approach to encode physical knowledge, i.e., applying DeepONet~\cite{lu2021learning} as an operator surrogate to correlate different physical terms.

\begin{figure}
    \centering
    \includegraphics[width=0.8\textwidth]{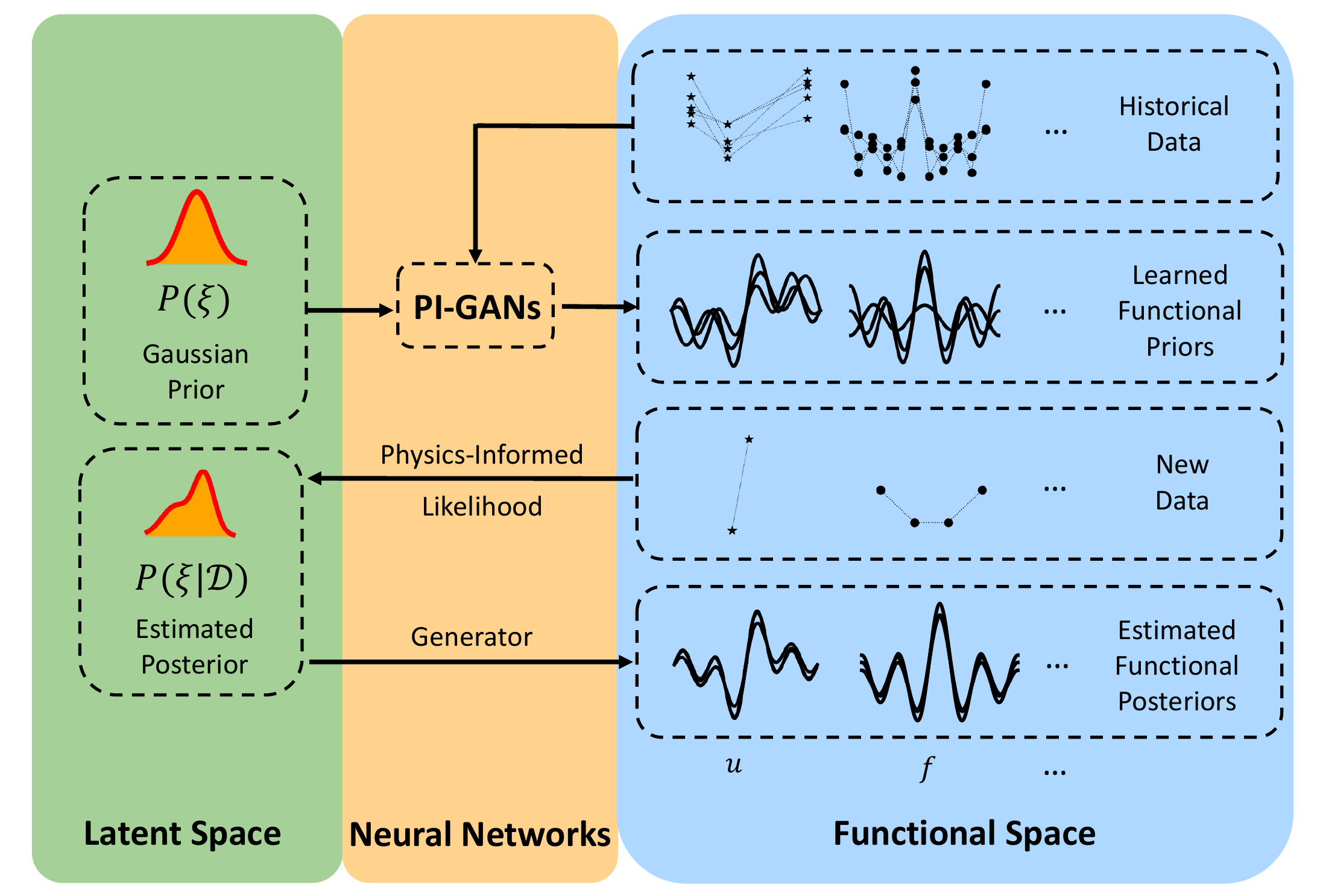}
    \caption{
    Schematic of learning functional priors and posteriors from data and physics. The ``PI'' part is based on either physics-informed neural networks with automatic differentiation or operator approximation in the form of DeepONet.
    % Schematic for Bayesian inference using PI-GANs as functional priors.  
    $P(\bm{\xi})$ is the prior distribution for input noise $\bm{\xi}$, i.e., $\mathcal{N}(0, \bm{I})$, 
    $P(\bm{\xi}|\mathcal{D})$ represents the posterior distribution of $\bm{\xi}$, where $\mathcal{D}$ is new data.
    % $P(D^{\mbox{target}}|\bm{\xi})$ is the likelihood for observations, which is usually a Gaussian distribution with zero mean and known variance, and $P(\bm{\xi}|D^{\mbox{target}})$ represents the posterior distribution. 
    % $D^{\mbox{source}}$: meta-training data in meta-learning or pre-training data in DeepONets, $D^{\mbox{target}}$: training data (blue circles) for an unseen target task; HMC: Hamiltonian Monte Carlo; DFM: deep flow-based model. 
    }
    \label{fig:schem}
\end{figure}

We illustrate the workflow in Fig.~\ref{fig:schem}. We first use PI-GANs to learn functional priors from historical data and physics. The generator in PI-GANs, which takes the coordinate $\bm{x}$ and Gaussian noise $\bm{\xi}$ as input to represent a stochastic process, will approximate the functional prior. Then, conditioned on the new data  $\mathcal{D}$, we can sample from the posterior of $\bm{\xi}$ with physics-informed likelihood using Hamiltonian Monte Carlo (HMC) or other techniques. In the end, with the generator and samples of $P(\bm{\xi}|\mathcal{D})$, we can have samples of the functional posterior.

Independently, \cite{patel2020gan,patelbayesian} also used GANs to learn the prior of image-type vectors for physical problems. However, instead of learning a \textit{vector} prior merely using the data for the quantity of interest as in \cite{patel2020gan,patelbayesian}, which are not always available in practice, here we apply physics-informed GANs to learn the \textit{functional} prior using data collected for various state variables. Moreover, as a benefit of functional prior, instead of using images as data, here we can tackle the case where the data are collected from scattered sensors, for which we need to consider not only the uncertainty from noise (i.e., \emph{aleatoric uncertainty}) but also the uncertainty from the possible gaps between sensors (i.e., \emph{epistemic uncertainty}).

 The novelty of the proposed method is three-fold: 
 \begin{itemize}
 \item In certain scenarios with historical data, the functional priors learned by the GANs/PI-GANs are superior to the artificially designed ones (e.g., Gaussian processes), in that they are more flexible since they can approximate a wide range of stochastic processes with neural networks, and also more reasonable since they reflect our knowledge from historical data. 
 \item Compared with BNNs, the Bayesian inference is performed in the latent space (GANs) instead of the parameter space (BNNs), which generally has much lower dimensions and is easier for posterior sampling.
 \item With physics encoded in both stages of prior learning and posterior estimation, either in a PDE-based or PDE-agnostic way, the method can integrate data from multiple fields.
 \end{itemize} 

The rest of the paper is organized as follows. In Sec.~\ref{sec:method}, we introduce the problem setup and the methodology. In Sec.~\ref{sec:results}, we show our computational tests, including 1D and 2D, regression and PDE problems. We summarize in Sec.\ref{sec:summary}. In \ref{sec:maml}, we introduce the model-agnostic meta-learning (MAML) method, which is compared against the proposed method in Sec.~\ref{sec:results}. A comparison between different generator architectures is presented in \ref{sec:comparison}. We include some discussion on why the Wasserstein GANs could fail, which suggests a possible direction of future research. The details of  learning the hyperparameters are in \ref{sec:dnn_arch}.

% We summarize the the novelty of the present method in the following three-fold: (i) the functional prior learned by the GANs/PI-GANs can be but not limited to a Gaussian process, suggesting that it is more flexible than the Gaussian process. Specifically, the functional priors learned by PI-GANs are superior to the artificially designed ones (e.g. Gaussian processes) in certain scenarios, e.g., meta-learning, since it reflects our knowledge from historical data; (ii) compared with BNNs the Bayesian inference is performed in the latent space, which generally has much lower dimensions than the BNNs; and (iii) the Bayesian inference is scalable with respect to big training dataset by employing the stochastic MCMC/HMC or deep flow-based generative models. 

\section{Methodology}
\label{sec:method}

As already mentioned in the Introduction, we assume that we have two sets of data: (a) historical data for learning the functional priors, and (b) new data for the posterior estimation. The historical data are denoted as $\mathcal{\overline{D}}$ and the new data are denoted as $\mathcal{D}$. The particular focus of the current study is on regression and PDE problems. For the latter case, both $\mathcal{\overline{D}}$ and $\mathcal{D}$ could consist of measurements for different state variables that are intrinsically connected by the physics.

Let us consider a partial differential equation (PDE) equation of the form 
\begin{equation}\label{eqn:Nequ}
\begin{aligned}
        \mathcal{N}_{\vx}u(\vx)&=f(\vx), \vx \in \Omega\\
        \mathcal{B}_{\vx}u(\vx)&=b(\vx), \vx \in \partial\Omega
\end{aligned}
\end{equation}
as an example, where $\mathcal{N}_{\vx}$ is a general differential operator, $\mathcal{B}_{\vx}$ is the boundary operator, $\Omega$ is the domain, $\partial\Omega$ is the boundary.

Suppose we place $\overline{N}_u$ $u$-sensors at $\{\bar{\vx}_{u}^{(i)}\}_{i=1}^{\overline{N}_u}$, $\overline{N}_f$ $f$-sensors at $\{\bar{\vx}_{f}^{(i)}\}_{i=1}^{\overline{N}_f}$, and $\overline{N}_b$ $b$-sensors at $\{\bar{\vx}_{b}^{(i)}\}_{i=1}^{\overline{N}_b}$. Multiple reads of these sensors would yield different vectors, i.e., the snapshots denoted as  $\mathcal{\overline{D}} = \{\overline{T}_j\}_{j=1}^M, \overline{T}_j \in \mathbb{R}^{\overline{N}_u+\overline{N}_f+\overline{N}_b}, \forall j$. These data are our historical data. Specifically, 
\begin{equation}\label{eqn:Tsnap}
\begin{aligned}
    \overline{T}_j &= (\overline{U}_j, \overline{F}_j, \overline{B}_j),\\
    \overline{U}_j &= (u_j(\bar{\vx}_{u}^{(i)}))_{i=1}^{\overline{N}_u},\\
    \overline{F}_j &= (f_j(\bar{\vx}_{f}^{(i)}))_{i=1}^{\overline{N}_f},\\
    \overline{B}_j &= (b_j(\bar{\vx}_{b}^{(i)}))_{i=1}^{\overline{N}_b},\\
\end{aligned}
\end{equation}
where $u_j, f_j$, and $b_j$ are different realizations of $u, f$ and $b$ that should satisfy Equation~\ref{eqn:Nequ}. For simplicity, here we assume that the sensors are of high-fidelity, and the observations can be viewed as noiseless. We will discuss the noisy cases later.

\begin{figure}
    \centering
    \includegraphics[width=0.9\textwidth]{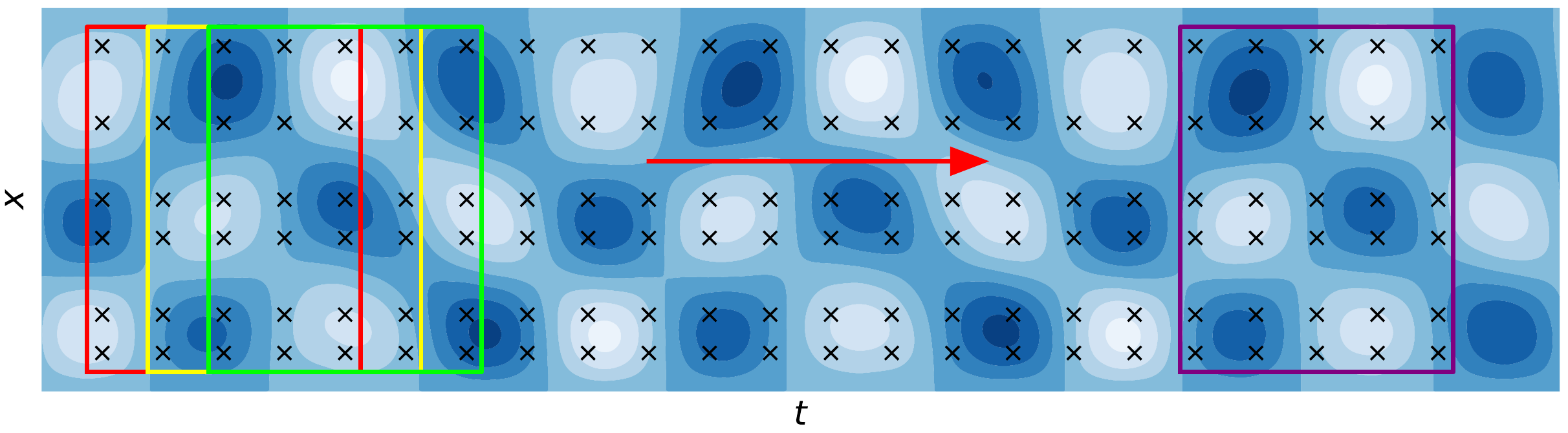}
    \caption{Schematic plot of using the sliding windows to generate fake snapshots in the spatio-temporal domain. The black crosses show the periodic reads of six sensors in the spatial domain. The red, yellow, green, and purple rectangles represent four windows; each window will generate a snapshot with 30 observations in the spatio-temporal domain.
    }
    \label{fig:sliding}
\end{figure}

We now present some examples for the ``historical data''. In the first example, the regression or the PDE problem is time-independent, i.e., $\Omega$ is only the spatial domain. This could happen in the cases when some terms in PDE are case dependent, e.g., flow resistance for different soils or rocks. In the second example, the regression or the PDE problem is time-dependent, i.e., $\Omega$ is the spatio-temporal domain. Typically, for such cases we can have periodic reads of some sensors scattered in the spatial domain, and thus we can use sliding windows to generate fake snapshots in the spatial-temporal domain. We illustrate this point with a schematic plot in Fig.~\ref{fig:sliding}.

In the posterior estimation stage we have the new data $\mathcal{D} = \mathcal{D}_u \cup \mathcal{D}_f \cup \mathcal{D}_b$, and
\begin{equation}
\label{eqn:testdata}
\begin{aligned}
    \mathcal{D}_u = \{(\vx_u^{(i)}, u^{(i)})\}_{i=1}^{N_u}, \\
    \mathcal{D}_b = \{(\vx_b^{(i)}, b^{(i)})\}_{i=1}^{N_b}, \\
    \mathcal{D}_f = \{(\vx_f^{(i)}, f^{(i)})\}_{i=1}^{N_f}, \\
\end{aligned}
\end{equation}
where $u^{(i)}$ is the noisy measurements of $u$ at $\vx_u^{(i)}$, i.e., $u^{(i)} = u(\vx_u^{(i)}) + \epsilon_{u}^{(i)}, \epsilon_{u}^{(i)} \sim \mathcal{N}(0, {\sigma_u^{(i)}}^2)$, and similarly for the other terms. It is important to note that the sensor locations for $\mathcal{\overline{D}}$ and $\mathcal{D}$ are different.

In the following, we will introduce how to learn the functional priors from data $\mathcal{\overline{D}}$ and physics, and how to make predictions with uncertainty quantification, i.e., estimate the posterior conditioned on the data $\mathcal{D}$.

% \subsection{Functional priors in meta-learning and DeepONets}

% In this section, we will first briefly review the meta-learning and DeepONets, and then we will demonstrate how to employ the GANs/PI-GANs to learn the functional priors in these methods. Finally, we present the Bayesian inference which is used to quantify the uncertainties in predictions using the functional priors learned by GANs/PI-GANs. 

\subsection{Data-driven physics-informed functional priors}

\subsubsection{Physics-informed Generative Adversarial Networks}

In general, generative adversarial networks (GANs) \cite{goodfellow2014generative} aim to use a generator neural network $G_{\bm{\eta}}$ with parameters $\bm{\eta}$ to approximate the data distribution $P_r$. Here, the data are vectors in a Euclidean space. The generator $G_{\bm{\eta}}$ takes random noise $\bm{\xi}\sim P(\bm{\xi})$ as input and outputs generated samples $G_{\bm{\eta}}(\bm{\xi})$ whose underlying distribution is $P_g$. GANs solve this problem by introducing another discriminator neural network $D_{\bm{\rho}}$, which takes a sample as input and outputs a real value indicating whether this sample is generated by $G_{\bm{\eta}}$ or real sample from $P_r$. The generator $G_{\bm{\eta}}$ and discriminator $D_{\bm{\rho}}$ are trained in an adversarial way. The hope is that the generated distribution $P_g$ will converge to the target one $P_r$, so that the discriminator cannot distinguish the generated samples and real ones. There are different versions of GANs; in Wasserstein GANs with gradient penalty (WGAN-GP) \cite{gulrajani2017improved} the formal loss functions for the generator and discriminator are 
\begin{equation} \label{eqn:ClassicGDloss}
\begin{aligned}
L_G &= -\mathbb{E}_{\bm{\xi}\sim P(\bm{\xi})} [ D_{\bm{\rho}}(G_{\bm{\eta}}(\bm{\xi}))], \\
L_D &= \mathbb{E}_{\bm{\xi}\sim P(\bm{\xi})} [ D_{\bm{\rho}}(G_{\bm{\eta}}(\bm{\xi}))] - \mathbb{E}_{T\sim P_r} [ D_{\bm{\rho}}(T)] + \lambda \mathbb{E}_{\hat{T} \sim P_i} (\Vert \nabla_{\hat{T}} D_{\bm{\rho}}(\hat{T}) \Vert_2 -1 )^2,
\end{aligned}
\end{equation}
where $P_i$ is the distribution induced by uniform sampling on interpolation lines between independent samples of $T$ and $G_{\bm{\eta}}(\bm{\xi})$, and $\lambda$ is the gradient penalty coefficient. Note that the loss function for the generator can be interpreted as the Wasserstein-1 distance between the generated distribution and the target data distribution, up to constants.

In this paper, we are not aiming to learn a distribution of vectors in a Euclidean space, but a distribution of functions that serves as the functional prior. The physics-informed GAN (PI-GAN) developed in \cite{yang2020physics} perfectly matches our goal. PI-GANs cannot only learn the distribution of functions from repeated reads of scattered sensors, but also incorporate the physics into the learning system so that the data collected from multiple terms can be leveraged when learning the functional distribution for the quantity of interest. In \cite{yang2020physics}, PI-GANs encode the physics in the form of PDEs with automatic differentiation as in physics-informed neural networks (PINNs) \cite{raissi2019physics}. In this paper, apart from PINNs, we also make a further step to utilize DeepONets as PDE-agnostic operator surrogate to incorporate the physics \cite{lu2021learning}. We introduce the details as follows.

Consider Equation~\ref{eqn:Nequ}; we will use generators taking the form of $\tilde{u}_{\bm{\eta}}(\vx;\bm{\xi})$, $\tilde{f}_{\bm{\eta}}(\vx;\bm{\xi})$, and $\tilde{b}_{\bm{\eta}}(\vx;\bm{\xi})$ to represent the functional prior for $u$, $f$ and $b$, which are functions of the coordinate $\vx$ and noise $\bm{\xi}\sim P(\bm{\xi})$. Here, $P(\bm{\xi})$ is the input noise distribution, and in this paper we set $P(\bm{\xi})$ as standard multivariate Gaussian, where $\bm{\eta}$ is the parameter. Note that $\tilde{u}_{\bm{\eta}}$, $\tilde{f}_{\bm{\eta}}$ and $\tilde{b}_{\bm{\eta}}$ are not independent and share parameters. We will discuss these generators' architectures and trainable parameters later.

On the one hand, we have real snapshots $\mathcal{\overline{D}} = \{\overline{T}_j\}_{j=1}^M$ defined in Equation~\ref{eqn:Tsnap}, which can be viewed as $({\overline{N}_u+\overline{N}_f+\overline{N}_b})$-dimensional samples drawn from a hidden distribution $P_r$. On the other hand, with the generators $\tilde{u}_{\bm{\eta}}(\vx;\bm{\xi})$, $\tilde{f}_{\bm{\eta}}(\vx;\bm{\xi})$, and $\tilde{b}_{\bm{\eta}}(\vx;\bm{\xi})$, we can generate ``fake snapshots'' $Q_{\bm{\eta}}(\boldsymbol{\xi})\in \mathbb{R}^{\overline{N}_u+\overline{N}_f+\overline{N}_b}$:
\begin{equation} \label{eqn:Gsnap}
\begin{aligned}
Q_{\bm{\eta}}(\boldsymbol{\xi}) & = ( \tilde{U}_{\bm{\eta}}(\boldsymbol{\xi}),\tilde{F}_{\bm{\eta}}(\boldsymbol{\xi}),\tilde{B}_{\bm{\eta}}(\boldsymbol{\xi})),\\
\tilde{U}_{\bm{\eta}}(\boldsymbol{\xi}) & = (\tilde{u}_{\bm{\eta}}(\bar{\vx}_{u}^{(i)}; \bm{\xi}))_{i=1}^{\overline{N}_{u}},\\
\tilde{F}_{\bm{\eta}}(\boldsymbol{\xi}) & = (\tilde{f}_{\bm{\eta}}(\bar{\vx}_{f}^{(i)}; \bm{\xi}))_{i=1}^{\overline{N}_{f}},\\
\tilde{B}_{\bm{\eta}}(\boldsymbol{\xi}) & = (\tilde{b}_{\bm{\eta}}(\bar{\vx}_{b}^{(i)}; \bm{\xi}))_{i=1}^{\overline{N}_{b}}.\\
\end{aligned}
\end{equation}

The discriminator neural network $D_{\bm{\rho}}$ parameterized by $\bm{\rho}$ takes a real or fake snapshot as input, and outputs a real number. If we use WGAN-GP, then the loss function for the generator parameters $\bm{\eta}$ and the discriminator parameters $\bm{\rho}$ are
\begin{equation} \label{eqn:GDloss}
\begin{aligned}
L_G &= -\mathbb{E}_{\bm{\xi}\sim P(\bm{\xi})} [ D_{\bm{\rho}}(Q_{\bm{\eta}}(\boldsymbol{\xi}))], \\
L_D &= \mathbb{E}_{\bm{\xi}\sim P(\bm{\xi})} [ D_{\bm{\rho}}(Q_{\bm{\eta}}(\boldsymbol{\xi}))] - \mathbb{E}_{T\sim P_r} [ D_{\bm{\rho}}(T)] + \lambda \mathbb{E}_{\hat{T} \sim P_i} (\Vert \nabla_{\hat{T}} D_{\bm{\rho}}(\hat{T}) \Vert_2 -1 )^2.
\end{aligned}
\end{equation}
Here, we set the penalty weight as $\lambda=0.1$. During the training, we update $\bm{\rho}$ and $\bm{\eta}$ iteratively with the ratio of $5:1$. Compared with the GAN losses in Equation~\ref{eqn:ClassicGDloss}, the only difference is to replace $G_{\bm{\eta}}(\bm{\xi})$, which is the direct output of the generator, with $Q_{\bm{\eta}}(\bm{\xi})$, which is the generated snapshots induced from the stochastic function formulated in Equation \ref{eqn:Gsnap}.

In Equation~\ref{eqn:Tsnap} we assume that the historical data are noiseless, so are the generated snapshots in Equation~\ref{eqn:Gsnap}. If the historical data are noisy, i.e., the entries of data are perturbed with independent random noise, then the entries of generated snapshots should also be augmented with noise of the same distribution, but the loss functions in Equation~\ref{eqn:GDloss} do not need to be changed.

\subsubsection{Encoding PDEs with physics-informed neural networks (PINNs)}
The first approach to incorporate physics is to use automatic differentiation to encode PDEs as in PINNs \cite{raissi2019physics}. In particular, if we know the PDE, then we can use a generative model $G_{\bm{\eta}}(\vx; \bm{\xi})$ as $\tilde{u}_{\bm{\eta}}(\vx; \bm{\xi})$, and hence the generators for $u(\vx)$, $f(\vx)$ boundary $b(\vx)$ can be written as:
\begin{equation}\label{eqn:generator_pinn}
\begin{aligned}
    \tilde{u}_{\bm{\eta}}(\vx; \bm{\xi}) &= G_{\bm{\eta}}(\vx; \bm{\xi}),\\
    \tilde{f}_{\bm{\eta}}(\vx; \bm{\xi}) &= \mathcal{N}_{\vx}G_{\bm{\eta}}(\vx; \bm{\xi}),\\
    \tilde{b}_{\bm{\eta}}(\vx; \bm{\xi}) &= \mathcal{B}_{\vx}G_{\bm{\eta}}(\vx; \bm{\xi}),\\
\end{aligned}
\end{equation}
respectively.

\subsubsection{DeepONets as PDE-agnostic operator surrogates}
\begin{figure}[H]
    \centering
    \includegraphics[width=0.5\textwidth]{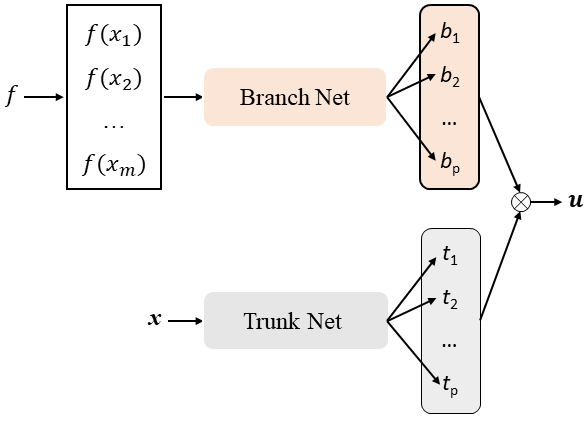}
    \caption{
    Schematic of DeepONet. $f$ is the input of  Branch Net,  $x_1, x_2, .., x_m$ are the discrete points to resolve the input function, $b_1, b_2, ..., b_p$ and $t_1, t_2, ..., t_p$ are the outputs for the Branch Net and Trunk Net, respectively, $\bm{u}$ is the output target function. If the operator takes multiple functions as input (e.g., source term $f$, boundary condition $b$, etc.), then we just need to concatenate the multiple vectors that resolve these functions as the input of the Branch Net.
    }
    \label{fig:deeponet}
\end{figure}

Apart from the PDE-based approach, as an operator surrogate, the recently developed DeepONet which has been justified theoretically to be a universal approximator to any continuous nonlinear operator \cite{lu2021learning} can also incorporate physics into the prior learning by ``bridging", i.e., inter-relating the different physical terms. As shown in Fig. \ref{fig:deeponet}, the DeepONet is composed of two sub-networks, i.e., the Branch Net (BN) and the Trunk Net (TN). The input for BN is a function, which is represented by a set of discrete function values at certain locations, i.e., $x_1, x_2, ..., x_m$, and the output of BN is a vector $[b_1, b_2, ..., b_p]$. In addition,  TN takes $\bm{x}$ as input and outputs a vector $[t_1, t_2, ..., t_p]$. The output of the DeepONet is the inner product of these two vectors as $\bm{u} = \sum^p_{j=1} b_j t_j$.

%  Note that  the number used to resolve different functions should be the same, i.e., $m$ is fixed for all input functions. 
 In the present study, we employ uniform grids to discretize the input functions in both one- and two-dimensional problems. In addition, the TN has no constraint on the input $\bm{x}$, suggesting that we can evaluate the output $\bm{u}$ at any location. Both the BN and TN employed in this study are multilayer perceptrons (MLP), which are trained simultaneously by minimizing  the mean squared error (MSE) between the given and predicted $\bm{u}$ from DNNs using the Adam optimizer. More details on the DeepONet can be found in \cite{lu2021learning,cai2021deepm}.

In particular, if we have data of $u, b$, and $f$, we can train a DeepONet as a solution operator, denoted as $\tilde{S}$, that maps the function $b$ and $f$ to the function $u(\vx) = \tilde{S}[f,b](\vx)$. 
We can then use two independent deep generative models $G^{(f)}_{\bm{\eta}_1}(\vx; \bm{\xi})$ and $G^{(b)}_{\bm{\eta}_2}(\vx; \bm{\xi})$ as $\tilde{f}_{\bm{\eta}_1}(\vx; \bm{\xi})$ and $\tilde{b}_{\bm{\eta}_2}(\vx; \bm{\xi})$, where the subscripts denote the parameters; the generators for $u(\vx)$, $f(\vx)$, and $b(\vx)$ are thus expressed as
\begin{equation}\label{eqn:generator_deepOnet1}
\begin{aligned}
    \tilde{u}_{\bm{\eta}_1, \bm{\eta}_2}(\vx; \bm{\xi}) &= \tilde{S}[G^{(f)}_{\bm{\eta}_1}(\cdot; \bm{\xi}), G^{(b)}_{\bm{\eta}_2}(\cdot; \bm{\xi})](\vx),\\
    \tilde{f}_{\bm{\eta}_1}(\vx; \bm{\xi}) &= G^{(f)}_{\bm{\eta}_1}(\vx; \bm{\xi}),\\
    \tilde{b}_{\bm{\eta}_2}(\vx; \bm{\xi}) &= G^{(b)}_{\bm{\eta}_2}(\vx; \bm{\xi}),\\
\end{aligned}
\end{equation}
respectively. Note that as inputs to $\tilde{S}$, $G^{(f)}_{\bm{\eta}_1}(\cdot; \bm{\xi})$ and $G^{(b)}_{\bm{\eta}_2}(\cdot; \bm{\xi})$ are two functions of $\vx$ with fixed $\bm{\xi}$.

If the boundary condition is fixed, then we only need data of $u$ and $f$ and train a DeepONet that maps $f$ to $u = \tilde{S}[f]$. Consequently, we only need one independent generative model $G_{\bm{\eta}}(\vx; \bm{\xi})$, and the generators for $u(\vx)$ and $f(\vx)$ are
\begin{equation}\label{eqn:generator_deepOnet1}
\begin{aligned}
    \tilde{u}_{\bm{\eta}}(\vx; \bm{\xi}) &= \tilde{S}[G_{\bm{\eta}}(\cdot; \bm{\xi})](\vx),\\
    \tilde{f}_{\bm{\eta}}(\vx; \bm{\xi}) &= G_{\bm{\eta}}(\vx; \bm{\xi}),\\
\end{aligned}
\end{equation}
respectively. This is the case that we will demonstrate in Sec.~\ref{sec:results}.
% \subsubsection{Meta-learning}
% Meta-learning is a paradigm where a machine learning model gains experience over multiple learning episodes - often covering a distribution of related tasks - and uses this experience to improve its future learning performance \cite{hospedales2020meta}. 

We will discuss the architecture of the generative model $G_{\bm{\eta}}(\vx; \bm{\xi})$ in Sec.~\ref{sec:architecture}.

\subsection{Posterior estimation with physics-informed likelihood}\label{sec:post}

Once $\tilde{u}_{\bm{\eta}}(\vx;\bm{\xi})$ is well-trained, it can be viewed as a map that transports a distribution of $\bm{\xi}$ to the functional distribution of $u$, and in particular, the standard multivariate Gaussian distribution to the learned functional prior of $u$. Similarly for other terms. Instead of performing posterior estimation in the functional space, we can now switch to the Euclidean space, estimate the posterior of $\bm{\xi}$ conditioned on the new data via Markov Chain Monte Carlo (MCMC). With the generators, the posterior distribution of $\bm{\xi}$ can then be transported to the posteriors of $u, b$ and $f$ in the functional space, which serve as our prediction with uncertainty quantification.

Given the new data $\mathcal{D} = \mathcal{D}_u \cup \mathcal{D}_f \cup \mathcal{D}_b$ defined in Equation~\ref{eqn:testdata}, the likelihood can be written as:
\begin{equation}
\label{eqn:ganlikelihood}
\begin{aligned}
    P(\mathcal{D}|\boldsymbol{\bm{\xi}}) &= P(\mathcal{D}_u|\bm{\xi}) P(\mathcal{D}_f|\bm{\xi}) P(\mathcal{D}_b|\bm{\xi}), \\
     P(\mathcal{D}_u|\bm{\xi}) &= \prod_{i=1}^{N_u} \frac{1}{\sqrt{2\pi{\sigma_u^{(i)}}^2}}\exp \left(-\frac{(\tilde{u}_{\bm{\eta}}(\boldsymbol{x}_{u}^{(i)}; \bm{\xi}) - u^{(i)})^2}{2{\sigma_u^{(i)}}^2}\right), \\
     P(\mathcal{D}_f|\bm{\xi}) &= \prod_{i=1}^{N_f} \frac{1}{\sqrt{2\pi{\sigma_f^{(i)}}^2}}\exp \left(-\frac{(\tilde{f}_{\bm{\eta}}(\boldsymbol{x}_{f}^{(i)}; \bm{\xi}) - f^{(i)})^2}{2{\sigma_f^{(i)}}^2}\right), \\
    P(\mathcal{D}_b|\bm{\xi}) &= \prod_{i=1}^{N_b} \frac{1}{\sqrt{2\pi{\sigma_b^{(i)}}^2}}\exp \left(-\frac{(\tilde{b}_{\bm{\eta}}(\boldsymbol{x}_{b}^{(i)}; \bm{\xi}) - b^{(i)})^2}{2{\sigma_b^{(i)}}^2}\right). \\
\end{aligned}
\end{equation}

Combined with the Gaussian prior $P(\bm{\xi}) = (2\pi)^{-d_{\bm{\xi}}/2}\exp(-\Vert \bm{\xi} \Vert^2/2)$, where $d_{\bm{\xi}}$ is the dimensionality of $\bm{\xi}$, 
the posterior is obtained from Bayes' theorem:
\begin{equation}\label{eqn:Bayes}
\begin{aligned}
P(\bm{\xi}| \mathcal{D}) = \frac{ P(\mathcal{D}|\bm{\xi})P(\bm{\xi})}{P(\mathcal{D})} \simeq P(\mathcal{D}|\bm{\xi})P(\bm{\xi}),
\end{aligned}
\end{equation}
where ``$\simeq$'' represents equality up to a constant. 
Equation~\ref{eqn:Bayes} provides the unnormalized density of the posterior, thus we can use MCMC, in particular, No-U-Turn (which is a Hamitonian Monte Carlo method with adaptive path lengths) \cite{hoffman2014no}, to sample from $P(\bm{\xi}| \mathcal{D})$ , denoted as $\{{\bm{\xi}}^{(i)}\}_{i=1}^M$. Consequently, we obtain the posteriors of $u$ from samples $\{\tilde{u}_{\bm{\eta}}(\boldsymbol{x}, {\bm{\xi}}^{(i)} ) \}_{i=1}^M$.  We focus mostly on the mean and standard deviation of $\{\tilde{u}_{\bm{\eta}}(\boldsymbol{x}, {\bm{\xi}}^{(i)} ) \}_{i=1}^M$, since the former represents the prediction of  $u(\boldsymbol{x})$ while the latter quantifies the uncertainty. Similarly for the other terms.

\subsection{Generator architecture}~\label{sec:architecture}
The generative models take the form of $G_{\bm{\eta}}(\vx; \bm{\xi})$, where $\bm{\eta}$ are the trainable parameters, $\vx$ and $\bm{\xi}$ are the inputs. It is a generalization of the Bayesian neural network $\tilde{u}(\vx; \bm{\theta})$ with parameter $\bm{\theta}$, which is used in~\cite{yang2021b} to solve physical problems. To get different functional priors, in $\tilde{u}(\vx; \bm{\theta})$ we can only tune the architecture and activation functions, or the priors distributions for $\bm{\theta}$. Here, in $G_{\bm{\eta}}(\vx;\bm{\xi})$, the prior for $\bm{\xi}$ is fixed, e.g., Gaussian, and we can tune $\bm{\eta}$ for different functional priors. 
% But we remark that $\tilde{u}(\vx; \bm{\theta})$ can also be viewed as a special case of the generator $G_{\bm{\eta}}(\vx;\bm{\xi})$, if we reparameterize $\bm{\theta}$ with $\bm{\xi}$ and $\bm{\eta}$, e.g., $\bm{\theta} = \bm{\xi} \odot \bm{\eta}_1 + \bm{\eta}_2$, $\bm{\eta} = (\bm{\eta}_1,\bm{\eta}_2)$.

Inspired by the Karhunen-Lo\`eve expansion, in this paper, we set $G_{\bm{\eta}}(\vx; \bm{\xi})$ as the inner product of two subnetworks, i.e., $G_{\bm{\eta}}(\vx;\bm{\xi}) = \tilde{g}_{\bm{\eta_1}}(\vx) \cdot \tilde{h}_{\bm{\eta_2}}(\bm{\xi})$, where $\tilde{g}_{\bm{\eta_1}}$ and $\tilde{h}_{\bm{\eta_2}}$ are two neural networks with $\bm{\eta_1}$ and $\bm{\eta_2}$ as parameters, respectively, and $\bm{\eta} = (\bm{\eta}_1, \bm{\eta}_2)$. The two subnetworks share the same output dimension $d_G$. We can see here that  $\tilde{g}_{\bm{\eta_1}}(\vx)$ acts as the ``basis'' and $\tilde{h}_{\bm{\eta_2}}(\bm{\xi})$ acts as the ``random variables''. Note that $\tilde{h}_{\bm{\eta_2}}(\bm{\xi})$ is not limited to be Gaussian, thus $G_{\bm{\eta}}(\vx;\bm{\xi})$ can represent non-Gaussian stochastic processes.

Moreover, in this paper we set $d_G$ the same as the dimension of $\bm{\xi}$, and $\tilde{g}_{\bm{\eta_1}}$ is a MLP. In our preliminary study, we empirically found that for the neural network $\tilde{h}_{\bm{\eta_2}}$, a ResNet-like architecture performs better than a vanilla MLP. Specifically, we set $\tilde{h}_{\bm{\eta_2}}(\bm{\xi}) = \bm{\xi} + \tilde{h}^*_{\bm{\eta_2}}(\bm{\xi})$, where $\tilde{h}^*_{\bm{\eta_2}}$ is a MLP. More comparisons on the generator architectures are presented in \ref{sec:comparison}.

% In this paper we use the architecture consists of two subnetworks, the first one takes a noise $\bm{\xi}$ as input, and the other employs $\bm{x}$ as the input, the output of the generator is the inner-product of the outputs of these two subnetworks, i.e., 
% \begin{align}
%     G_{\bm{\theta}_g}(\bm{x}, \bm{\xi}) = \sum^{D_g}_{i=1} \phi_{\bm{\theta}_{\bm{x}}}(\bm{x}) \psi_{\bm{\theta}_{\xi}}(\bm{\xi}),
% \end{align}
% where $D_g$ is the number of dimensions of the output for each subnetwork, $\psi_{\bm{\theta}_{\bm{xi}}}$ represents the first neural networks  parameterized by $\bm{\theta}_{\bm{\xi}}$,   $\phi_{\bm{\theta}_{\bm{x}}}$ is  the second neural networks parameterized by $\bm{\theta}_{\bm{x}}$, and $\bm{\theta}_g = \bm{\theta}_{\bm{\xi}} \cup \bm{\theta}_{\bm{x}}$ denotes all the hyperparameters in the generator.

Finally, we remark that while we take Equation~\ref{eqn:Nequ} as an example in this section, where only $u$, $b$ and $f$ are involved in the PDE, it is not hard to generalize the method to other cases with more terms. For example, if we have another function (or variable) $k$ in the left-hand-side of the equation, then we can use another neural network, which takes $(\bm{x}, \bm{\xi})$ (or $\bm{\xi}$) as input to represent $k$ ($\bm{\xi}$ should be shared with other terms), and the DeepONet should map $(k,f)$ (and optionally $b$) to $u$.

\section{Results and Discussion}
\label{sec:results}

In this section, we first employ the proposed method for regression in meta-learning. Specifically, the performance of the present approach will be compared to a baseline meta-learning method, i.e., model agnostic meta-learning (MAML) \cite{finn2017model} (details on meta-learning and MAML can be found in \ref{sec:maml}). Then in Sec.~\ref{sec:meta_pde}, we test 1D forward and inverse PDE problems with physics encoded by automatic differentiation as in PINNs. We also apply the proposed approach in conjunction with DeepONets for reactive transport in heterogeneous porous media. In particular, we infer the fractional order in a diffusion-reaction system and the conductivity field for a porous media flow problem with uncertainties. Finally, we test the method on a regression problem to demonstrate how to tackle time- and space-dependent problems. Details, e.g.,  architectures, training steps, setup in HMC, etc., for each case are presented in \ref{sec:dnn_arch}.

% [NN architecture, hyperparameters here.]

\subsection{A pedagogical example: learning functional priors for meta-learning}\label{sec:meta_func}

We start with a regression problem, which has been used as benchmark in meta-learning \cite{finn2017model}, to demonstrate the performance of the present method. Specifically, the performance of the proposed method will be compared to a baseline meta-learning method, i.e., model agnostic meta-learning (MAML) \cite{finn2017model}, in the function approximation case  (more details on meta-learning and MAML can be found in \ref{sec:maml}). To form the task distribution $p(\mathcal{T})$ (see more details on $p(\mathcal{T})$ in \ref{sec:maml}), we consider a family of functions as follows:
\begin{align}
    u &= A \sin(\omega x), ~ x \in [-1, 1], \\
    A & \sim \mathcal{U}([1, 3]), ~ \omega \sim \mathcal{U} ([2, 12]), 
\end{align}
where $\mathcal{U}$ represents uniform distribution. Similar as the setup in MAML \cite{finn2017model}, we assume that we have prior knowledge on this task distribution, which is reflected by the historical data or meta-training data $\overline{\mathcal{D}}$. Our goal is to make effective use of the prior knowledge to improve the model performance by recalling relevant knowledge learned from $\overline{\mathcal{D}}$. Note that in all cases, we assume that the task labels, e.g., $A$, and $\omega$, are unknown, and we only have access to the data $\overline{\mathcal{D}}$. 

% Specifically, we have numbers of pairs of input/output, i.e., $(\bm{x}, \bm{u})_{i=1, ..., N}$ as meta training data. Then we assume that we have a few measurements on $u$, i.e., $(\bm{x}, \bm{u})_K$ for a new task that is different from $(\bm{x}, \bm{u})_{i=1, ..., N}$. We now would like to infer $u$ for $x \in [-1, 1]$.

Here we randomly draw $N = 2,000$ pairs of $(A, \omega)$ to generate historical data $\overline{\mathcal{D}}$ for learning the whole task distribution. For each $u$ sample, we use 30 equidistant sensors to resolve it.  In addition, we assume that we only have a few new noisy measurements ${\mathcal{D}}$ at the posterior estimation stage. As for the historical data $\overline{\mathcal{D}}$, we test two different cases: Case (1) $\overline{\mathcal{D}}$ are noise free; and Case (2) $\overline{\mathcal{D}}$ have larger noise scale than the new noisy measurements used at the posterior estimation stage (i.e., ${\mathcal{D}}$). The first case can happen when the data come from high-fidelity simulations, or are collected from high-fidelity sensors in test site, while the second case can happen when the data are collected years or decades ago so that the simulations or sensors are of low-fidelity.

\begin{figure}[H]
    \centering
    \subfigure[]{\label{fig:meta_funca}
    \includegraphics[width=0.3\textwidth]{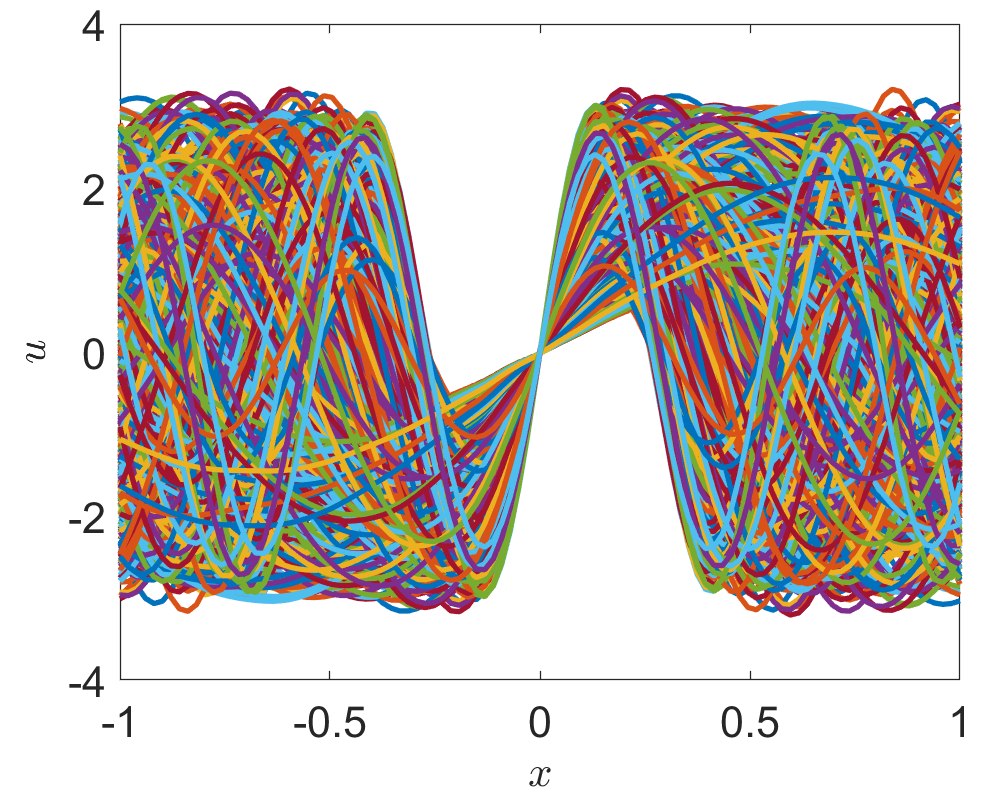}
    \includegraphics[width=0.3\textwidth]{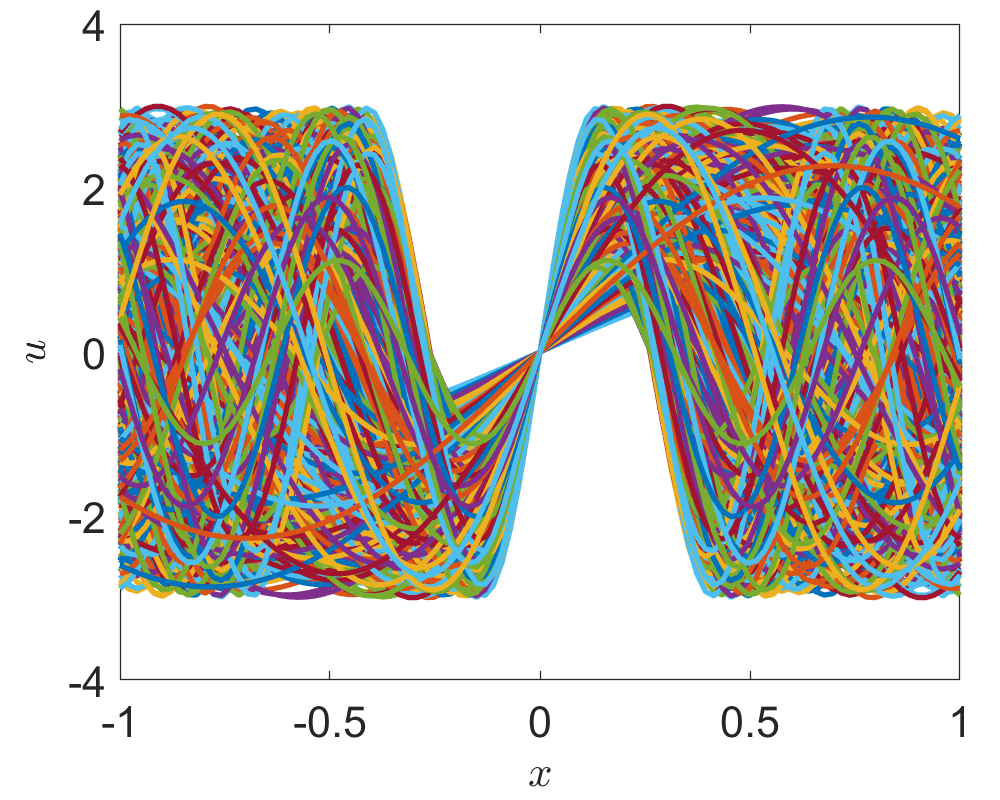}}
     \subfigure[]{\label{fig:meta_funcb}
    \includegraphics[width=0.3\textwidth]{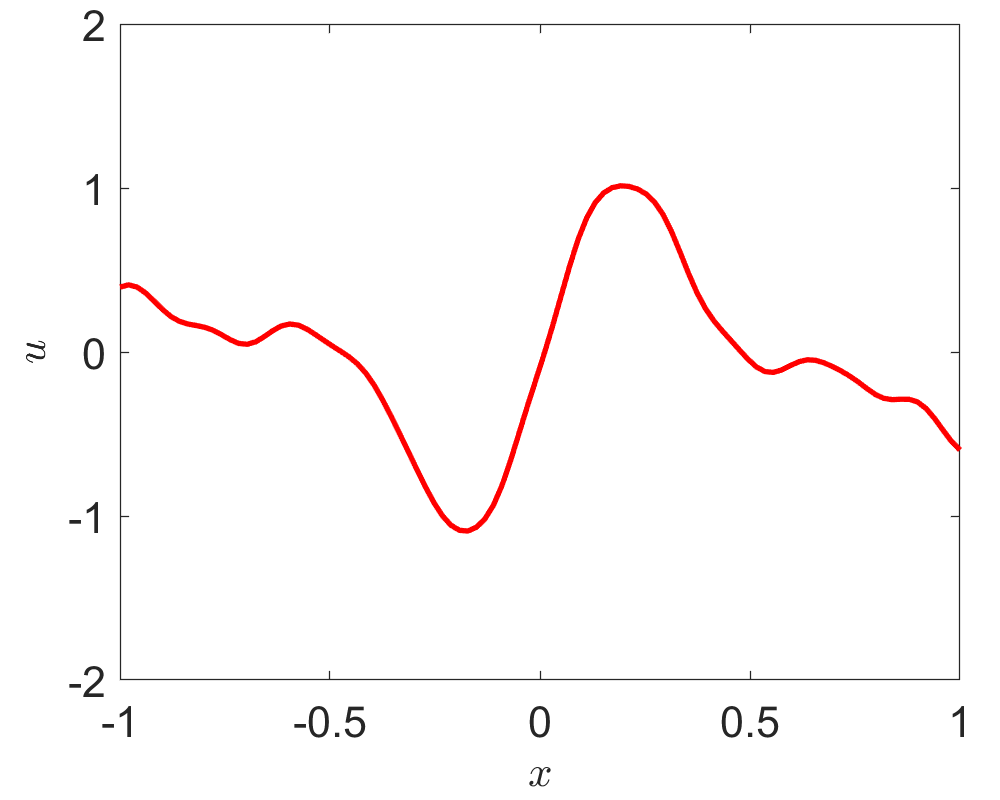}}
    \subfigure[]{\label{fig:meta_funcc}
    \includegraphics[width=0.3\textwidth]{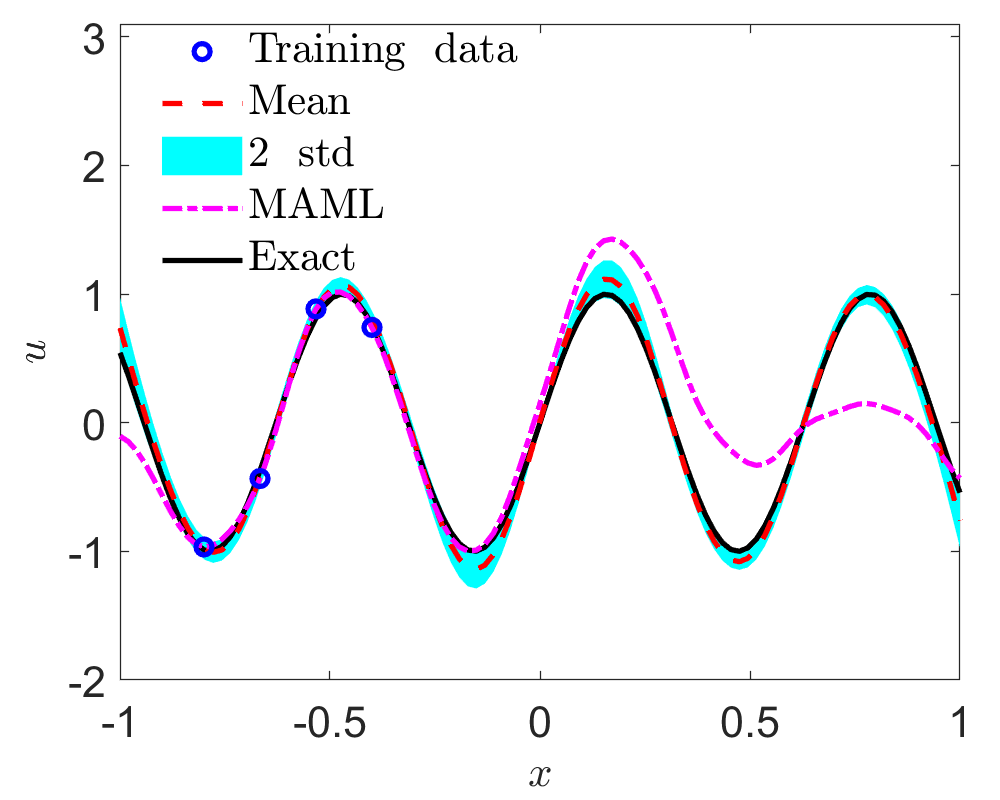}
    \includegraphics[width=0.3\textwidth]{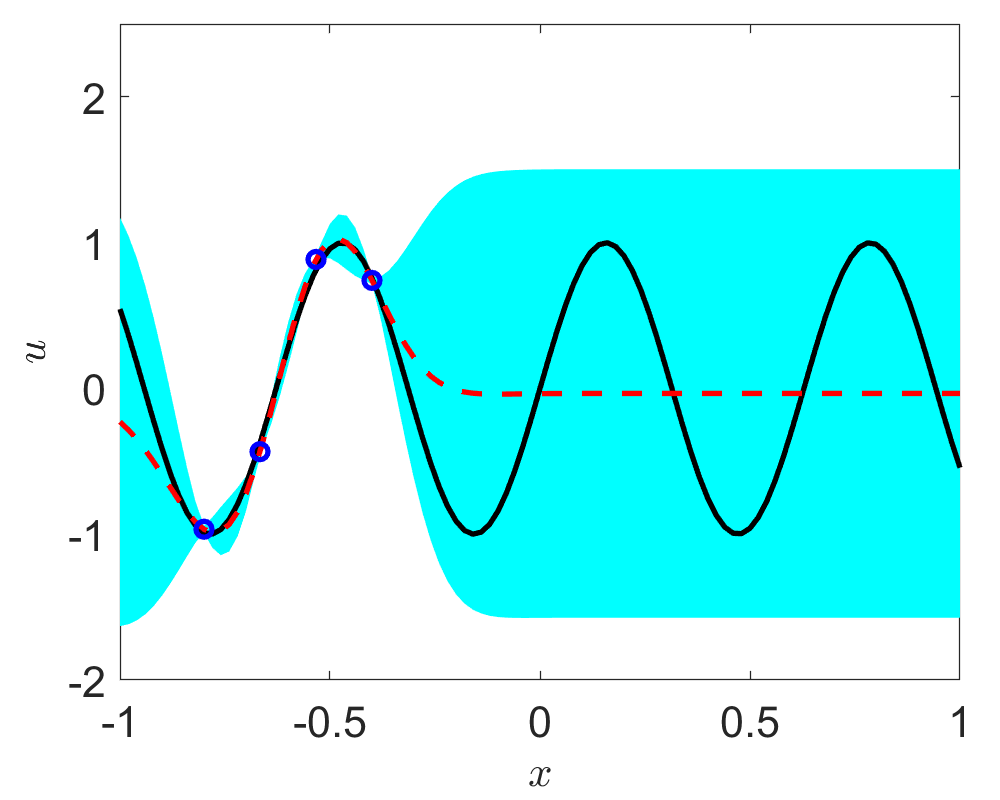}}
    \subfigure[]{\label{fig:meta_funcd}
    \includegraphics[width=0.3\textwidth]{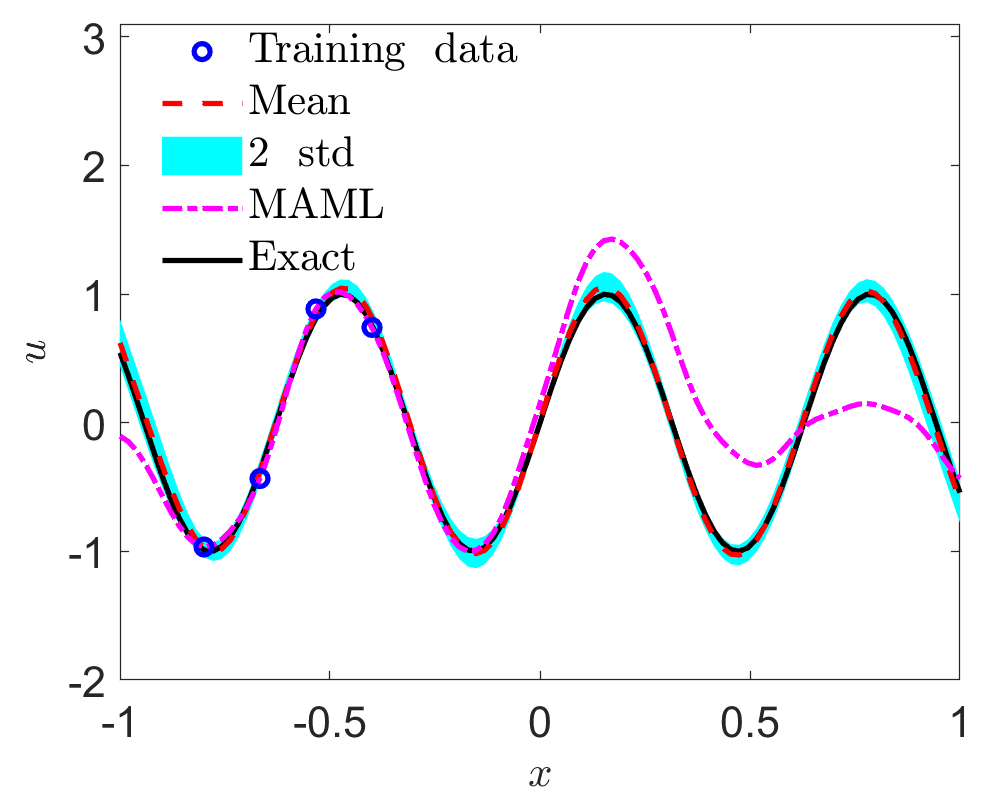}}
     \subfigure[]{\label{fig:meta_funce}
    \includegraphics[width=0.35\textwidth]{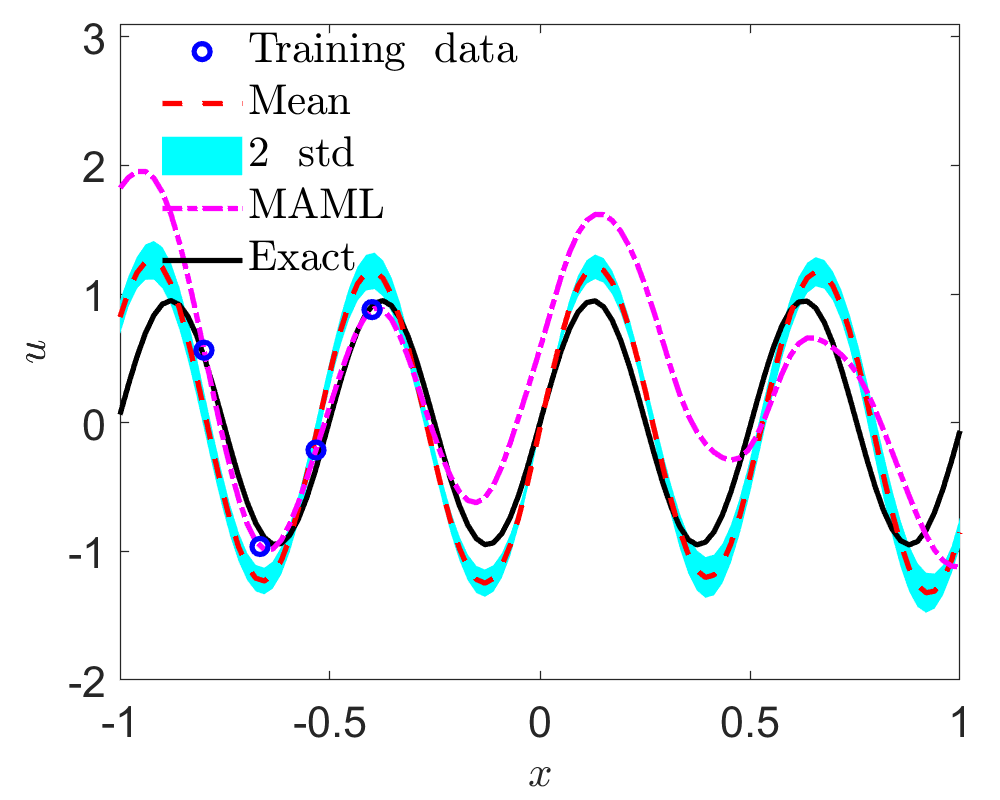}}
    \caption{
    Meta-learning for function approximation with uncertainty quantification. 
    (a) First and second column: Functional prior for $u$: 1,000 random samples from $u$.
    The first column is the learned functional prior, and the second column is the exact samples for $u$. 
    (b) Predictions from the MAML after meta-training. 
    (c) First column: Predicted $u$ from the present approach as well as MAML \cite{finn2017model} using clean $\overline{\mathcal{D}}$ and noisy $\mathcal{D}$ with noise scale: $\mathcal{N}(0, 0.05^2)$.
    Second column: Predictions for $u$ using GPR with a square kernel. 
    (d) Predicted $u$ for noisy $\overline{\mathcal{D}}$ and ${\mathcal{D}}$ from the present method. The noise scales for $\overline{\mathcal{D}}$ and ${\mathcal{D}}$ are  $\mathcal{N}(0, 0.1^2)$ and $\mathcal{N}(0, 0.05^2)$, respectively.
    (e) Predicted $u$ for the case where the ground truth (black solid line) is out of the space of the learned functional prior. Noise for $\overline{\mathcal{D}}$ and ${\mathcal{D}}$ are $\mathcal{N}(0, 0.1^2)$ and $\mathcal{N}(0, 0.05^2)$, respectively. Mean: predicted mean; 2 std: predicted two standard deviations. 
    }
    \label{fig:meta_func}
\end{figure}

The generator takes a Gaussian noise and the coordinate $\bm{x}$ as input and outputs $\tilde{u}_{\bm{\eta}}(\bm{x}; \bm{\xi})$. An illustration of the learned functional prior for Case (1) is presented in Fig. \ref{fig:meta_funca}. Upon completion of learning the functional prior, we assume that we have only 4 noisy measurements for a new task, i.e., $y = \sin(10 x)$, which are equidistantly distributed in $x \in [-0.8, -0.4]$.  Specifically, the noise scale for the measurement is assumed to satisfy a Gaussian distribution, i.e., $\mathcal{N}(0, 0.05^2)$. The objective now is to infer $u$ in the whole domain using the learned functional prior as well as the few measurements. 

As shown in the first column of Fig. \ref{fig:meta_funcc}, the predicted means from the present method are in good agreement with the exact solution, while the predicted uncertainties are quite small in the whole domain, which is consistent with the computational errors between the predicted means and the exact solution. By learning the functional prior, the generator has a memory on each task in the task distribution (as shown in the first column of Fig. \ref{fig:meta_funca}), which will be recalled during the inference of a new task. Therefore, we can use only partial measurements on the new task to fully resolve the function. Furthermore, we present the results from MAML for comparison (Details on the setup of MAML are in \ref{sec:maml}). As displayed in the first column of Fig. \ref{fig:meta_funcc}, the predictions at the region where we have measurements agree well with the exact solution. For regions without measurements, the MAML is still able to infer the phase for the sine function, but great discrepancy is observed between the predicted and exact amplitude. Note that all tasks share the same model initialization in the MAML, as shown in the Fig. \ref{fig:meta_funcb}. Consequently, the MAML can only learn prior knowledge on the ``average'' of all tasks from the historical data. Employing the ``average'' as a starting point, plenty of data are still required to resolve the details of a new task. Therefore, the present method is capable of providing more accurate predictions than the MAML in this particular problem.

We also present the results from the Gaussian Processes Regression (GPR) with a square kernel, in which the prior is optimized by maximizing the marginal likelihood \cite{rasmussen2003gaussian}.   As shown in the second column of Fig. \ref{fig:meta_funcc}, the computational errors between the predicted means and the exact solution are bounded by the two standard deviations, but the uncertainties are much larger than the results from the present method, which indicates the effectiveness of the learned functional prior from data. In other words, the prior knowledge learned from the historical data is able to improve the model performance for new tasks.

We proceed to test the performance of the present method for Case (2), i.e., noisy historical data. Similar as in Case (1),  we randomly draw $N = 2,000$ pairs of $(A, \omega)$ to generate the training data and  30 equidistant sensors to resolve each sample of $u$. In addition, each measurement in the training data is assumed to be perturbed by a Gaussian noise, i.e.,  $\mathcal{N}(0, 0.1^2)$. All the other parameters, e.g., architecture of GANs, number of training steps, training data at the posterior estimation stage, etc., are kept the same as in Case (1). The results are illustrated in Fig. \ref{fig:meta_funcd}, which are similar as the results in Case (1) (first column in Fig. \ref{fig:meta_funcc}) and will not be discussed in detail here.

Finally, we test the case in which the training data used for posterior estimation are from a function out of the space of the learned functional prior. In particular, we  assume that we have  4 noisy measurements from  $u = 0.95 \sin(12.5 x)$ at the posterior estimation stage, with the same sensor placement as above. We employ the same functional prior learned in Case (2), and the predictions are depicted in Fig. \ref{fig:meta_funce}. As shown, neither the present method nor MAML can provide accurate predictions, but the computational errors from the present approach are smaller than MAML for this particular case. Since the prior distribution did not cover the new data, it is reasonable that the posterior, which is based on the prior, cannot match the ground truth of the function using only 4 noisy measurements.
% The above results indicate that more accurate results can be obtained if the historical data can cover the task distribution better. 

% \subsection{PDE problems with PINNs and DeepONets}

% \subsubsection{Noisy Observation for GAN}

\subsection{PINNs: Forward and Inverse PDE problems}
\label{sec:meta_pde}
We proceed to consider the following nonlinear diffusion-reaction system governed by: 
\begin{equation}\label{eq:meta_pde}
\begin{split}
    D \partial^2_x u - k_r u^3 &= f, ~ x \in [-1, 1], \\
    u(-1) &= u(1) = 0,
\end{split}
\end{equation}
where $u$ is the solute concentration, $D = 0.01$ is the diffusion coefficient, $k_r$ is the chemical reaction rate, and $f$ is a source term. We fix the exact solution for this system in both the forward and inverse problem, which is expressed as
\begin{align}\label{eq:meta_pde}
    u = (x^2 - 1) \sum^4_{i=1} \left[\omega_{2i-1} \sin(i \pi x) + \omega_{2i} \cos(i \pi x) \right],
\end{align}
where $\omega_{i}$ are uniformly sampled from $\mathcal{U}([0,1])$, $i = 1,2,...,8$.
The source term $f$ can then be derived based on \Eqref{eq:meta_pde} given the chemical reaction rate $k_r$.

\begin{figure}[H]
    \centering
    \subfigure[]{\label{fig:meta_forwarda}
    \includegraphics[width=0.3\textwidth]{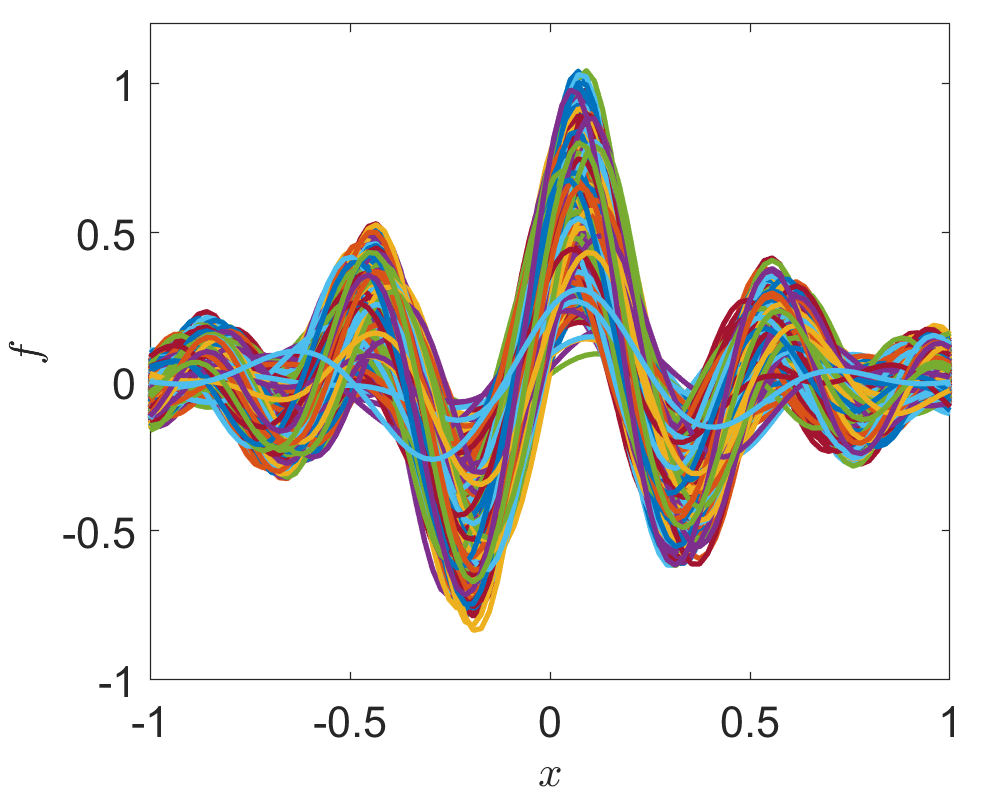}
    }
    \subfigure[]{\label{fig:meta_forwardb}
    \includegraphics[width=0.3\textwidth]{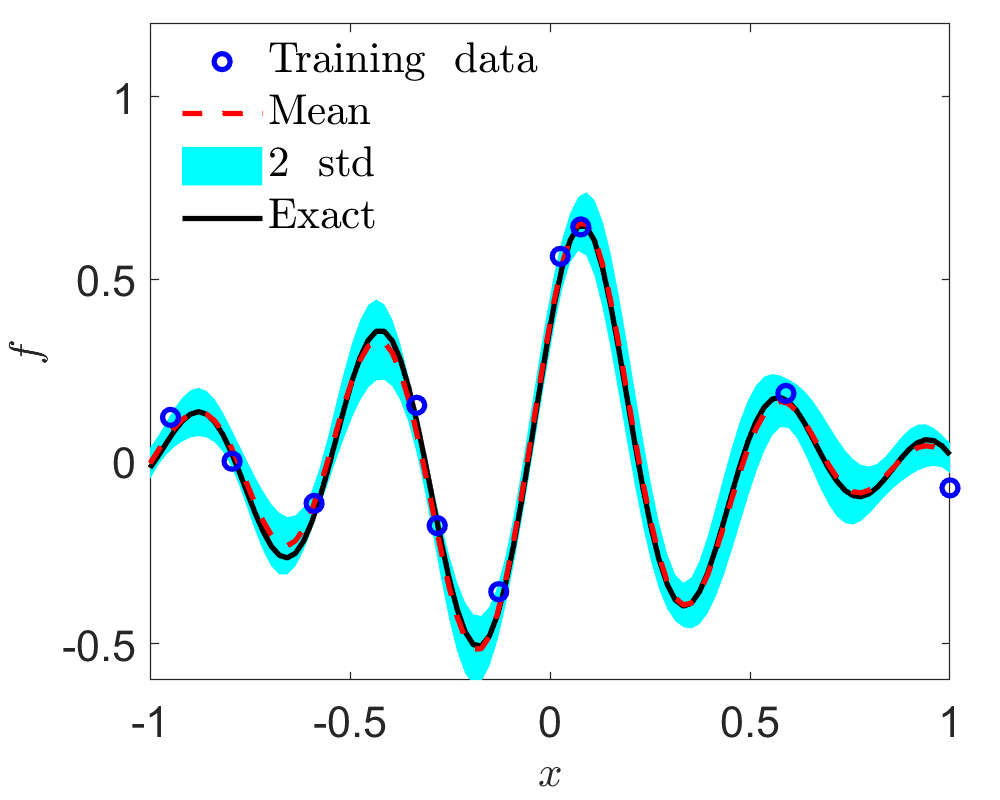}
    \includegraphics[width=0.3\textwidth]{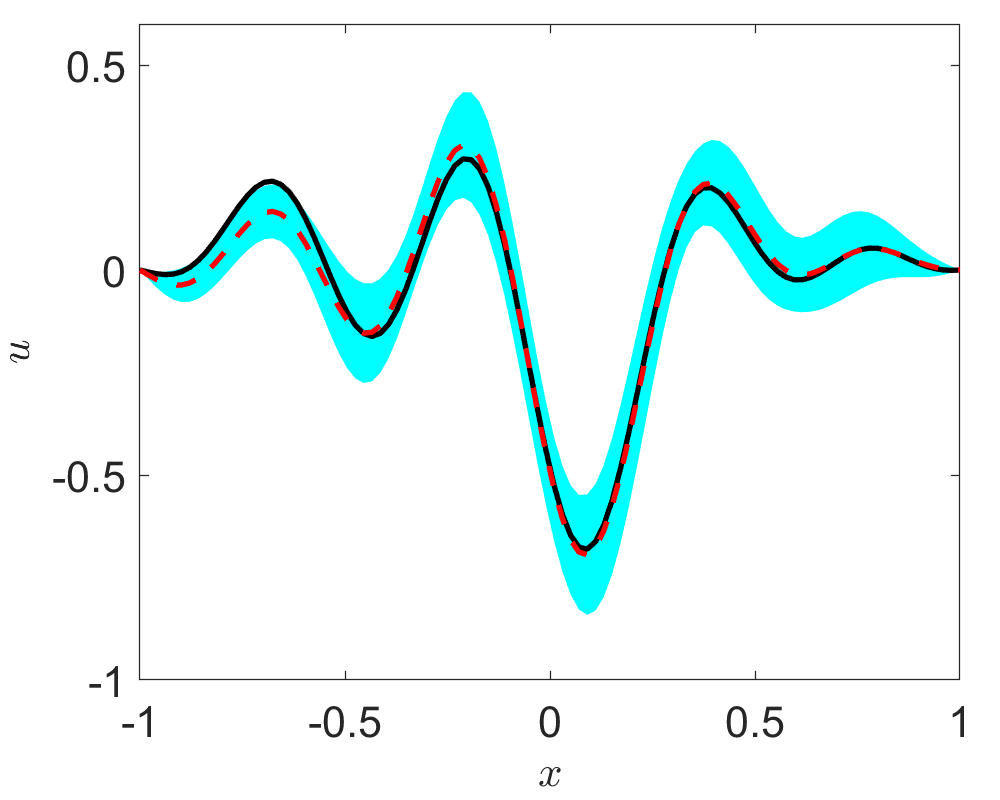}
    }
    \subfigure[]{\label{fig:meta_forwardc}
    \includegraphics[width=0.3\textwidth]{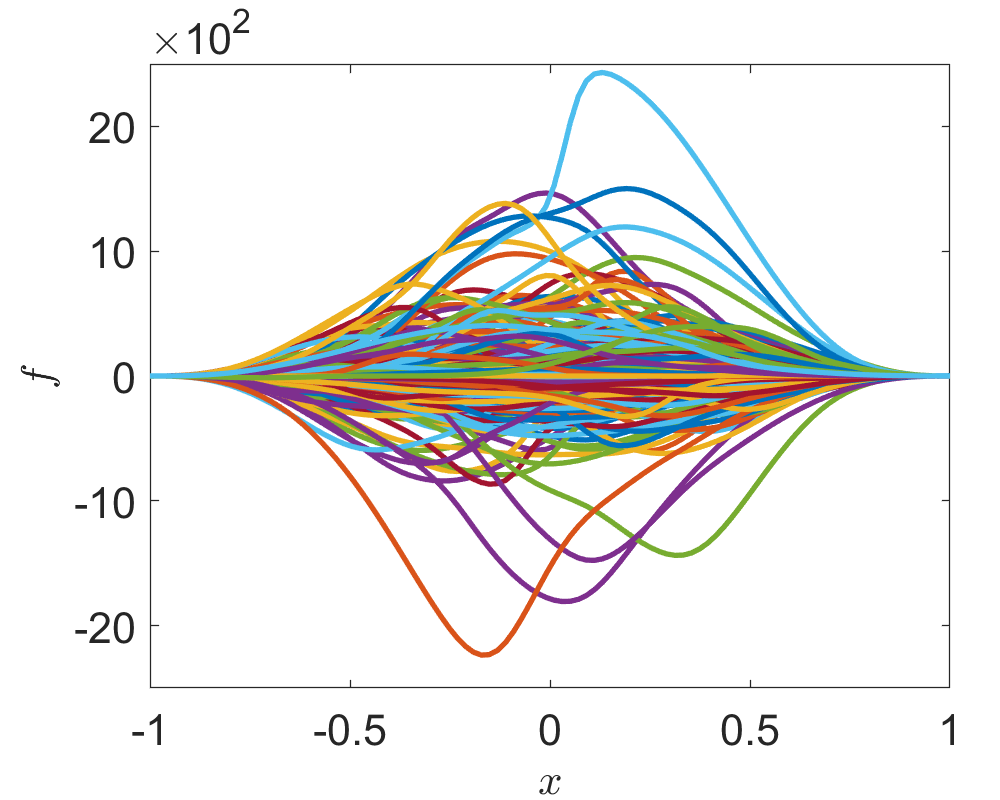}
    }
    \subfigure[]{\label{fig:meta_forwardd}
    \includegraphics[width=0.3\textwidth]{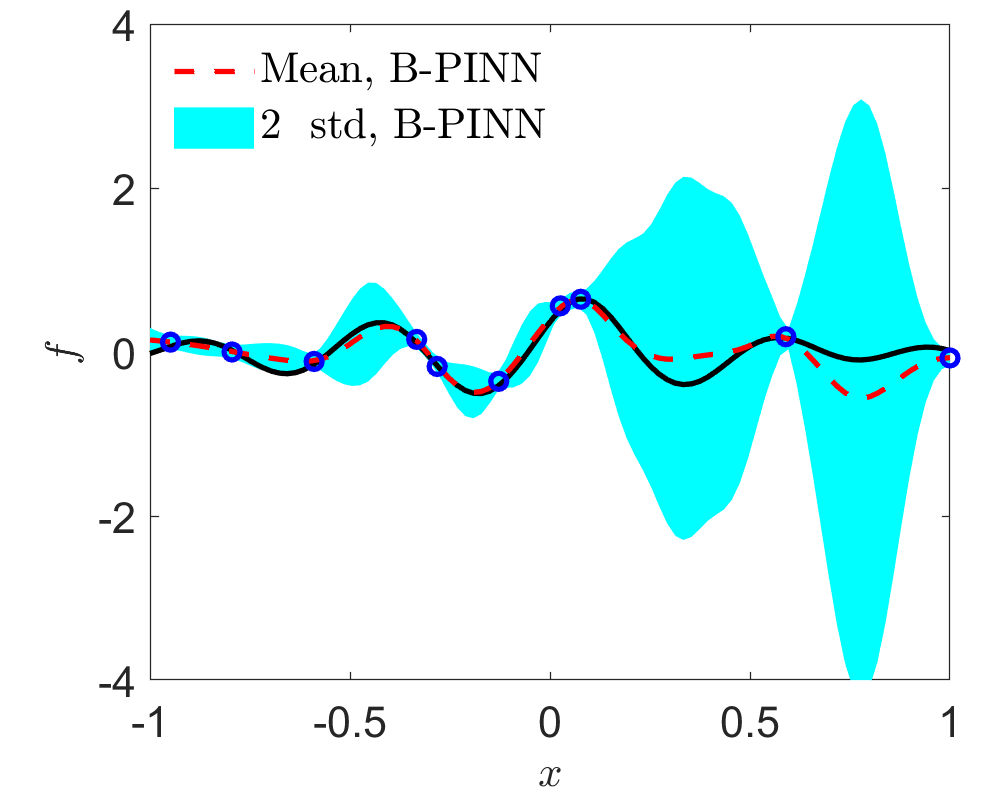}
    \includegraphics[width=0.3\textwidth]{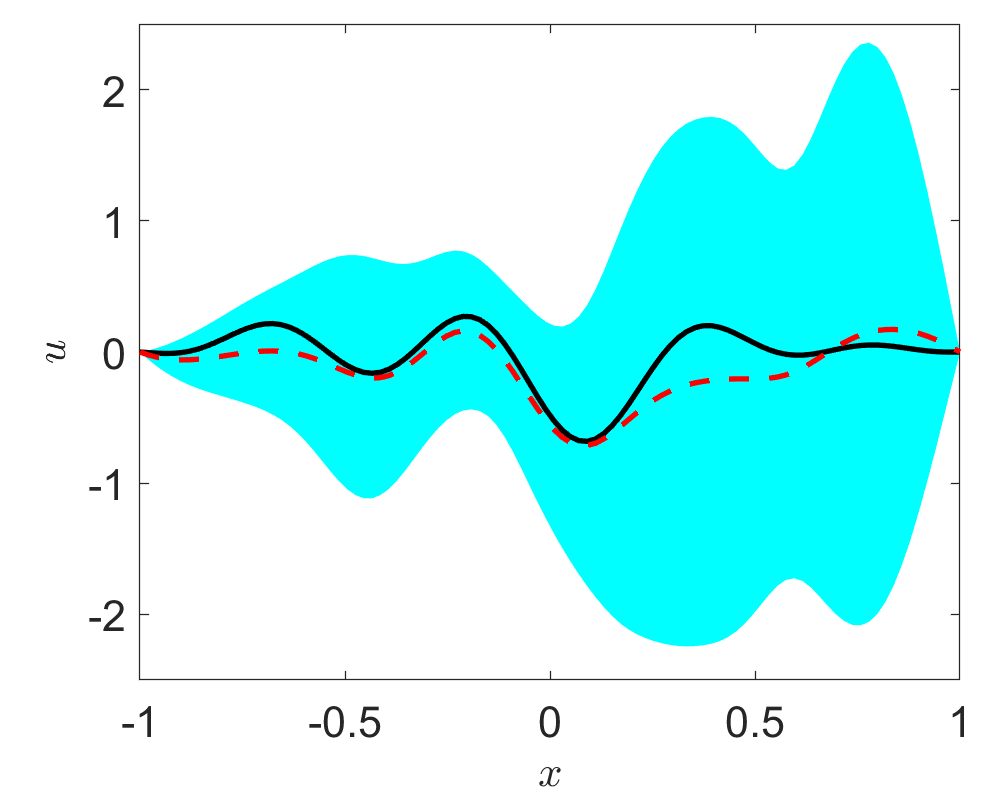}
    }
    \caption{
    Meta-learning for solving forward problem with uncertainty quantification in $D \partial^2_x u - k_r u^3 = f$.
    (a) Functional priors for $f$: 1,000 random samples for $f$.
    (b) From left to right: Predicted $f$ and $u$ from the present method.
    (c) Prior for $f$ from B-PINN. In the BNN, the prior for each weight and bias is identical, i.e., $\mathcal{N}(0, 1^2)$.
    (d) From left to right: Predicted $f$ and $u$ from B-PINN.
    Mean: predicted mean value; 2 std: predicted two standard deviations.
    % The mean and standard deviation are computed using 1,000 posterior samples for the input noise of GANs. 
    }
    \label{fig:meta_forward}
\end{figure}

We first consider a forward problem, in which we assume that $k_r = 0.2$ is a known constant. Similar as in the setup in Sec. \ref{sec:meta_func}, we employ 10,000 samples of $f$ together with the boundary condition of $u$ as the historical data. For each $f$ sample we use 40 equidistant sensors to resolve it. We illustrate the functional prior for $f$ in Fig. \ref{fig:meta_forwarda}. The objective here is to infer both $u$ and $f$ if we have partial measurements on $f$ for a new task. Specifically, 10 random measurements on $f$ are collected at the posterior estimation stage here.

The generator takes Gaussian noise and the coordinate $\bm{x}$ as input and outputs $\tilde{u}_{\bm{\eta}}(\bm{x}; \bm{\xi})$. We can then obtain the right hand side $f$ based on \Eqref{eq:meta_pde} using the automatic differentiation as in PINNs \cite{raissi2019physics,yang2020physics}. 
Once the functional prior is obtained, we then employ the HMC method to estimate the posterior distributions for $f$ and $u$, which are displayed in Fig. \ref{fig:meta_forwardb}. We observe that: (1) The predicted means for both $f$ and $u$ are in good agreement with the exact solution; and (2) The computational errors for both $f$ and $u$ are bounded by two standard deviations in the whole domain, i.e., $x \in [-1, 1]$.
% and (3) The predicted uncertainties , i.e., standard deviation, increases in the region, e.g., $x \in [-0.5, 0]$, without measurements as expected. 

In addition to the present method, the B-PINN developed in \cite{yang2021b} can also be used for quantifying uncertainties in predictions for PDE problems. We then present the results from the B-PINNs for comparison. Note that the priors for B-PINNs, e.g., architecture of BNNs, and prior distributions for weights and biases used in this case are kept the same as in \cite{yang2021b}. As shown in Fig. \ref{fig:meta_forwardd}, the computational errors between the predicted means and exact solutions for $f$ and $u$ are bounded by the  standard deviations, but the errors or predicted uncertainties are much larger than the results from the present method. The above results demonstrate that the prior learned from the historical data is quite informative and is able to enhance the predicted accuracy for the new unseen task.

\begin{figure}[H]
    \centering
    \subfigure[]{ \label{fig:meta_inversea}
    \includegraphics[width=0.4\textwidth]{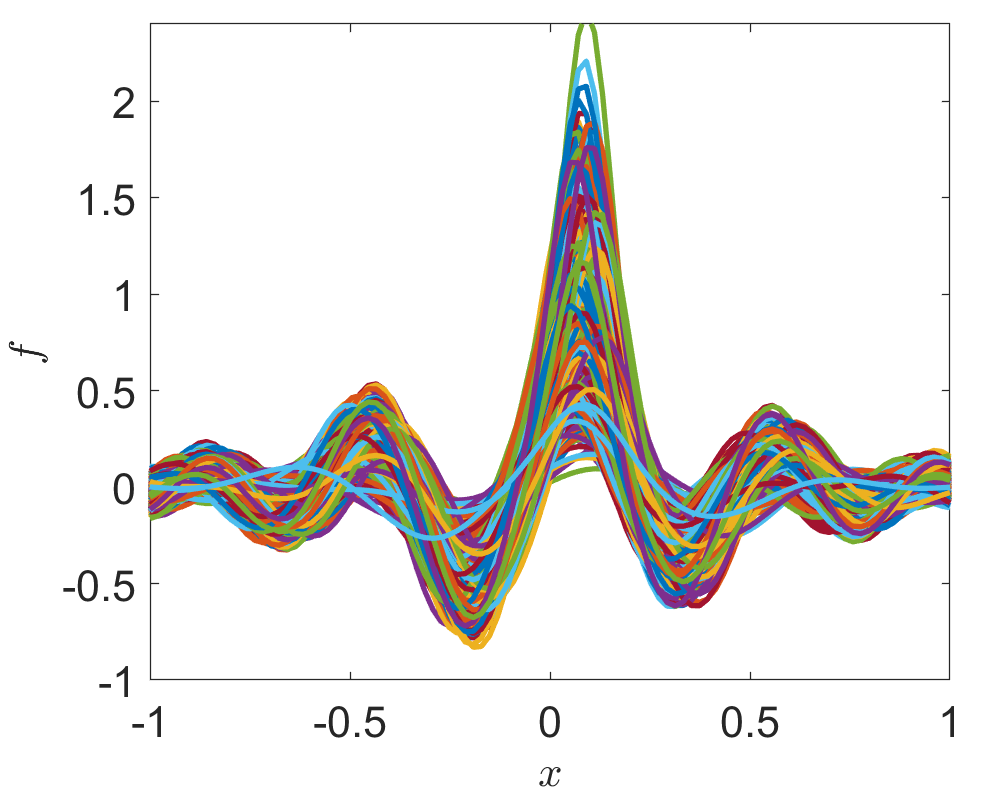}
    \includegraphics[width=0.4\textwidth]{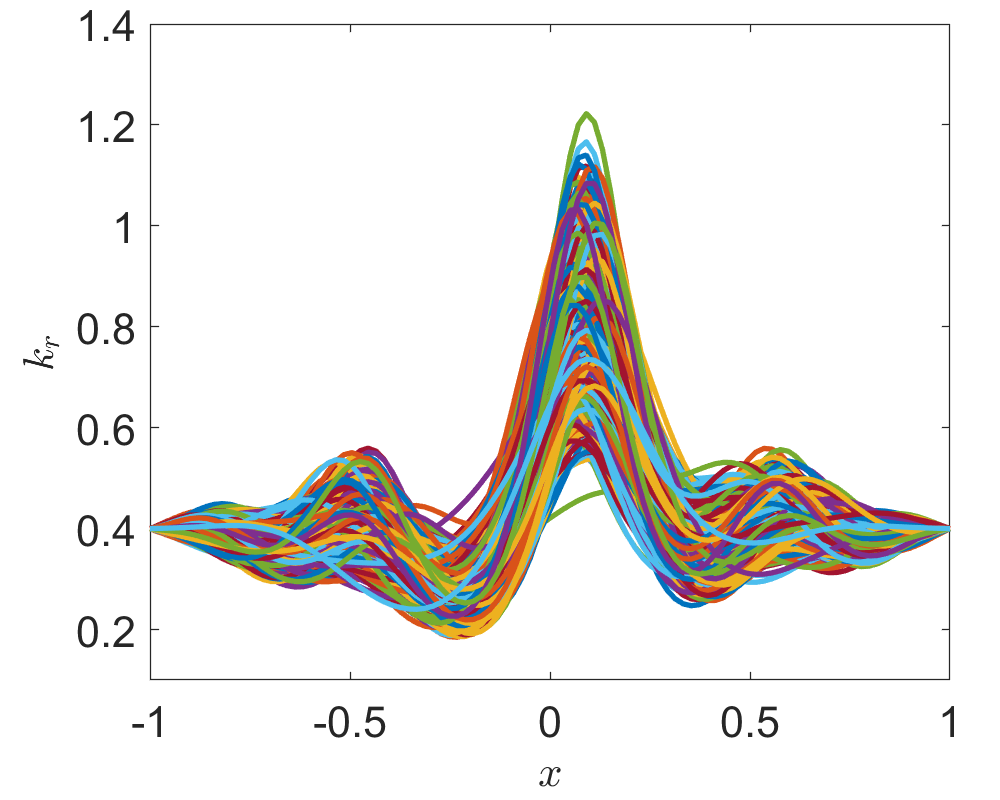}
    }
    \subfigure[]{ \label{fig:meta_inverseb}
    \includegraphics[width=0.3\textwidth]{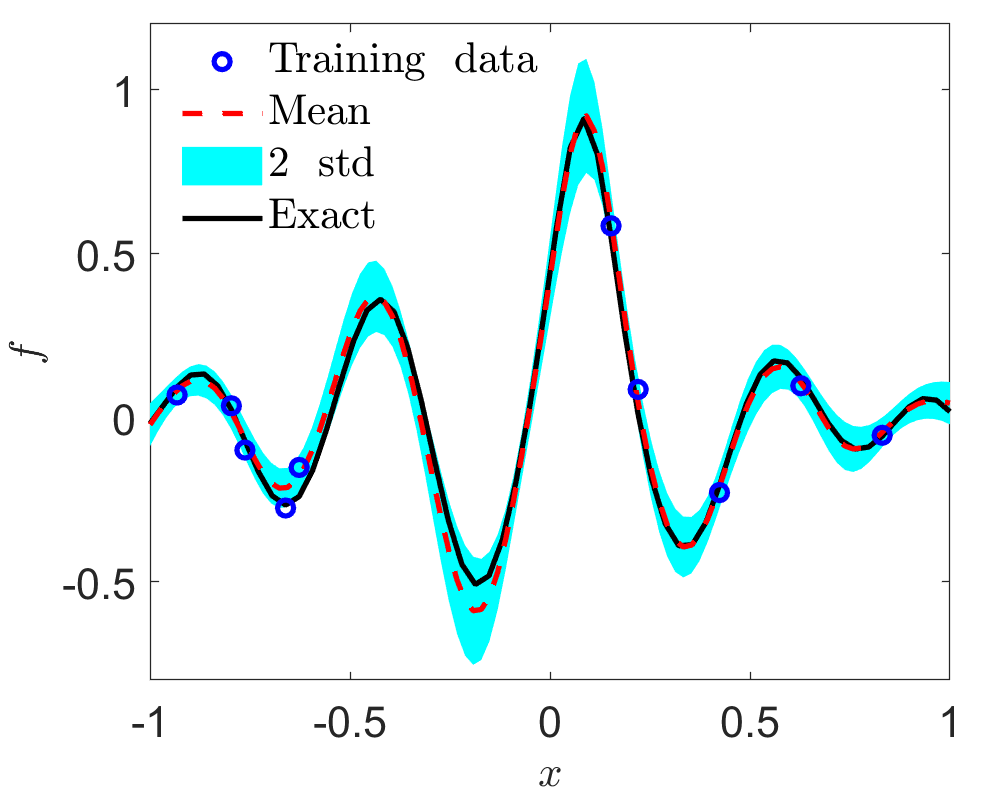}
    \includegraphics[width=0.3\textwidth]{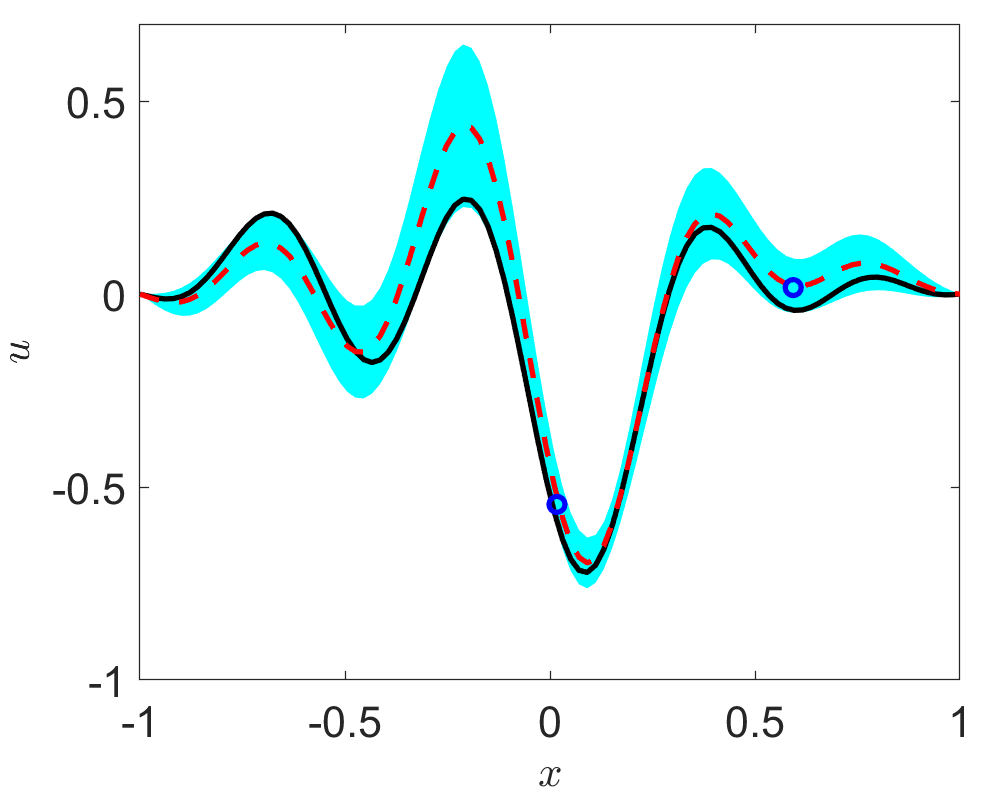}
    \includegraphics[width=0.3\textwidth]{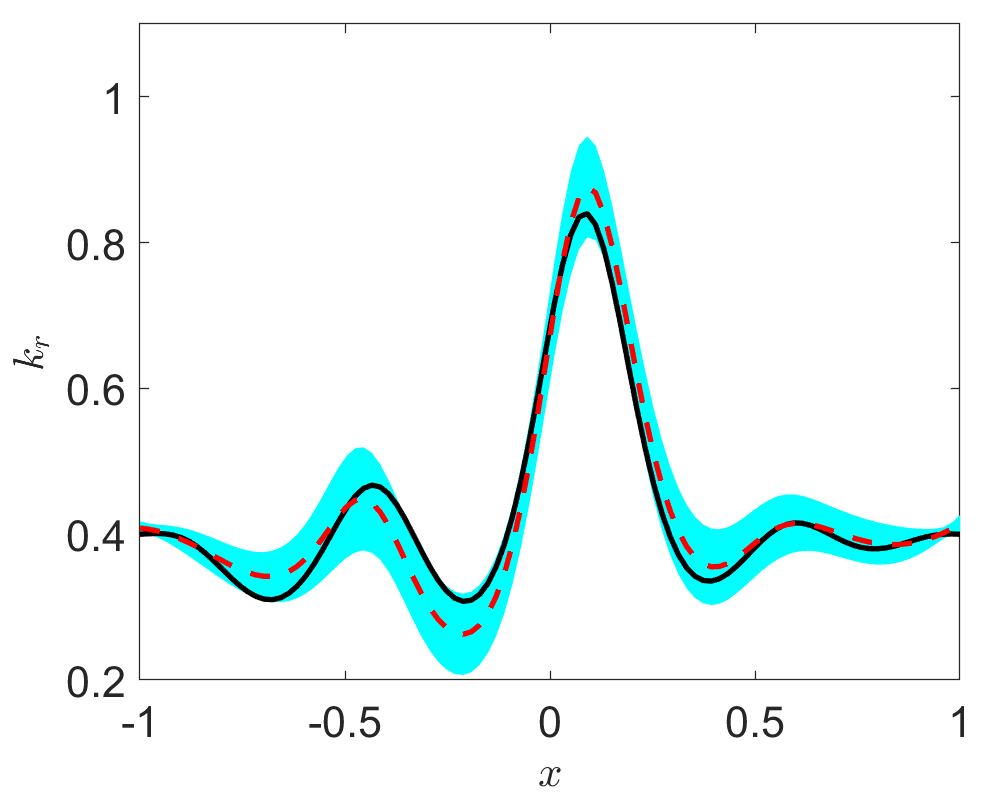}
    }
    \caption{
    Meta-learning for solving inverse problem with uncertainty quantification in $D \partial^2_x u - k_r u^3 = f$. 
    (a) Functional priors for $f$ and $k_r$: 1,000 random samples for both $f$ and $k_r$.
    (b) Predicted $f$, $u$, and $k_r$ from present method. 
    Mean: predicted mean value; 2 std: predicted two standard deviations.
    }
    \label{fig:meta_inverse}
\end{figure}

We now consider an inverse problem, in which $k_r$ is an unknown field. The solution $u$ is assumed to be the same as used in the forward problem, the exact reaction rate is set as a nonlinear function of the solute concentration, i.e.,  $k_r = 0.4 \exp(-u)$, and $f$ can then be derived from \Eqref{eq:meta_pde}. Similarly, we assume that we have 10,000 pairs of $(k_r, f)$ as the historical data for learning the functional priors of $k_r$ and $f$. For each $k_r/f$ sample, we use 40 equidistant sensors to resolve it. Two illustrations for the functional priors of $k_r/f$ are displayed in Fig. \ref{fig:meta_inversea}. We would like to infer $k_r$ with uncertainties based on partial observations on $f$ and $u$ for a new task. In particular, we employ 10 and 2 measurements for $f$ and $u$ as the training data at the posterior estimation stage, respectively.

Here, we employ two generators to generate $k_r$ and $u$, respectively. In particular, the generators share the same input, i.e.,  Gaussian noise $\bm{\xi}$ and the coordinate $\bm{x}$. The right hand side $f$ can then be obtained based on  \Eqref{eq:meta_pde} using automatic differentiation. 
The predicted $f$, $u$ and $k_r$ are displayed in Figs. \ref{fig:meta_inverseb}. We can see that: (1) The predicted means for $f$, $u$ and $k_r$ are in good agreement with the exact solution; and (2) The computational errors for $f$, $u$, and $k_r$ are bounded by  two standard deviations in the whole domain, i.e., $x \in [-1, 1]$.

% \subsection{Uncertainty quantification for DeepONets}
% In this section, we employ the present method to quantify uncertainties in two applications for reactive transport in porous media using DeepONets. In particular, we first test a fractional diffusion with nonlinear reaction in porous media, and then we consider a two-dimensional flow through heterogeneous porous media.

\subsection{DeepONets: Fractional diffusion in heterogeneous porous media}
\label{sec:frac}
We now apply the proposed method to a nonlinear diffusion-reaction system in porous media. In particular,  fractional diffusion is used in the modeling due to the heterogeneity of porous media, which is expressed as
\begin{align}\label{eq:fraceq}
    D \partial^\alpha_x u - k_r u^3 = f, ~ x \in [-1, 1],\;
    u(-1) = u(1) = 0,
\end{align}
where $D = 0.05$, $k_r = 1$, $\alpha$ is from 1 to 2, $\partial^\alpha_x u$ is the $\alpha$-th order Riesz fractional derivative of $u$, and $f$ is a source term. 

\begin{figure}[H]
    \centering
    \subfigure[]{\label{fig:fractionala}
    \includegraphics[width=0.3\textwidth]{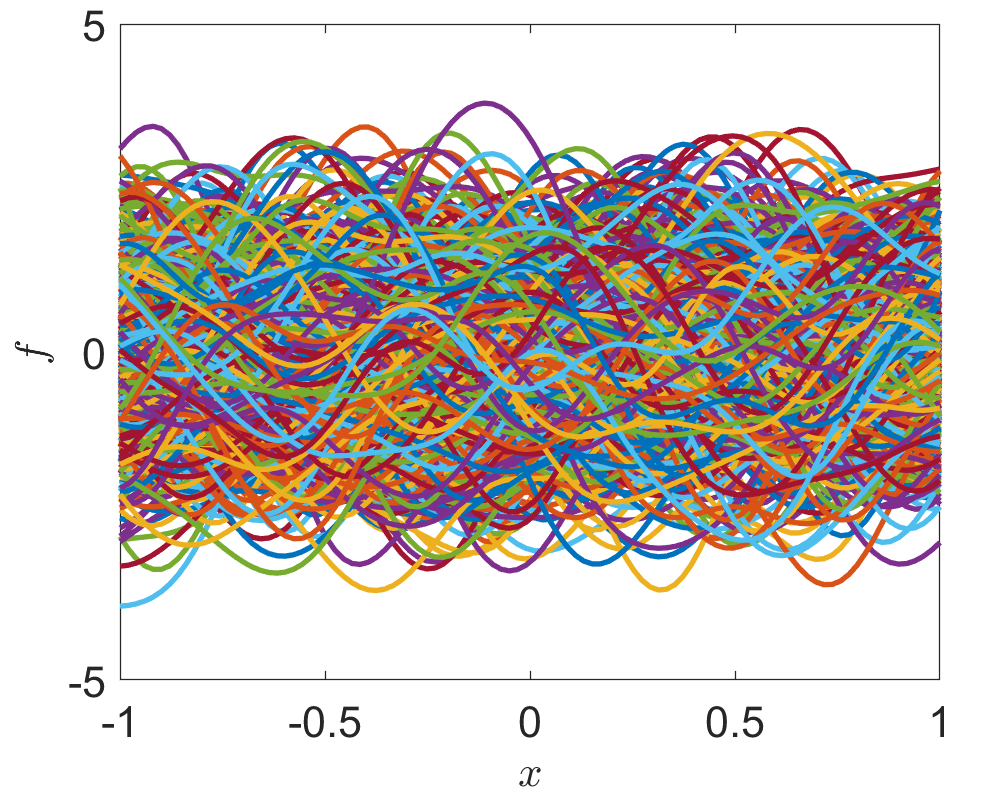}}
    \subfigure[]{\label{fig:fractionalb}
    \includegraphics[width=0.3\textwidth]{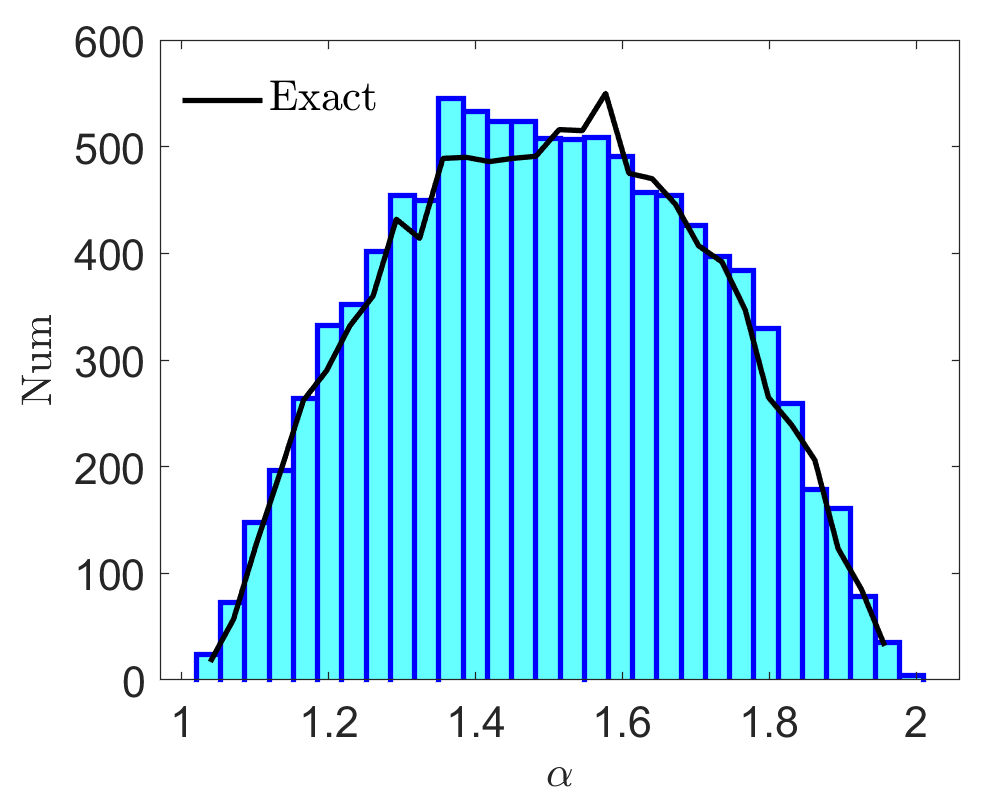}}\\
    \subfigure[]{\label{fig:fractionalc}
    \includegraphics[width=0.3\textwidth]{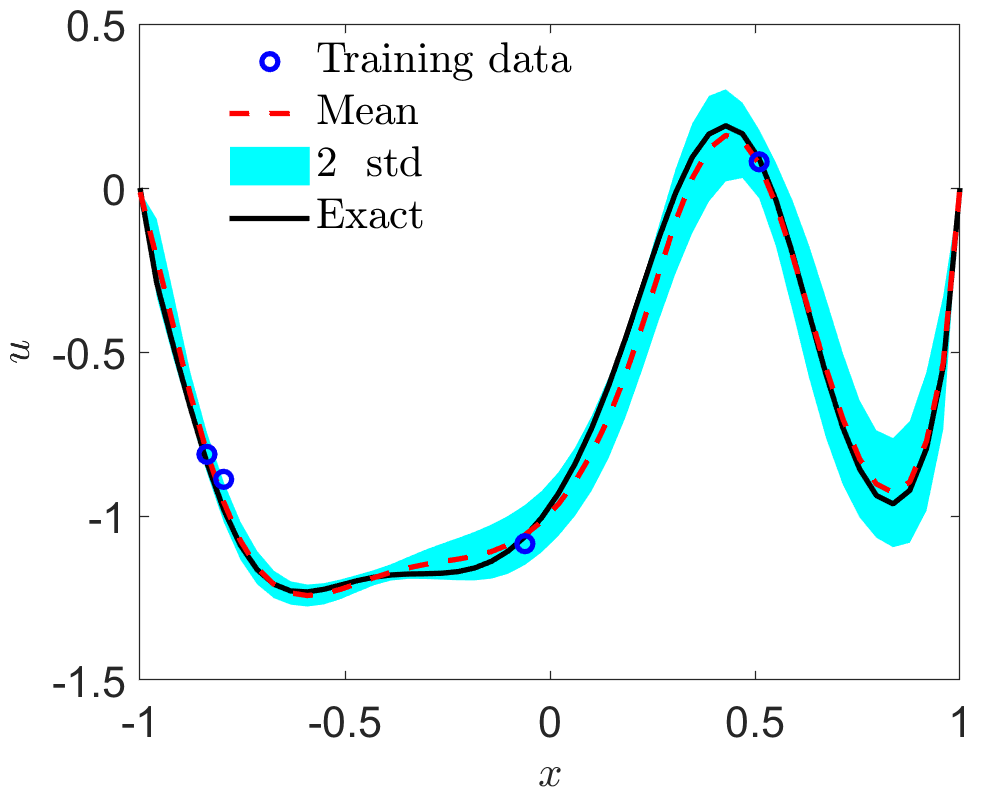}}
    \subfigure[]{\label{fig:fractionald}
    \includegraphics[width=0.3\textwidth]{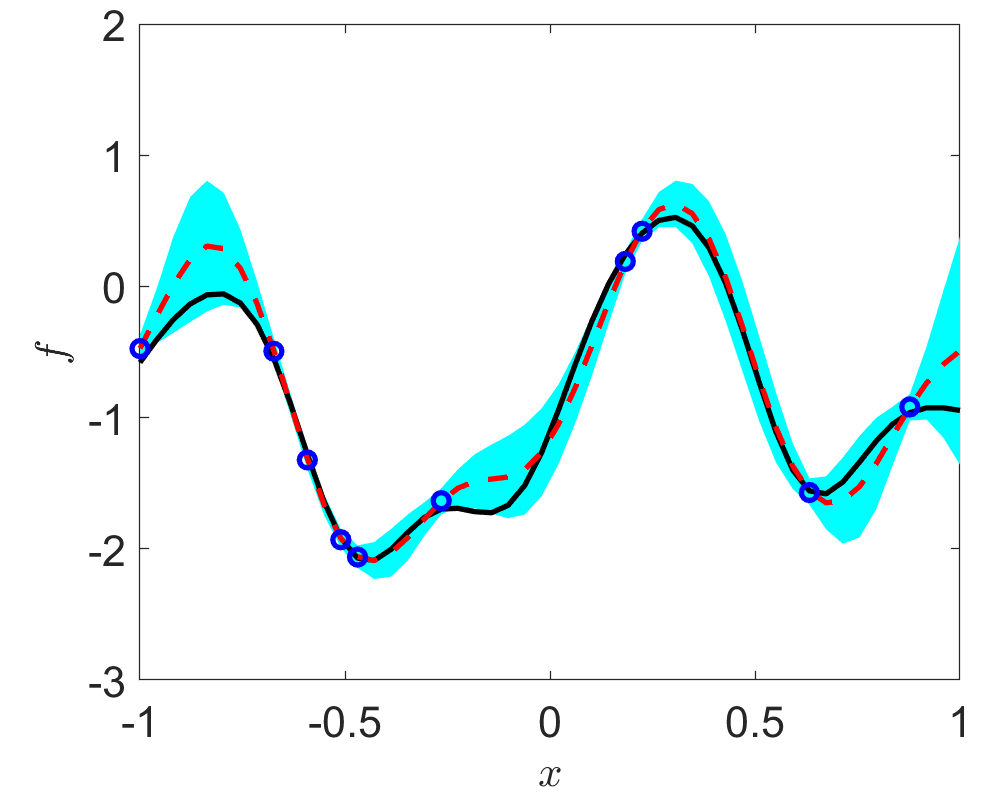}}
    \caption{
    DeepONet for fractional diffusion and nonlinear reaction system in heterogeneous porous media with uncertainty quantification: $D \partial^\alpha_x u - k_r u^3 = f$.
    (a) Functional prior for $f$: 1,000 random samples for $f$.
    (b) Histogram graph for the learned prior for $\alpha$ (10,000 samples).
    (c) Predicted $u$. 
    (d) Predicted $f$.
    Blue circles: noisy measurements at the posterior estimation stage. Mean: predicted mean value; 2 std: predicted two standard deviations. 
    }
    \label{fig:fractional}
\end{figure}

We employ two generators $G^{(f)}_{\bm{\eta}_1}(x; \bm{\xi})$ and $G^{(\alpha)}_{\bm{\eta}_2}(\bm{\xi})$; the first one takes  Gaussian noise $\bm{\xi}$ and spatial coordinate $x$ as input and outputs $f(x)$; the second one takes the same noise $\bm{\xi}$ as input and outputs $\alpha$. For the DeepONet $\tilde{S}$, the input is a concatenation of $\alpha$ and the vector that resolves $f$, and the output is $u$. 
To generate training data for GANs, we assume that 
\begin{equation}\label{eq:a_f_onet}
    \begin{split}
        \alpha &\in 1 + \frac{1}{1 + \exp(-a)}, ~ a \in \mathcal{N}(0, 1^2), \\
     f & \thicksim \mathcal{GP}(0, \mathcal{K}), ~ \mathcal{K} = \exp \left( - \frac{(x - x')^2}{2 l^2}\right),\\
    x,&~ x' \in [-1, 1], ~l = 0.2.
    \end{split}
\end{equation}
% \begin{align}\label{eq:a_f_onet}
%     \alpha &\in 1 + \frac{1}{1 + \exp(-a)}, ~ a \in \mathcal{N}(0, 1^2), \\
%      f & \thicksim \mathcal{GP}(0, \mathcal{K}), ~ \mathcal{K} = \exp \left( - \frac{(x - x')^2}{2 l^2}\right),\\
%     x,&~ x' \in [-1, 1], ~l = 0.2.
% \end{align}
where $\mathcal{GP}$ represents a Gaussian process with kernel $\mathcal{K}$. We then randomly draw 10,000 samples for both $\alpha$ and $f$ in \Eqref{eq:a_f_onet}, and 50 equidistant points are utilized to resolve $f$. As for DeepONet, we solve \Eqref{eq:fraceq} using the the spectral method that employs Jacobi functions~\cite{MaoKar2018SINUM} based on the same $\alpha$ and $f$ for training GANs. The obtained pair data $(\alpha, f, u)$ are utilized in the training of DeepONet. Note that GANs and DeepONet are trained separately. Upon completion of training the GAN as well as the DeepONet, we can obtain $\tilde{S}[G^{(f)}_{\bm{\eta}_1}(\cdot; \bm{\xi}), G^{(\alpha)}_{\bm{\eta}_2}(\bm{\xi})](\vx)$, a physics-informed surrogate model with functional priors for $u$.

% We employ two generators $G^{(\alpha)}_{\bm{\eta}_2}$ and $G^{(f)}_{\bm{\eta}_1}(x; \bm{\xi})$; the first one takes the Gaussian noise $\bm{\xi}$ as input and outputs $\alpha$, and the second one takes the same $\bm{\xi}$ and spatial coordinate $x$ as input and outputs $f(x)$. While for the DeepONet, the input is the concatenation of $\alpha$ and the vector that resolves $f$, and the output is $u$. 

In the stage of posterior estimation, we now assume that we have 4 and 10 random noisy measurements for $u$ and $f$, respectively. In addition, the noise scales for $u$ and $f$ are the same, i.e., $\mathcal{N}(0, 0.05^2)$. The objective is first to infer $u$ and $f$ in the whole domain, i.e., $x \in [-1, 1]$, and second to estimate the unknown fractional order $\alpha$, with uncertainties.

The predicted $u$ and $f$ based on the learned functional priors are displayed in Figs. \ref{fig:fractionalc}-\ref{fig:fractionald}. We see that the computational errors are bounded by the predicted two standard deviations. Moreover, the exact and predicted $\alpha$ are 1.4523 and  $1.5187 \pm 0.1333$ (mean $\pm$ one standard deviation), respectively. 

% and (2) the predicted uncertainty becomes larger in the regions without measurements for both $u$ and $f$.
% We would like to discuss that it is difficult to obtain measurements on the fractional order $\alpha$ in real applications. To demonstrate the capability of the present method for estimating the unknown parameters for physical problems, we assume that we have measurements on $\alpha$ to get its prior distribution. A more practical setup for this problem is to directly map $f$ to $u$ using the DeepONet since (1) the chemical reaction which is generally related to the pore structure is difficult to formulate, suggesting the governing equation for the diffusion-reaction system is agnostic, and (2) measurements on both the source term and concentration field can be obtained in applications.  

% \input{2Dflow}

\subsection{DeepONet: 2D Flow in heterogeneous porous media}
\label{sec:2dflow}
% Estimation of hydraulic conductivity fields given limited measurements is crucial for reactive transport in porous media, e.g., contaminant transport in soil. In geological applications, direct measurements on hydraulic conductivity fields are quite expensive. Meanwhile, indirect measurements on hydraulic heads are much cheaper to obtain. A widely used approach is to bridge the direct and indirect measurements using the governing equation and solve an inverse problem to predict the hydraulic conductivity field  \cite{meng2020composite,zheng2020physics}. 

Here we aim to estimate the 2D hydraulic conductivity field based on partial observations of the hydraulic conductivity fields and hydraulic heads. In particular, DeepONet is utilized to inter-relate the hydraulic conductivity fields and hydraulic heads. 
We consider the following two-dimensional flow through heterogeneous porous media, which is governed by the following equation \cite{zheng2020physics}:
\begin{align}\label{eq:flow_2d}
    \nabla \cdot \left( K(\bm{x}) \nabla h(\bm{x})\right) = 0, ~ \bm{x} = (x, y), ~x, y \in [0, 1],
\end{align}
with boundary conditions
\begin{equation}
\begin{split}
    h(0, y) &= 1, ~ h(1, y) = 0, \\
    \partial_{\bm{n}} h(x, 0) &= \partial_{\bm{n}} h(x, 1) = 0,
\end{split}
\end{equation}
where $K$ is the hydraulic conductivity, and $h$ is the hydraulic head. Generally, $K$ is determined by the pore structure. To take different structures into consideration, we can then use a stochastic process to describe $K$ \cite{zheng2020physics}. Here, we apply the following model to describe $K$, which is widely used to mimic the real conductivity field \cite{zheng2020physics}, i.e., $K = \exp(F(\bm{x}))$, with $F(\bm{x})$ denoting a truncated Karhunen-Lo\`eve (KL) expansion for a certain Gaussian process. In particular, we keep the leading 100 terms in the KL expansion for the Gaussian process with zero mean and the following kernel:
\begin{equation}
\begin{split}
    \mathcal{K}(\bm{x}, \bm{x}') &= \exp\left[\frac{-(x - x')}{2l^2_1} + \frac{-(y - y')^2} {2 l^2_2} \right],  \\
    x, ~x' &\in [0, 1], ~y, ~y' \in [0, 1], ~l_1 = l_2 = 0.25.
\end{split}
\end{equation}

In the generator $G^{(K)}_{\bm{\eta}}(\bm{x}; \bm{\xi})$,  100-dimensional Gaussian noise and $\bm{x}$ serve as the input, and the output is the hydraulic conductivity $K(\bm{x})$. For the DeepONet, the function input is $K(\bm{x})$ and the output is the hydraulic head $h(\bm{x})$.
For the training of GAN, we randomly draw 30,000 samples from the truncated KL expansion for $K(\bm{x})$ \cite{zheng2020physics}, and a $20 \times 20$ uniform grid is employed to resolve $K(\bm{x})$.  As for the DeepONet, we utilize the \emph{finite-element-based Partial Differential Equation Toolbox} in Matlab to solve \Eqref{eq:flow_2d} using the same $K(\bm{x})$ for training GAN. Subsequently, the obtained pair data, i.e.,  $(K, h)$ are utilized for the training of DeepONet.  Upon completion of training GANs as well as DeepONet, we can obtain $\tilde{S}[G^{(K)}_{\bm{\eta}}(\cdot; \bm{\xi})](\vx)$, a physics-informed surrogate model as functional priors for $h$.

\begin{figure}[H]
    \centering
    \subfigure[]{\label{fig:onet_2da}
    \includegraphics[width = 0.9\textwidth]{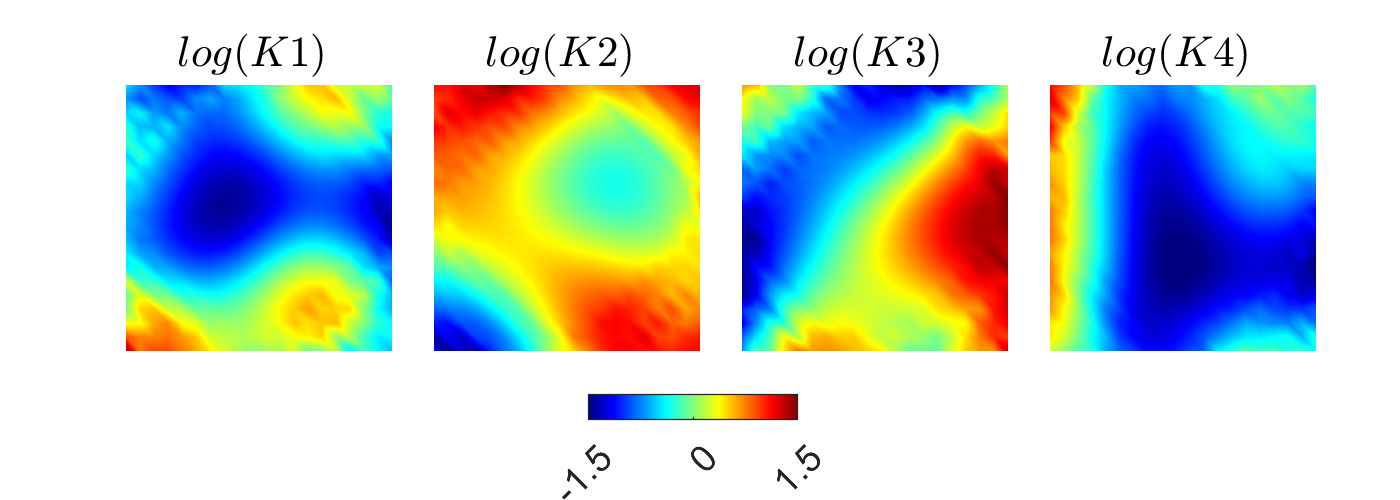}}
    \subfigure[]{\label{fig:onet_2db}
    \includegraphics[width = 0.9\textwidth]{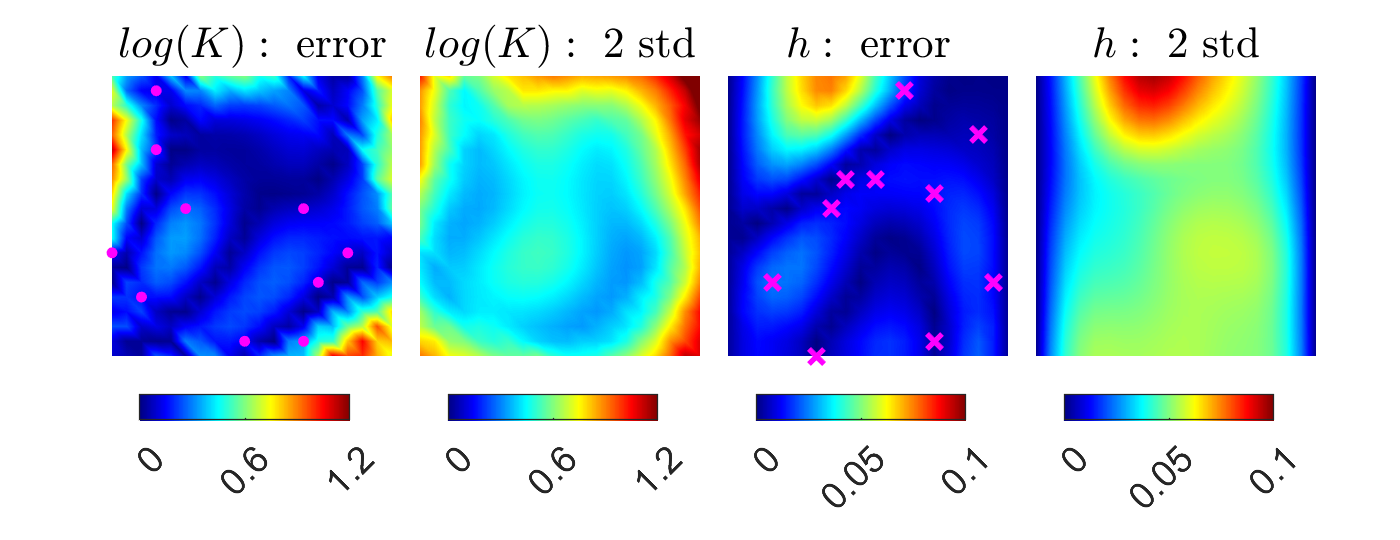}}
    \caption{
    DeepONet for 2D flow in heterogeneous porous media with uncertainty quantification: $\nabla \cdot \left( K(\bm{x}) \nabla h(\bm{x})\right) = 0$.
    (a) Four representative samples for the conductivity field. 
    (b) Predicted $K$ and $h$ with uncertainties.
    error: difference between the predictive mean and the ground truth.
    % error: $|\psi - \psi^*|$, where $\psi$ represent predicted means of any variables (e.g., $\log(K)$, $h$, etc); $\psi^*$ denotes the exact value; 
    std: standard deviation. Magenta dots: training data for $K$ at the posterior estimation stage; Magenta cross: training data for $h$ at the posterior estimation stage. 
    }\label{fig:onet_2d}
\end{figure}

% On the completion of training GANs and DeepONet, the physics-informed surrogate model for $h = \tilde{S}[G^{(K)}_{\bm{\eta}}(\cdot; \bm{\xi})](\vx)$ can then be obtained as functional priors. 

In the stage of posterior estimation, we assume that we have 10 random noisy measurements for both $K$ and $h$ (Fig. \ref{fig:onet_2d}), and the noise scales for both $\log(K)$ and $h$ are the same, i.e., $\mathcal{N}(0, 0.05^2)$. Note that noise is added to the $\log(K)$ field rather than $K$ here. The objective is to infer $K$ and $h$ in the whole field with uncertainties. We then present the predicted $K$ and $h$ based on the learned function priors in Fig. \ref{fig:onet_2d}. We can see that, first, the computational errors between the predicted means and exact solutions for both $K$ and $h$ are bounded by two standard deviations, and, second, the predicted uncertainties increase for locations with no measurements, as expected.

Finally, we would like to point out that  while the truncated KL expansion for a Gaussian process is employed to mimic the real conductivity field here, the present method can be readily applied to learn the non-Gaussian log conductivity field considered in \cite{meerschaert2013hydraulic,kang2019coupled}.

\subsection{Time- and space-dependent problem}

In this section, we test the proposed method for a time- and space-dependent problem. In particular, we use the data related to the experiments performed by the Norwegian Deepwater Programme (NDP) in the MARINTEK Offshore Basin on steel catenary risers (SCR) with high length-to-diameter ratio risers \cite{NDP_SCR}. The outer diameter of the riser $d = 14mm$, length of the riser $L = 12.5m$, incoming flow velocity $v = 0.12m/s$. The experimental data was collected by accelerometer sensors along the riser, in both the in-line (IL) and cross-fLow (CF) directions. With the data for acceleration, we reconstructed the displacements in the IL direction using a Fourier expansion. The data used in this section are the reconstructed displacements of the riser in the spatial-temporal domain, denoted as $u(x,t)$.

\begin{figure}[H]
    \centering
    \subfigure[]{
    \includegraphics[width=0.8\textwidth]{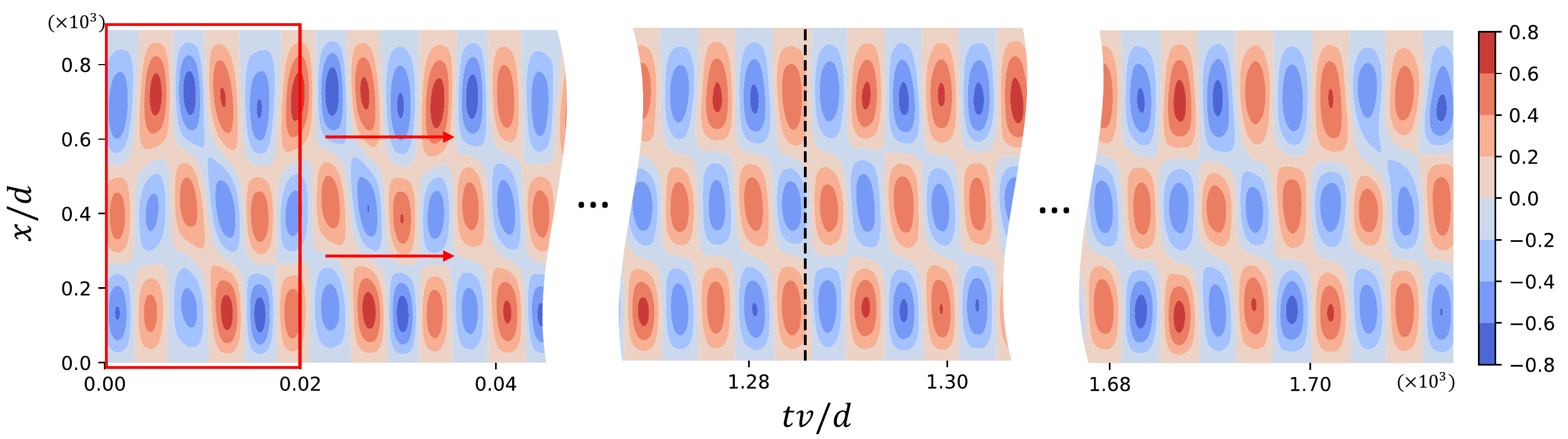}
    \label{fig:riserdisplacement}}
    \subfigure[]{
    \includegraphics[width=0.23\textwidth]{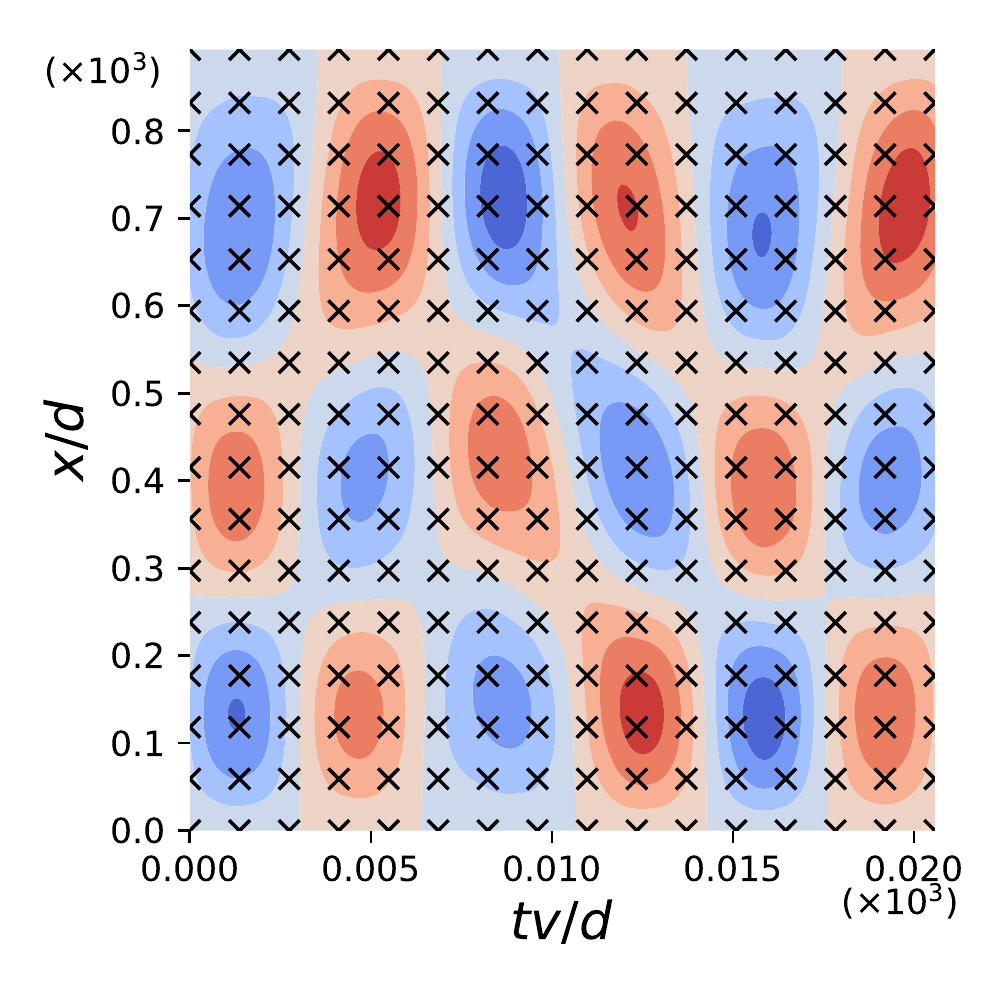}
    \includegraphics[width=0.23\textwidth]{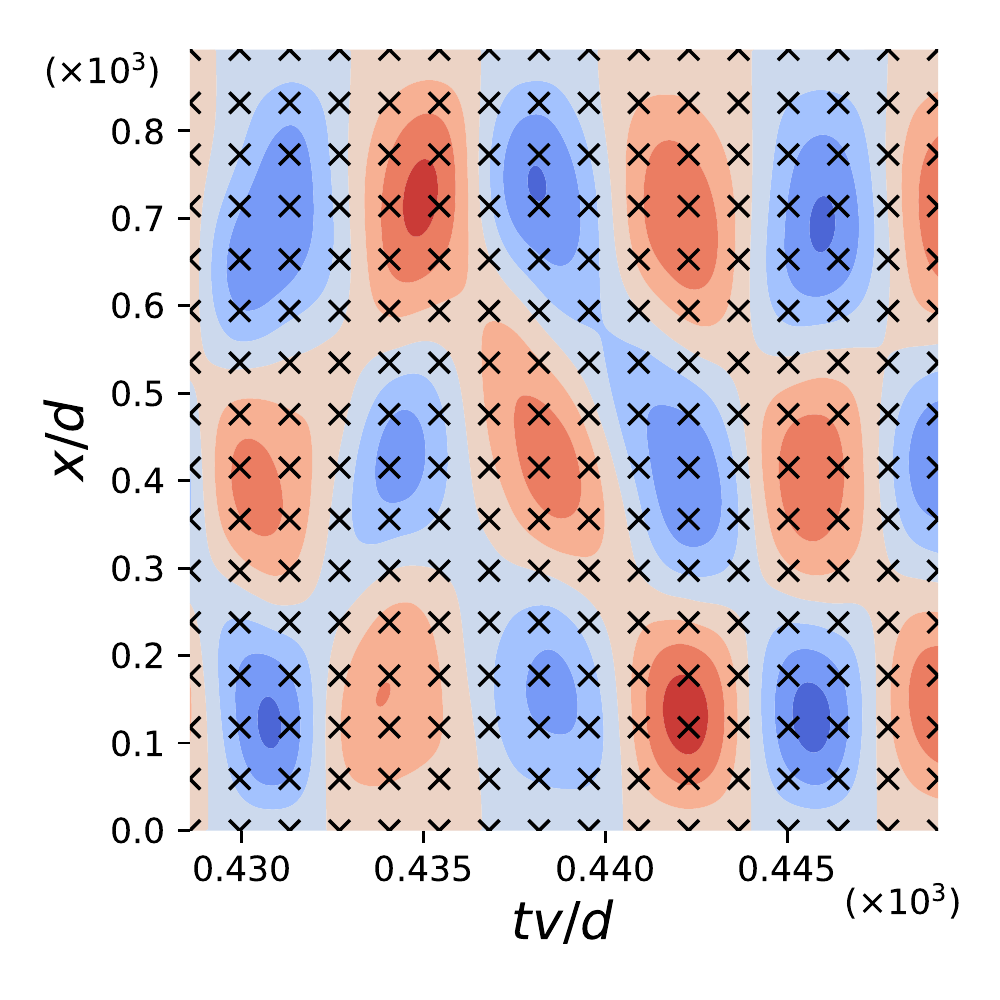}
    \includegraphics[width=0.23\textwidth]{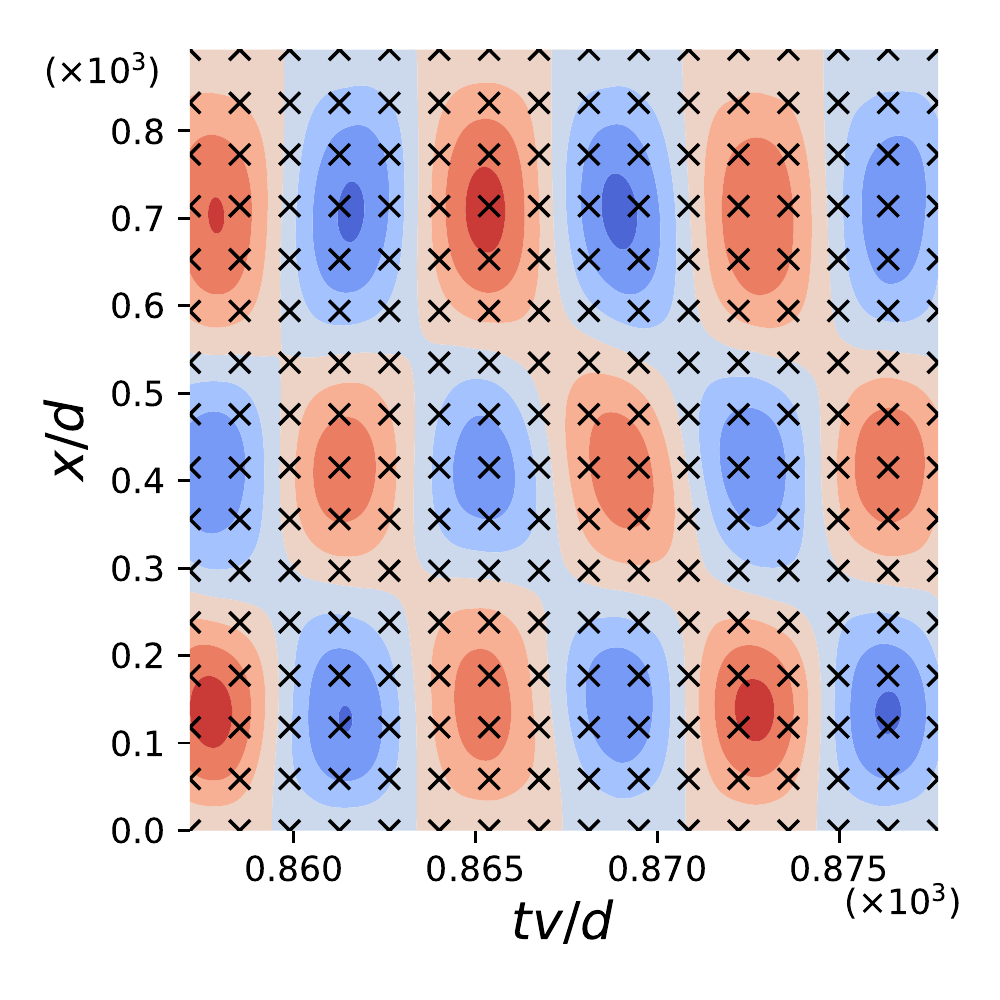}
    \includegraphics[width=0.23\textwidth]{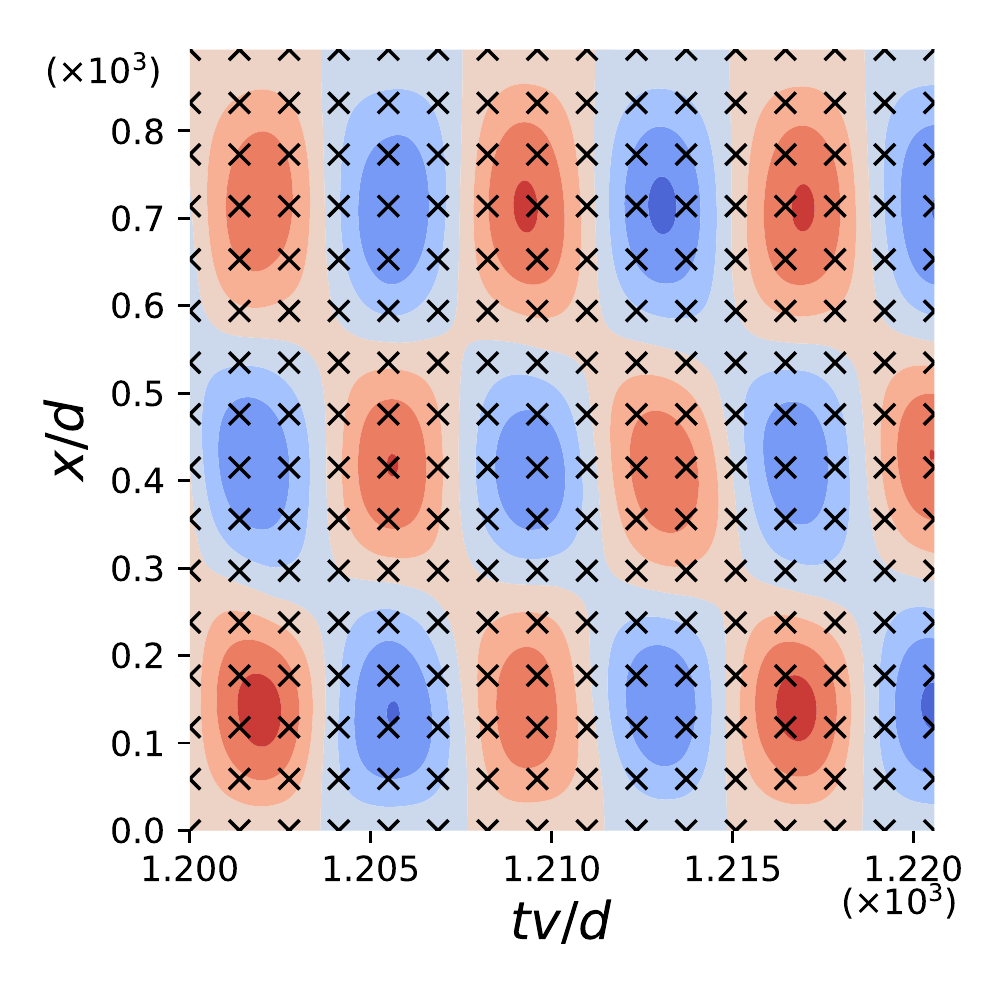}
    \label{fig:risersnaps}
    }
    \caption{
    Data for learning the time- and space-dependent functional priors. 
    (a) Dimensionless  displacement $u(x,t)/d$. The vertical dashed line represents the split between the training dataset (for learning the functional prior) and the test dataset (for posterior estimation). The red rectangular shows the sliding window.
    (b) Four examples of snapshots generated by the sliding window. The black crosses represent the historical data for learning the functional prior.
    }
    \label{fig:riserprior}
\end{figure}

We use the first 3/4 of $u(x,t)/d$ to learn the functional priors, and test on the rest 1/4. The dimensionless displacement $u(x,t)/d$ and the split are visualized in Fig.~\ref{fig:riserdisplacement}. To mimic the experiments, let us assume that we place 16 sensors to read the noiseless displacement with frequency $100$Hz. We can then use a sliding window to generate the training snapshots with a sliding step  $\Delta t = 0.01s$, i.e., $\Delta tv/d \approx 0.0857$. The window covers all the spatial domain and covers $t_w = 2.4s$ (i.e., $t_w v/d \approx 20.6$) in the temporal domain, which is close to three vortex shedding periods of the flow around the long marine riser. Since we read the sensors with frequency $100$Hz, in the window we have 241 reads for each sensor. However, for this problem fewer reads are sufficient to resolve $u(x,t)$ in the window. Therefore, in each window we equidistantly select 16 out of 241 reads to generate one training snapshot, i.e., each snapshot consists of $16 \times 16 = 256$ data points since 16 sensors are used. The total number of snapshots to learn the prior is $14,760$, and we visualize four examples of the snapshots in Fig.~\ref{fig:risersnaps}.

\begin{figure}[H]
    \centering
    \subfigure[]{
    \includegraphics[width=0.4\textwidth]{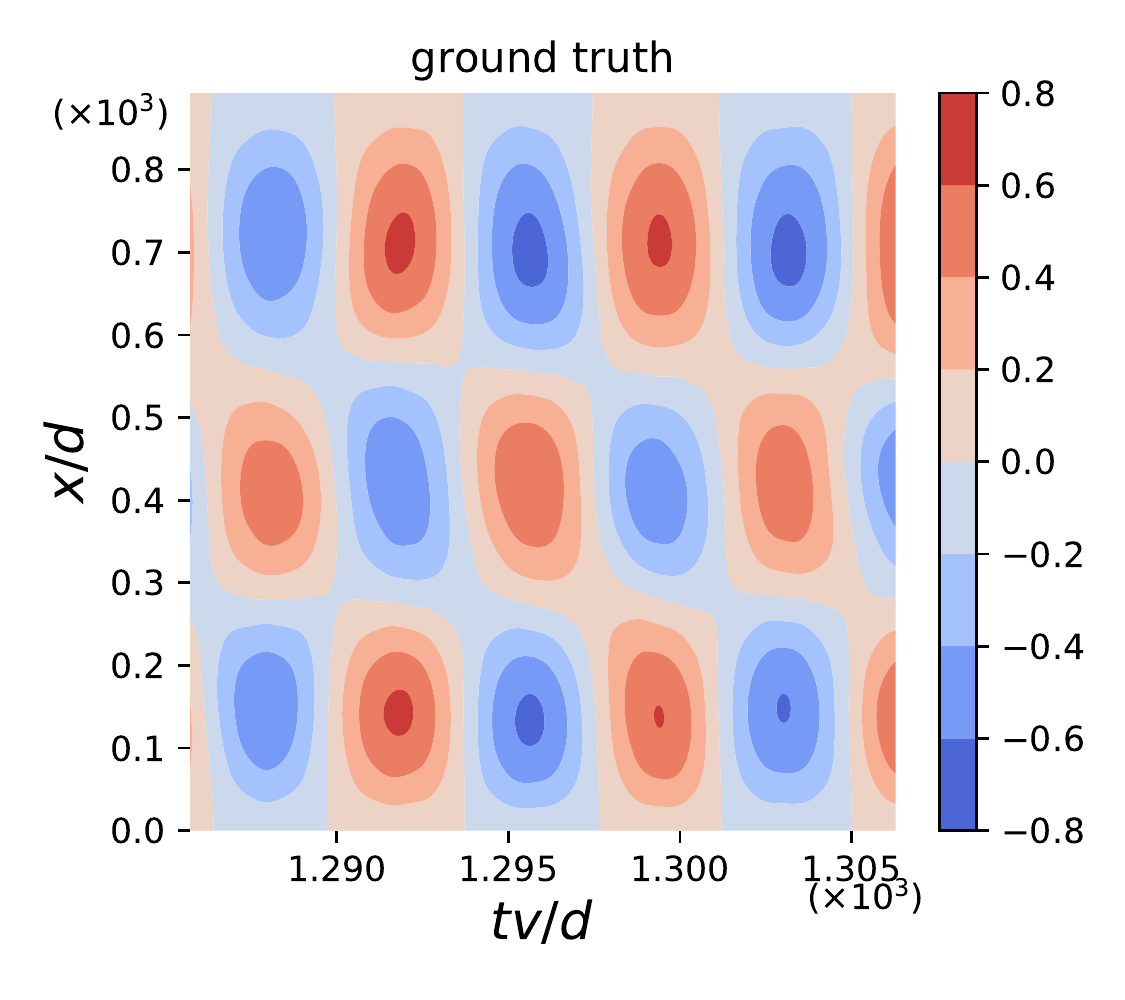}
    \label{fig:risergt}}
    % \subfigure[]{
    % \includegraphics[width=0.4\textwidth]{Figures/Riser/1000CF-1x6-gt-15000.pdf}
    % \label{fig:risererror}}
    \subfigure[]{
    \includegraphics[width=0.3\textwidth]{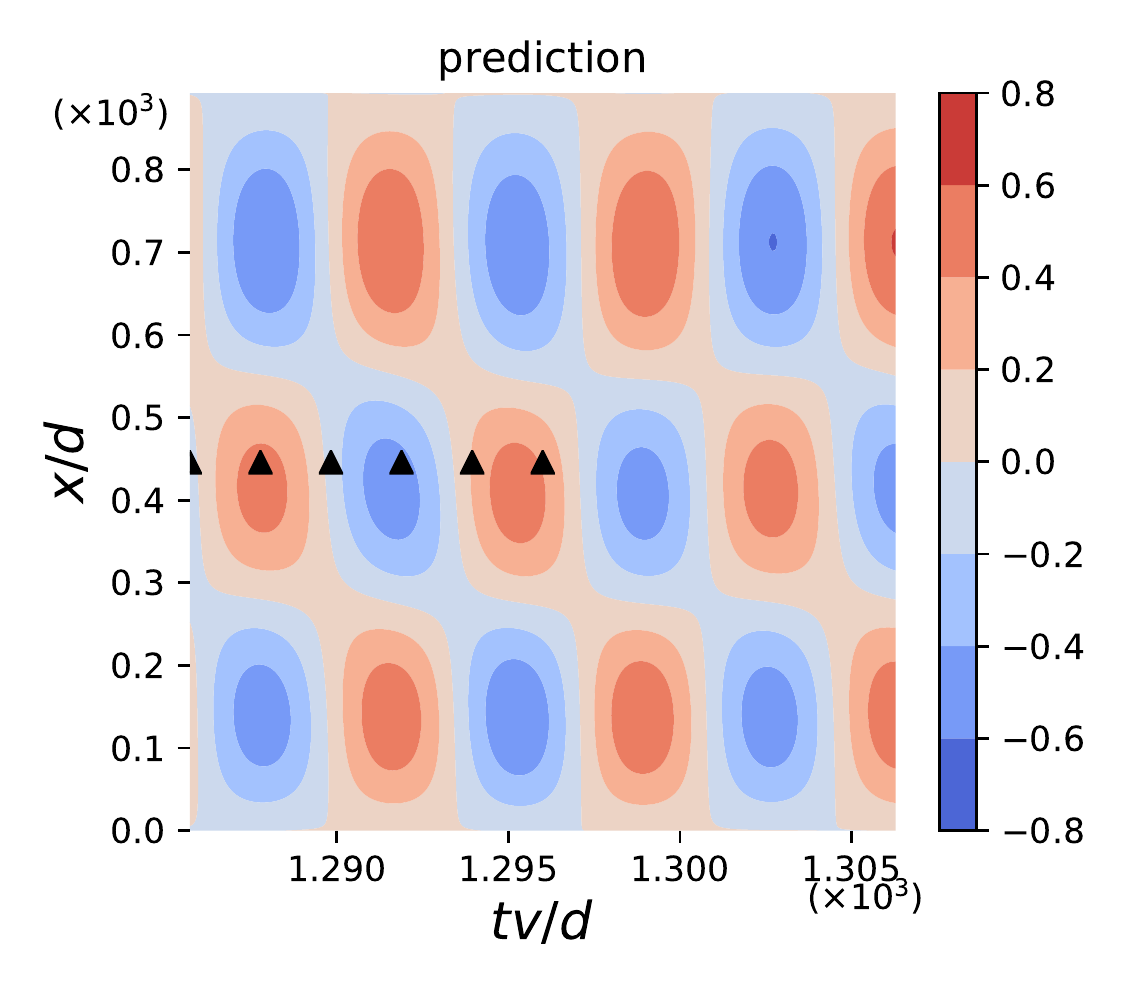}
    \includegraphics[width=0.3\textwidth]{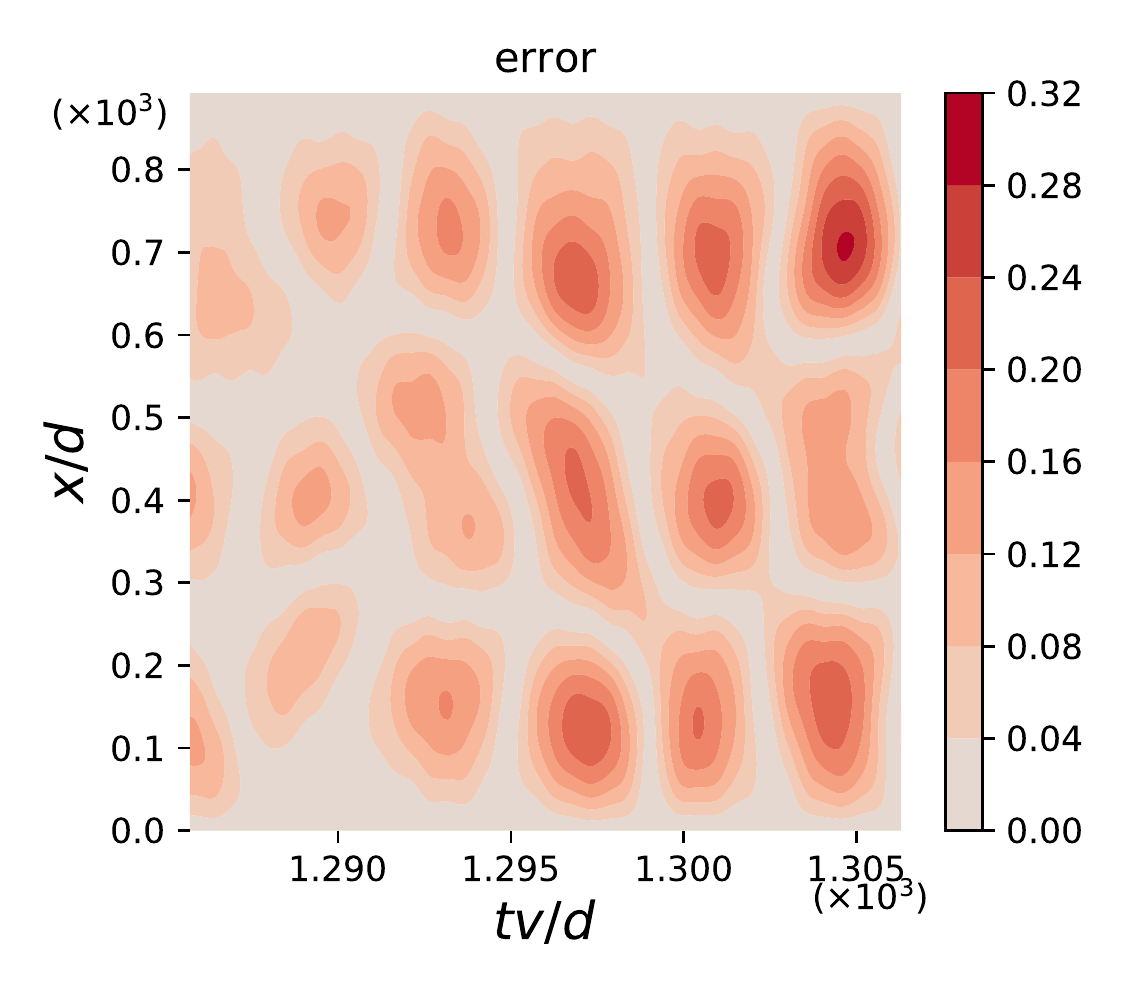}
    \includegraphics[width=0.3\textwidth]{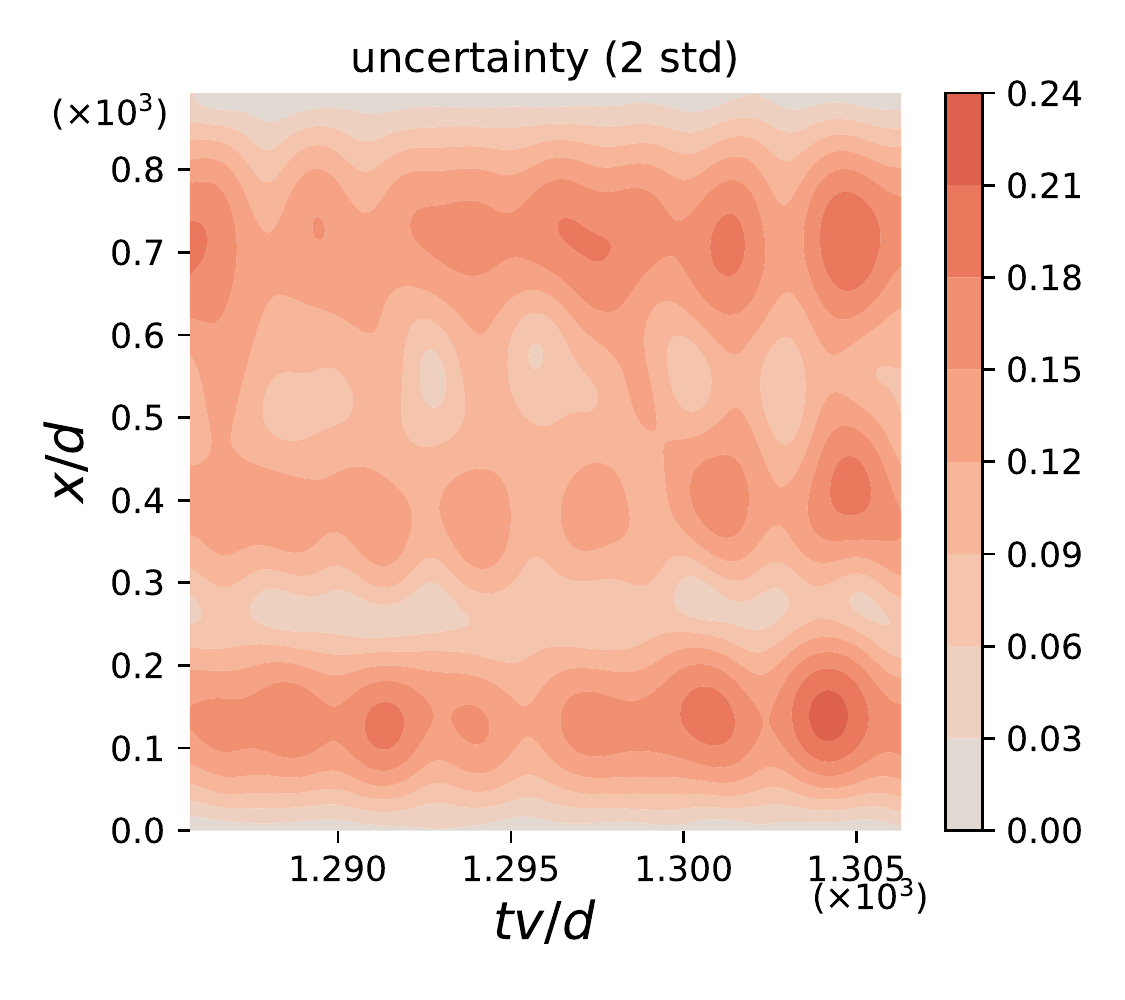}
    \label{fig:riser1x6}}
    \subfigure[]{
    \includegraphics[width=0.3\textwidth]{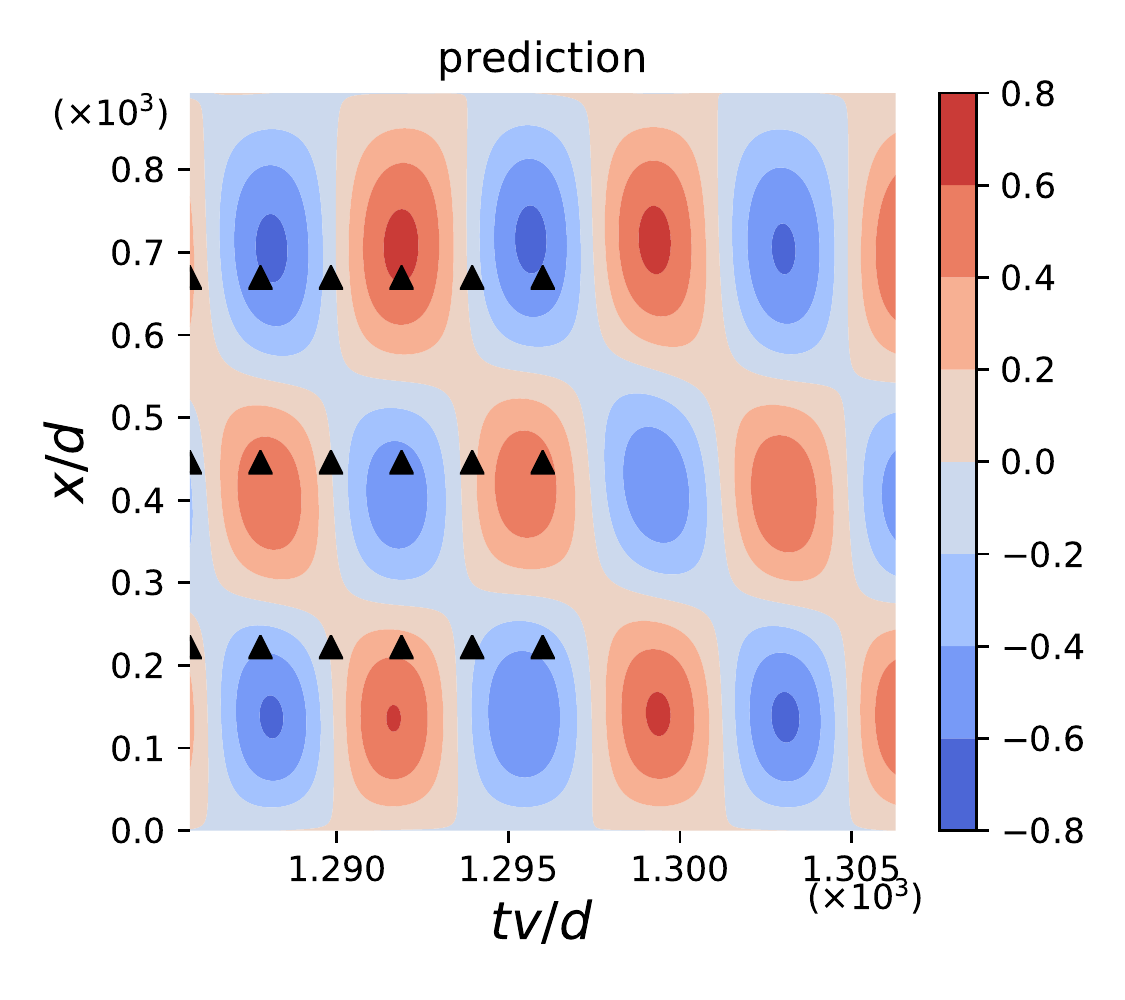}
    \includegraphics[width=0.3\textwidth]{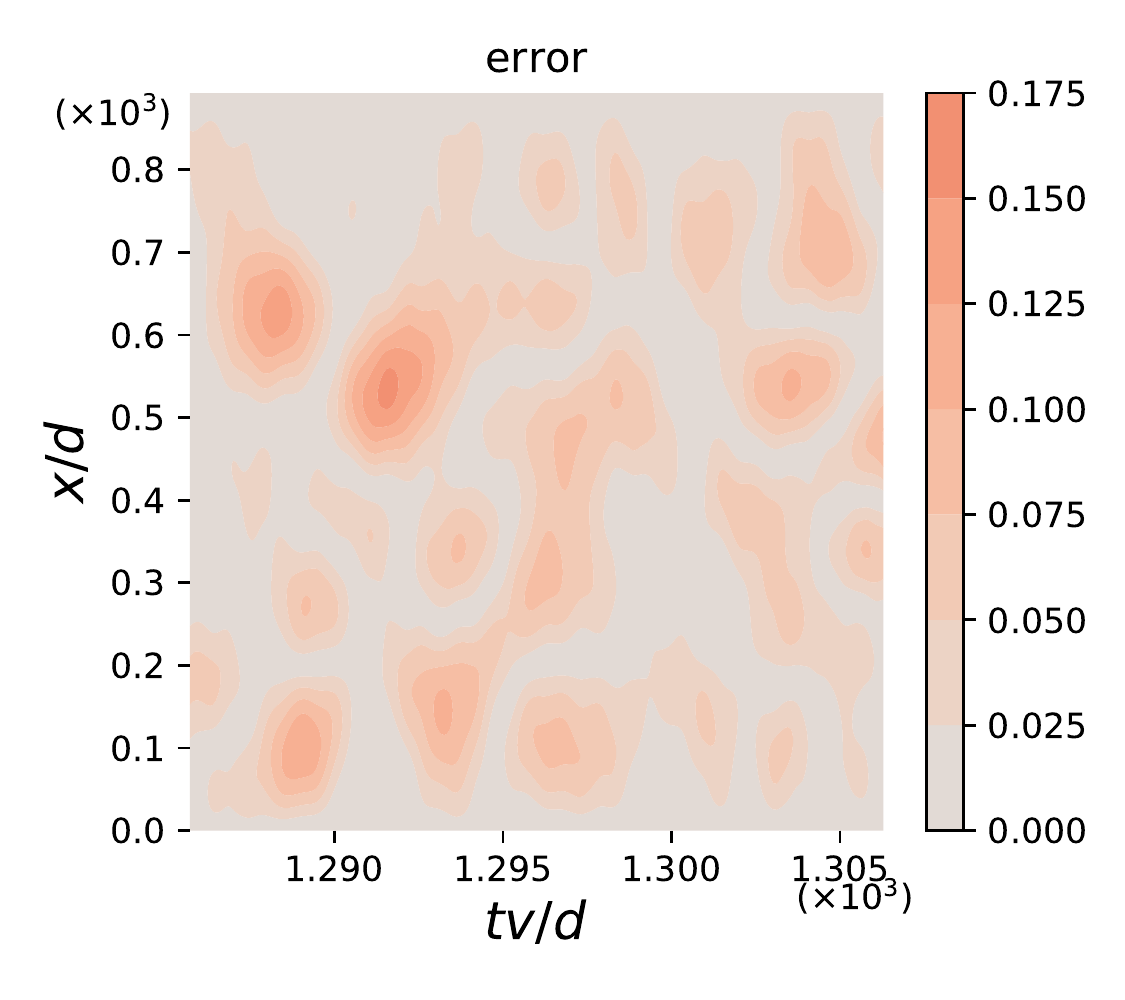}
    \includegraphics[width=0.3\textwidth]{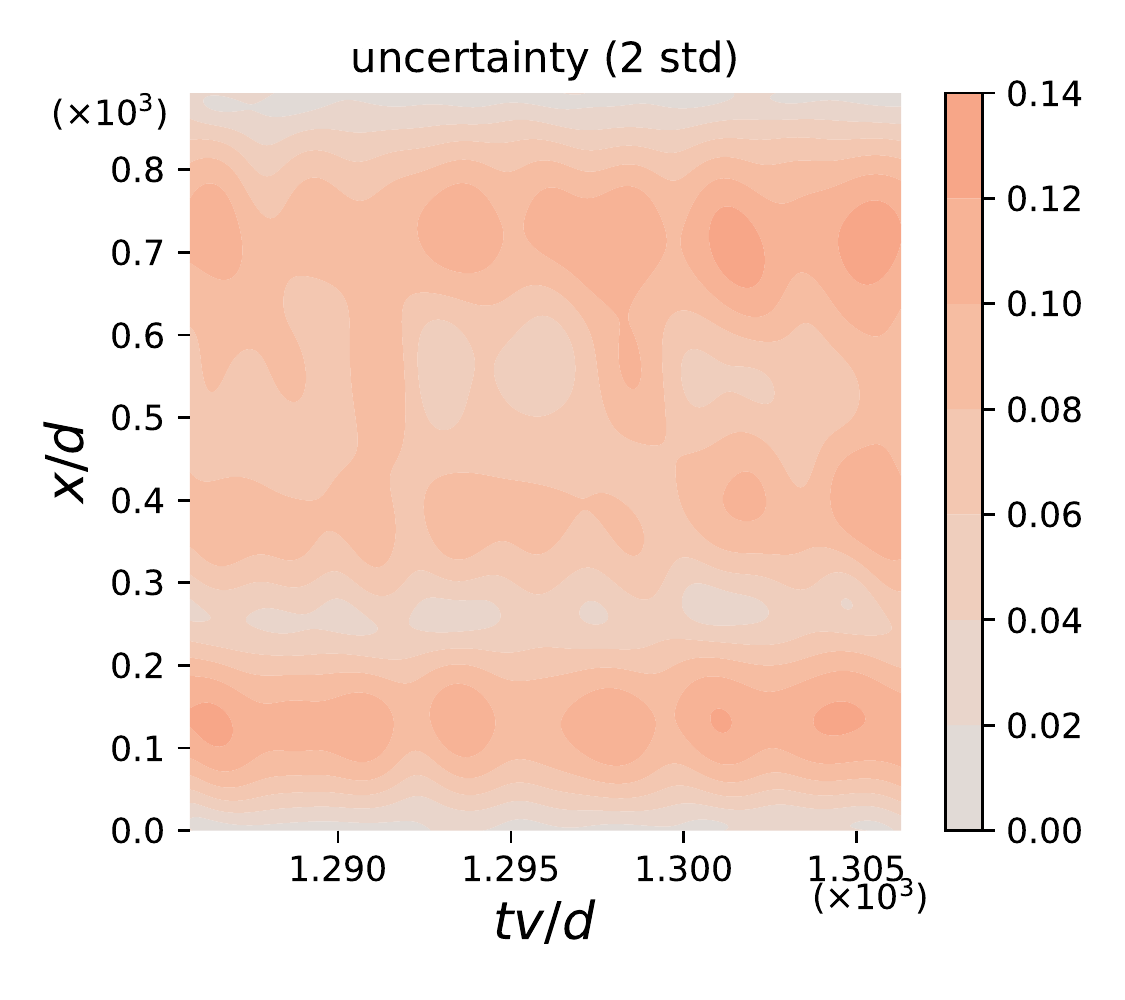}
    \label{fig:riser3x6}}
    % \caption{The predictive results using the learned functional prior. (a) The ground truth of the displacement in a time window. (b) The corresponding prediction, i.e., the mean of the posterior conditioned on six observations with noise size 0.1 (marked by the black crosses). (c) The uncertainty, i.e., the two standard deviations of the posterior. (d) The error of the prediction. For this case, the error is bounded by two standard deviations in about 88\% of the spatial-temporal domain, and bounded by one standard deviation in about 49\% of the spatial-temporal domain.}
    \caption{
    Predicted displacements in a window, using the learned functional prior. (a) The ground truth of the dimensionless displacement in a time window. (b-c) The corresponding prediction (mean of posterior), the error of the prediction, and the uncertainty (2 standard deviations of posterior), with 1 or 3 sensors.
    }
    \label{fig:riserpost}
\end{figure}

\begin{figure}[H]
    \centering
    \subfigure[]{
    \includegraphics[width=0.4\textwidth]{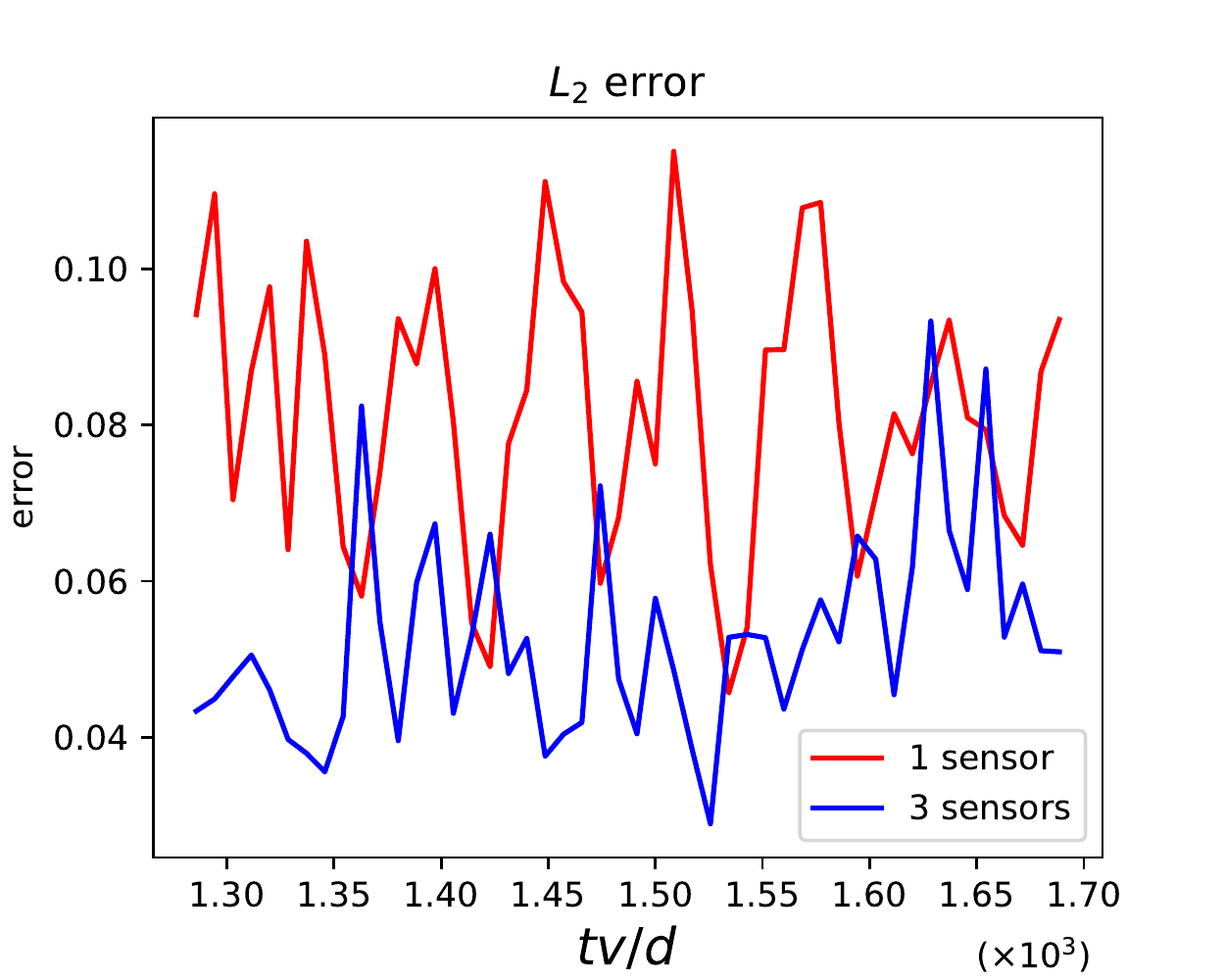}
    \label{fig:risererror}}
    \subfigure[]{
    \includegraphics[width=0.4\textwidth]{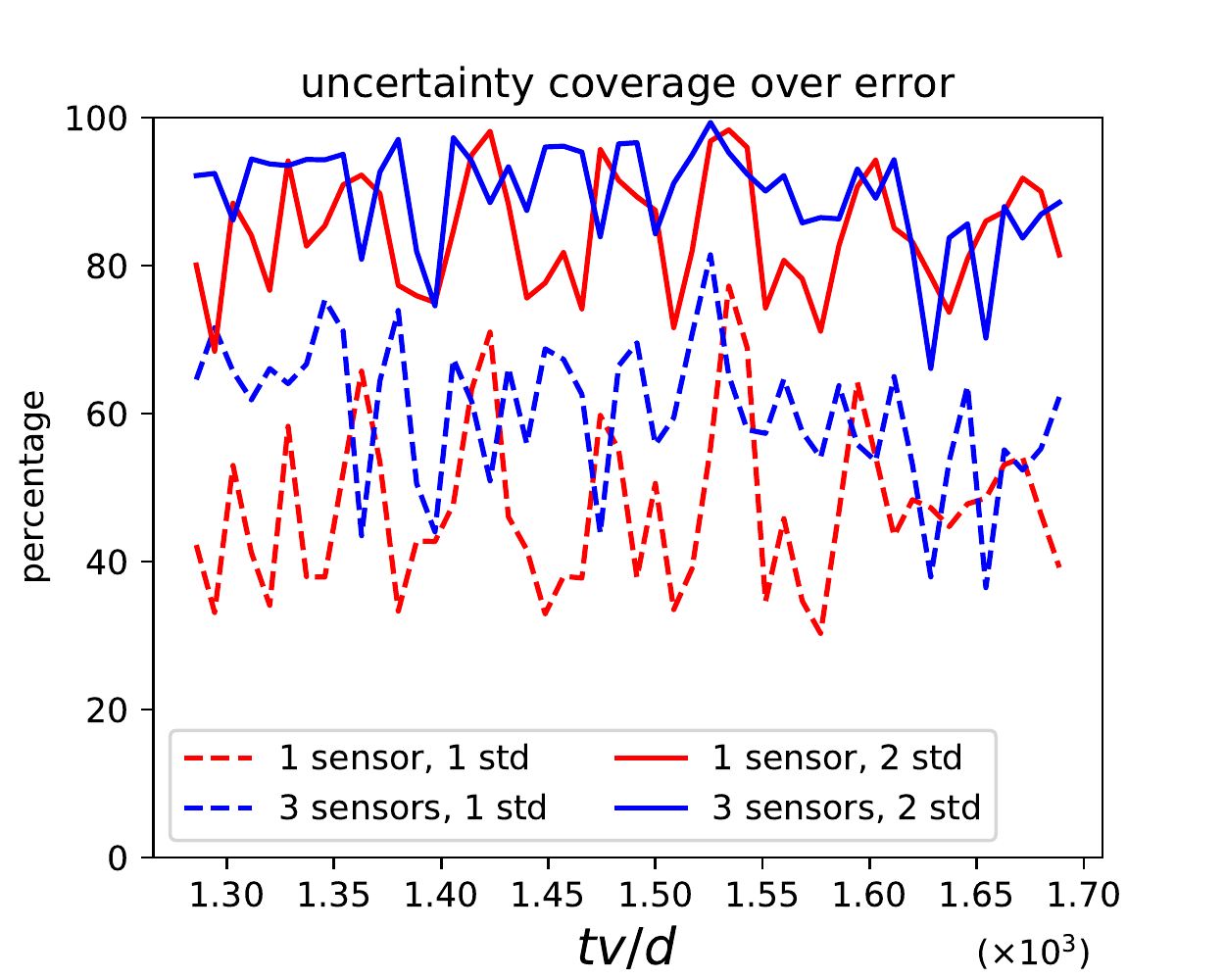}
    \label{fig:riseruqcover}}
    \caption{
    Predicted displacements for all the testing cases/windows using the learned functional prior. (a) $L_2$ error in each window. (b) Percentage of the area where the error is bounded by one or two standard deviations, in each window.
    }
    \label{fig:riserpost2}
\end{figure}

In the test stage, we still work in the sliding windows with the same size as in the training stage. However, we assume that we have less sensors placed in the spatial domain. In particular, we test with two scenarios. In the first scenario, we assume that we place one sensor to collect new data, while in the second scenario, we assume that we place three sensors to collect new data. For each scenario, we test 48 cases: the starting time of the sliding window is $t = 150, 151, 152,...,197s$, respectively, i.e., $tv/d = 1.286, 1.294,...,1.689$. For each case, we read $u(x,t)/d$ 6 times on each sensor, with additional noise drawn from $\mathcal{N}(0,0.1^2)$. Two examples of the sensor placement and corresponding results are illustrated in Fig.~\ref{fig:riser1x6} and \ref{fig:riser3x6} with ground truth in Fig.~\ref{fig:risergt}. The $L_2$ error and the uncertainty coverage over error for all the cases are illustrated in Fig.~\ref{fig:riserpost2}. Here, we can see that in general the error is reduced if we increase the number of sensors from 1 to 3. Also, in general the error is bounded by two standard deviations in most of the area, and bounded by one standard deviations in about half of the area. Such results show that the uncertainty we predict is reasonable.

\section{Summary}
\label{sec:summary}

We have developed a novel method to learn functional priors and posteriors from data and physics using historical data. In particular, we use PI-GANs with PINNs or DeepONets to learn functional priors from historical data and physics. Such learned functional priors are superior to the artificially designed ones, as they are more flexible and more informative. We started from a pedagogical example to show that our method, which leverages the knowledge of historical data in the Bayesian framework, is better than the standard meta-learning algorithm MAML and standard Gaussian process regression. We then tested 1D and 2D forward and inverse PDE problems using PINNs or DeepONets to encode physics, as well as a real-world time- and space-dependent regression problem. In these problems, we show that the proposed method can give good predictions and reasonable uncertainty quantification, with relatively small number of sensors, which can be attributed to the informative functional priors that reflect our knowledge from historical data. Specially, the present method is capable of handling problems in 100 dimensions (parameter space) as shown in Sec. \ref{sec:2dflow}, which makes it a promising tool for quantifying uncertainties in high-dimensional parametric PDEs.

In \ref{sec:comparison} we compared different neural network architectures for generators and found that for certain architectures, vanilla HMC could fail for posterior estimation with the prior learned by GANs. We present a detailed discussion in \ref{sec:whyfail}. Such issue suggests that further research is required to optimize the neural network architecture for the proposed method, and more specifically investigate what version of GANs is more suitable for such tasks.

\section*{Acknowledgement}
X. Meng, L. Yang, and G. E. Karniadakis would like to acknowledge the support of PhILMS grant DE-SC0019453, OSD/AFOSR MURI grant FA9550-20-1-0358, and the NIH grant U01 HL142518.

\appendix
\section{Model-Agnostic Meta-Learning}
\label{sec:maml}
Meta-learning is a machine learning paradigm that is commonly understood as {\emph{learning to learn}} \cite{hospedales2020meta}. In particular, a machine learning model gains prior knowledge over multiple learning episodes given training data, which often covers a distribution of related tasks, and then this learned prior knowledge is used to improve the model performance for a new task in the future \cite{hospedales2020meta}. Generally, meta-learning has two stages, i.e., the meta-training stage and the meta-testing stage (which corresponds to the prior learning and posterior estimation stage in the present study). The former is to gain prior knowledge from training data, which we refer to as historical data in this study; the latter is to make predictions for a new unseen task based on the learned prior knowledge as well as new training data. Among all meta-learning models, we focus on the {\emph{neural-network}} meta-learning, which is highly expressive due to the power of deep neural networks. Our particular interest is in the model agnostic meta-learning (MAML), which is one of the most popular and efficient neural-network meta-learning approaches and has been successfully used in many fields, such as regression, classification, etc. \cite{finn2017model}. 

In MAML, we have training data for a certain number of tasks, which are from the underlying task distribution $p(\mathcal{T})$. Such training data can be viewed as an analogue of the historical data in this paper. As for the training, MAML has an inner and outer optimization, as displayed in Algorithm \ref{alg:maml}. Upon completion of the meta-training stage, the learned hyperparameters will be used as the initialization at the meta-testing stage. 
% Specifically, we assume that we have $(K + K')$ data points for each task.

\begin{algorithm}[H]
\caption{Model Agnostic Meta-Learning}
\label{alg:maml}
\begin{algorithmic}
\Require: $p(\mathcal{T})$: distributions over tasks
\Require: $\alpha,~ \beta$: learning rate used in optimization
\State Randomly initialize hyperparameters $\bm{\theta}$ in neural networks
\While{$\bm{\theta}$ not converged}
    \State Sample batch of tasks $\mathcal{T}_i \sim p(\mathcal{T})$,\;
    \For{all $\mathcal{T}_i$}
    \State Sample $K$ data points $\mathcal{D}_i=\{(\bm{x}^{j}, \bm{y}^{j})\}_{j=1}^K$ from $\mathcal{T}_i$,\;
    \State Compute the MSE $\mathcal{L}_{\mathcal{T}_i}(\bm{\theta})$ using $\mathcal{D}_i$, \;
    \State Perform one step inner optimization: $\bm{\theta}'_{i} \gets \bm{\theta} - \alpha \nabla_{\bm{\theta}} \mathcal{L}_{\mathcal{T}_i}(\bm{\theta})$,\;
    \State Sample $K'$ data points $\mathcal{D}'_i=\{(\bm{x}'^{j}, \bm{y}'^{j})\}_{j=1}^{K'}$ from $\mathcal{T}_i$,\;
  \EndFor
\State Perform outer optimization: $ \bm{\theta} \gets \bm{\theta} - \beta \sum_{\mathcal{D}'_i}\nabla_{\bm{\theta}} \mathcal{L}_{\mathcal{T}_i}(\bm{\theta}'_i)$ using all $\mathcal{D}'_i$. \;
\EndWhile 
\State Use $\bm{\theta}$ as the initialization at the meta-testing stage.
\end{algorithmic}
\end{algorithm}

For the modeling in Sec. \ref{sec:meta_func}, the employed fully-connected neural network has 2 hidden layers with 40 neurons per layer, the hyperbolic tangent function is used as the activation function, and $\alpha = 0.01$, $\beta = 0.001$, which are similar as in \cite{finn2017model}. At the meta-training stage, we assume that we have the same number of training data for each task, i.e., 30 equidistant points. We randomly select $K = 20$ of them for the inner optimization, and the remaining are for the outer optimization. The number of meta-training steps is set as 100,000. As the meta-testing stage, we first employ the Adam optimizer until the loss is smaller than $10^{-3}$, and then switch to LBFGS-B optimization until convergence, i.e., the error between two adjacent steps is less than $10^{-8}$.

\section{Comparison of the Generator Architectures}\label{sec:comparison}

Here we conduct a comparison on the performance of generators with different structures using an example of one-dimensional regression. Specifically, a specified Gaussian process is utilized as the functional prior, i.e.,
\begin{equation}
\begin{split}\label{eq:gp_prior}
    \bm{u} & \thicksim \mathcal{GP}(0, \mathcal{K}), ~ \mathcal{K} = \exp \left( - \frac{(x - x')^2}{2 l^2}\right),\\
    x,&~ x' \in [-1, 1], ~l = 0.2. 
\end{split}
\end{equation}
We then assume that we have access to 10 random noisy measurements, the objective is to infer the target function for $x \in [-1, 1]$ with uncertainties using the prescribed prior as well as the measurements. Particularly, two specific architectures for the generators will be tested, i.e., the first one introduced in Sec. \ref{sec:architecture} of the main text (Gen I), and the second one is from \cite{yang2020physics} (Gen II).  The results from the GPR with the specified prior will serve as the reference.

For Gen I, we employ the same architecture for $\tilde{g}_{\bm{\eta}_1}$ and $\tilde{h}^*_{\bm{\eta}_2}$, i.e., 2 hidden layers with 64 neurons per layer. The hyperbolic tangent function is used as the activation function. As for Gen II, we also employ a DNN with 2 hidden layers with 64 neurons per layer and the same activation function, i.e., hyperbolic tangent function.  The discriminators for both test cases are kept the same, i.e., 3 hidden layers with 128 neurons per layer,  the activation function for which is leaky ReLu function. In addition, the dimensions for the input noise $\bm{\xi}$ is 40, which is kept the same in all tests. 

\begin{figure}[H]
    \centering
    \subfigure[]{\label{fig:gp_priora}
    \includegraphics[width=0.3\textwidth]{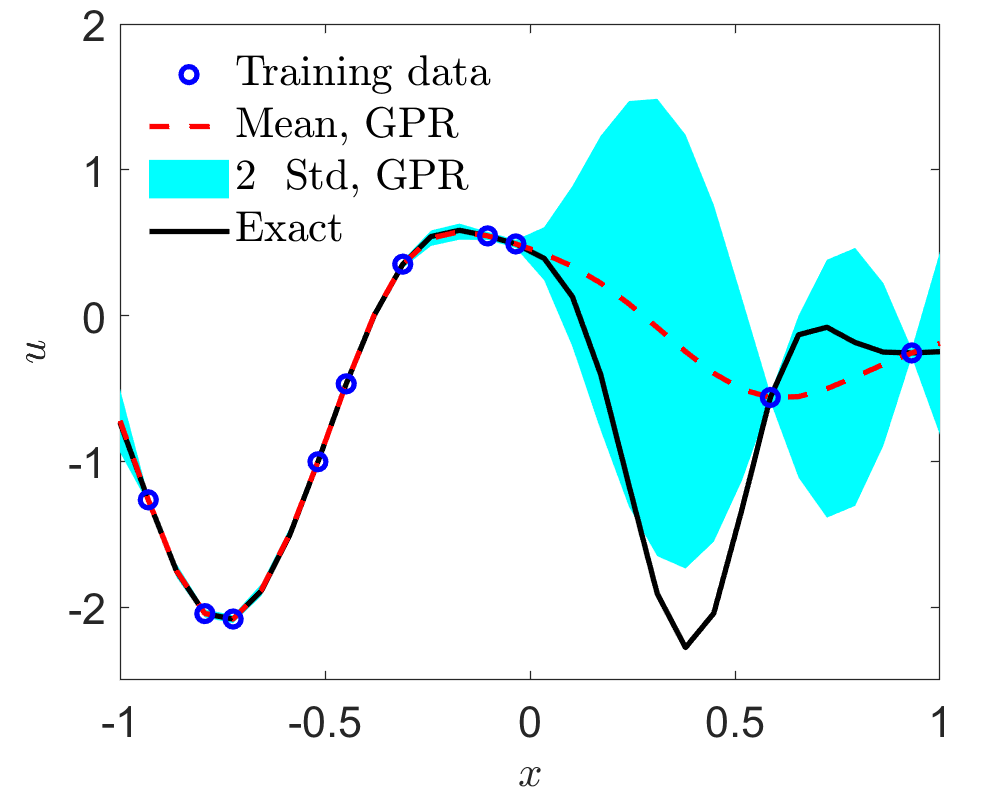}}
    \subfigure[]{\label{fig:gp_priorb}
    \includegraphics[width=0.3\textwidth]{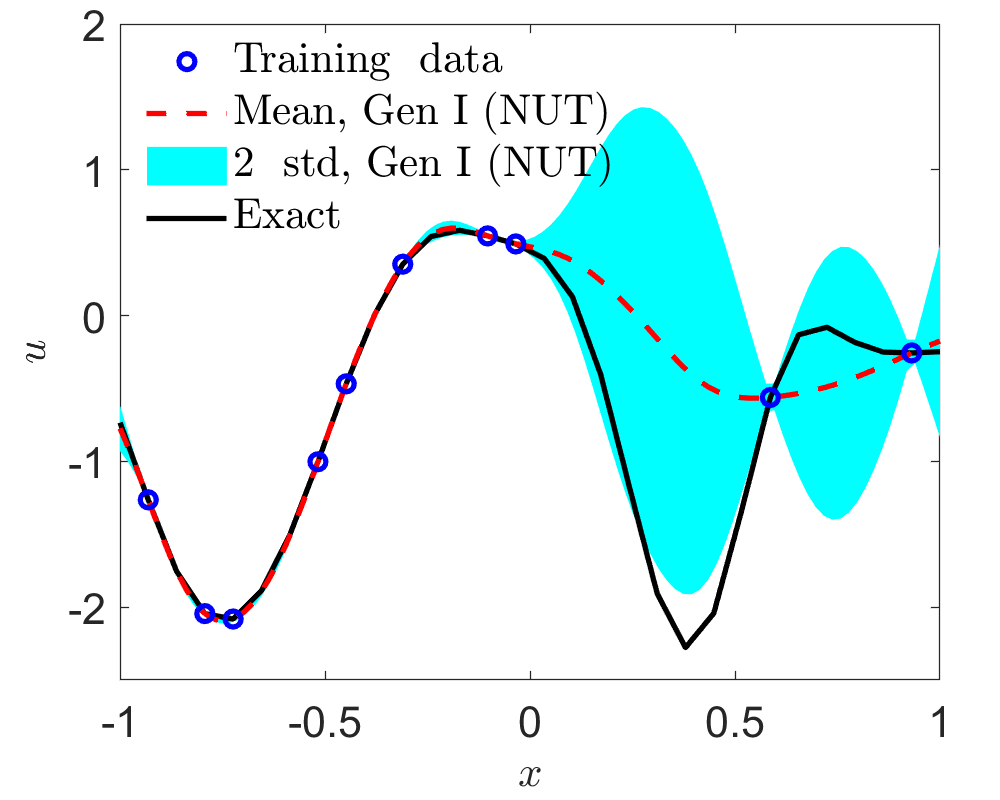}}
    \subfigure[]{\label{fig:gp_priorc}
    \includegraphics[width=0.3\textwidth]{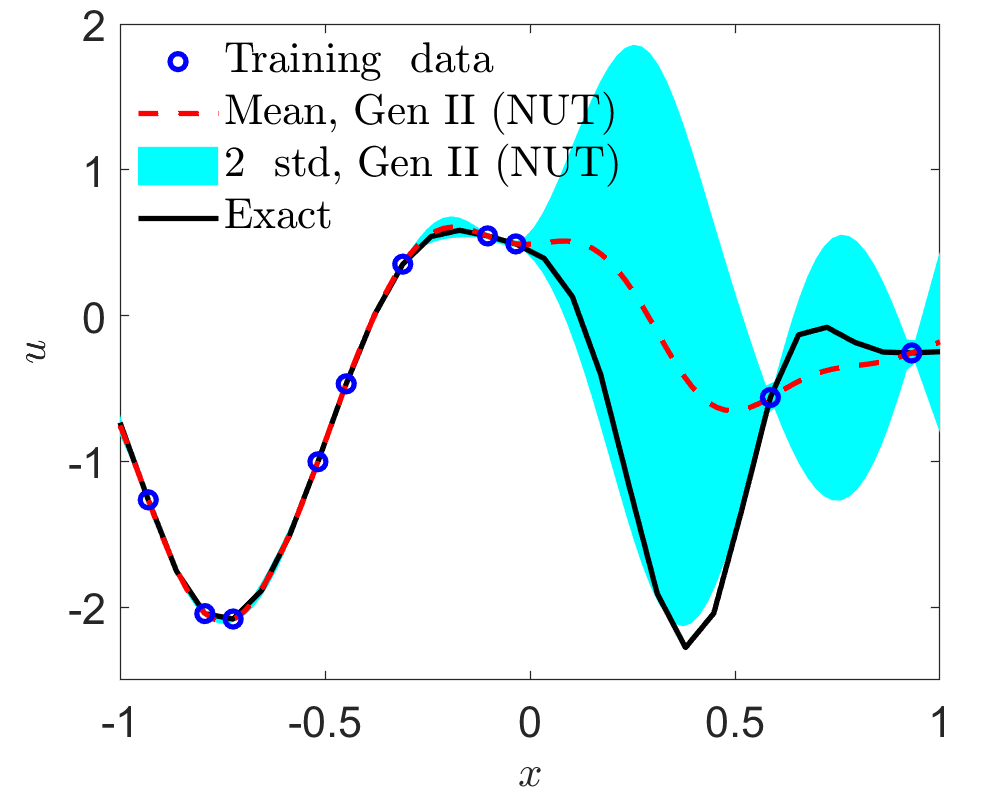}}
    \subfigure[]{\label{fig:gp_priord}
    \includegraphics[width=0.3\textwidth]{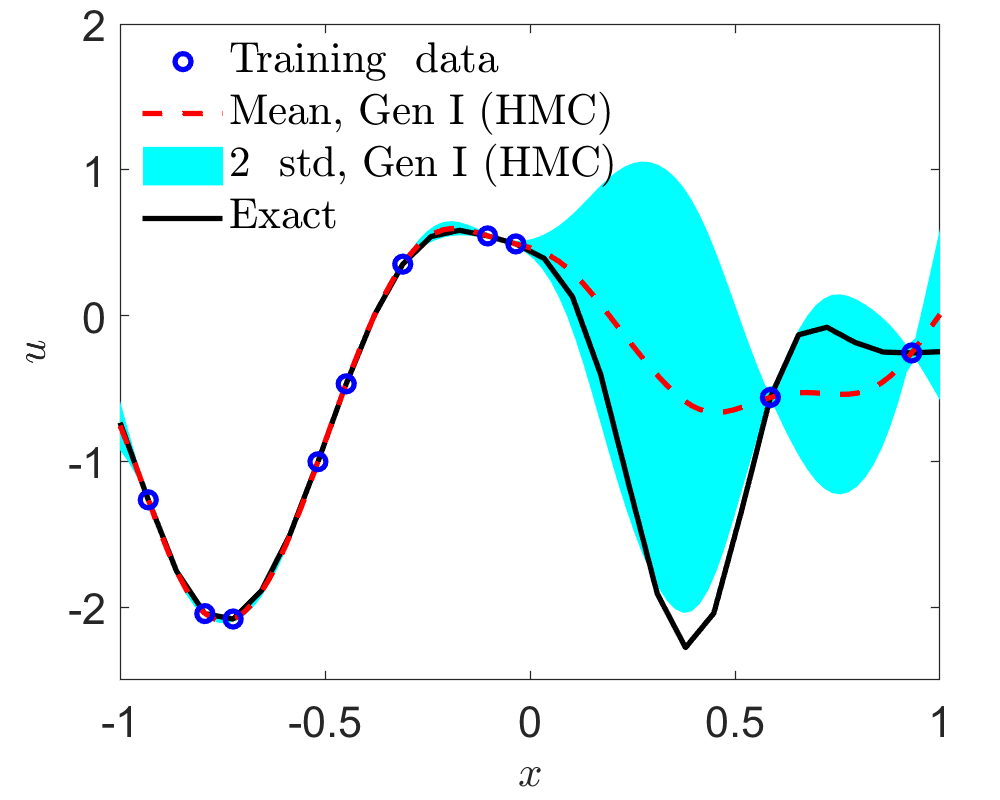}}
    \subfigure[]{\label{fig:gp_priore}
    \includegraphics[width=0.3\textwidth]{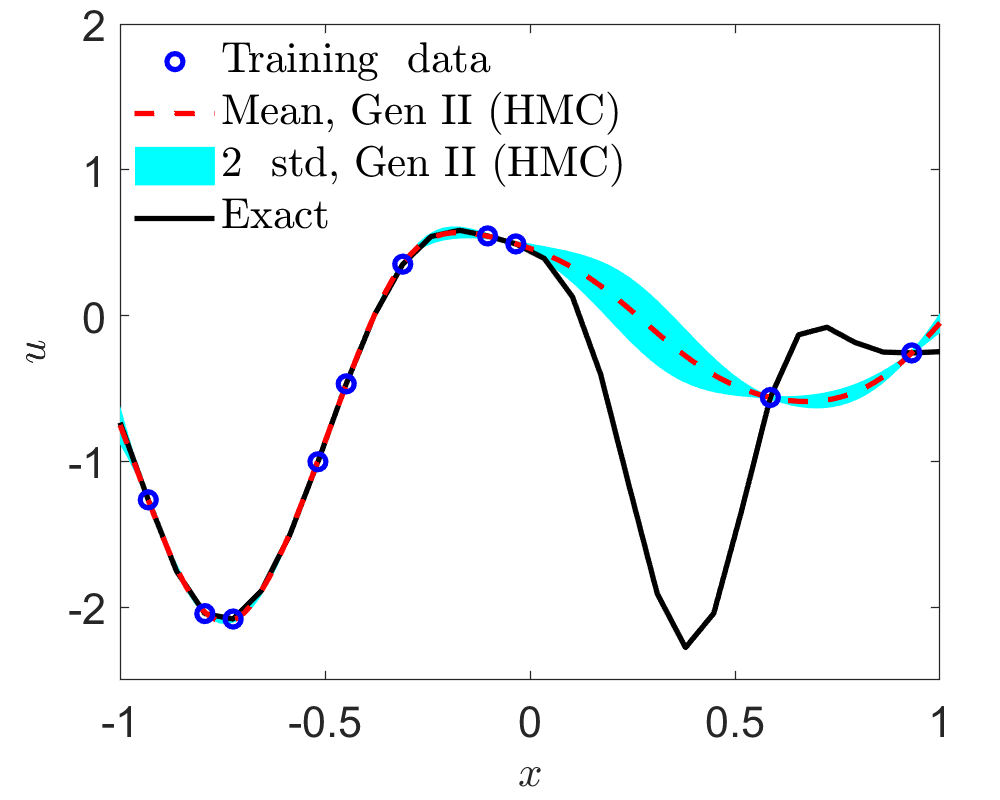}}
    \caption{
    Function approximation using GP as functional prior: Predictions for $u$ using 10 noisy measurements.
    (a) GPR. 
    (b) Gen I with NUT: The neural network architecture used in the main text.
    (c) Gen II with NUT.
    (d) Gen I with HMC.
    (e) Gen II with HMC.
    }\label{fig:gp_prior}
\end{figure}

We randomly sample 10,000 samples from \Eqref{eq:gp_prior} for the training of GANs, and 30 equidistant points are employed to resolve each sample of $u$. The inputs for the generator are 40-dimensional Gaussian noise $\bm{\xi}$ and $x$.  The details for optimization,  e.g., learning rate, number of training steps, can be found in \ref{sec:dnn_arch}.  To check the convergence of GANs, we compute the covariance matrix using 10,000 generated samples for $u$ on the completion of training. The mean square error (MSE) between the calculated and exact covariance matrix is less than $0.2\%$ for all test cases, suggesting the convergence of GANs. At the posterior estimation stage, to reduce the effect of noise on the predicted accuracy, we set the noise as a Gaussian distribution with zero mean and a relatively small variance, i.e., $\mathcal{N}(0, 0.01^2)$. We then employ the No-U-Turn \cite{hoffman2014no} in {\emph{Tensorflow Probability}} \cite{lao2020tfp}  to estimate the posterior of $\bm{\xi}$ and illustrate the predictions in Figs. B.\ref{fig:gp_priorb}-B.\ref{fig:gp_priorc}.  As shown, the results from Gen I and II are reasonable: (1) the uncertainty increases at locations where we have no measurements, and (2) the computational errors between the predicted mean and the exact solution are mostly bounded by the predicted two standard deviations. Moreover, the results from Gen I agree better with those from GPR (Fig. B.\ref{fig:gp_priora}).

We would also like to discuss the computational cost for Gen I and II. In the above test cases, we set the burnin step as 20,000 in No-U-Turn (the initial step and target acceptance rate are introduced in \ref{sec:dnn_arch}) to guarantee the convergence, the computational time for Gen I and II are around 15 minutes and 8 hours (Intel (R) Xeon (R) CPU E5-2643 @ 3.3 GHz), respectively.

We also utilize the vanilla Hamitonian Monte Carlo (HMC) in {\emph{Tensorflow Probability}} \cite{lao2020tfp} for posterior estimation. As reported in \cite{hoffman2014no}, the HMC's best performance occurs as the acceptance rate is around 0.65. We then carefully tune the burnin step, time step as well as the Leapfrog step in HMC to achieve the best acceptance rate. In particular, we present the results for Gen I and II in Figs. B.\ref{fig:gp_priord}-B.\ref{fig:gp_priore} (in both cases the acceptance rates are around 0.62), we see that Gen I can still provide similar results as those from No-U-Turn (Fig. B.\ref{fig:gp_priorb}), but the results from Gen II are quite different from those from No-U-Turn (Fig. B.\ref{fig:gp_priorc}) as well as the reference solution (Fig. B.\ref{fig:gp_priora}).  Possible reasons are discussed in \ref{sec:whyfail}.

% To ensure the convergence of posterior estimation, the burnin step is set as 20,000 in all test cases, and the acceptance rate is set as 0.6. It takes about 8 hours for the to finish the sampling process, while it only takes about 15 mins to the ResGANs. (Intel (R) Xeon (R) CPU E5-2643). Note that we use a relatively large burnin step here to ensure the convergence of  posterior estimation. For most test cases in our main text, 2,000 burnin steps are enough to obtain converged results, which generally takes 2-5 mins for posterior estimation.

\section{Why Wasserstein GANs Could Fail}~\label{sec:whyfail}

In our preliminary study, we found that vanilla HMC could fail for posterior estimation with the learned prior, especially for the second neural network architecture (Gen II) in ~\ref{sec:comparison}. In particular, the uncertainty is underestimated. This could happen even if the statistics of the learned prior, e.g., the covariance, converges to the target ones. 

We attribute this phenomenon to that we aim to train the learned prior converging to the historical data distribution in the Wasserstein sense, since we are using Wasserstein GANs, but this is not quite ideal for accurate posterior estimation. There are mainly two reasons:
\begin{itemize}
    \item  A series of generated multivariate distributions converge to the target distribution in the Wasserstein distance does not imply that the statistics of the conditional distributions converge to the statistics of the target conditional distribution. In other words, even if the learned prior converges to the distribution of historical data in the Wasserstein sense, the posterior corresponding to the learned prior may not converge to the posterior corresponding the distribution of historical data.
    \item Even if the conditional distributions converge to the target conditional distribution, the corresponding distributions in the input noise space could be multi-modal, and hard to be sampled from with vanilla HMC method.
\end{itemize}

\begin{figure}[H]
\centering
\includegraphics[width=1.0\textwidth]{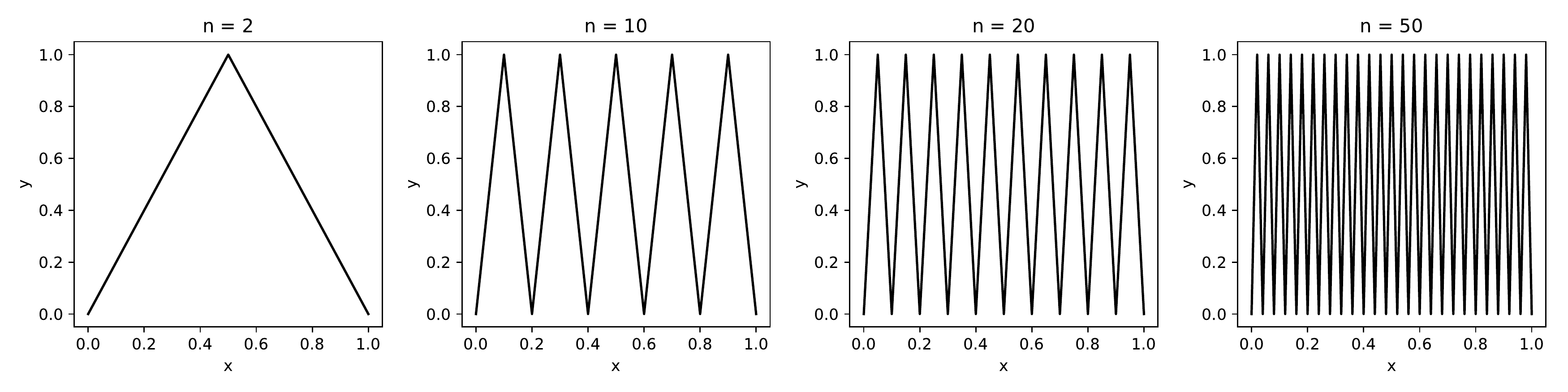}
% \begin{subfigure}{\textwidth}
%     \centering
%     \includegraphics[width=1.0\textwidth]{zigzag.pdf}
%     \caption{}
%     \end{subfigure}
\caption{Illustration of the generated distributions $Q_n$ for $n=2, 5, 20, 50$. $Q_n$ is the uniform distributions on the graph of the function.}
\label{fig:zigzag}
\end{figure}

Here we construct an example. In the main text, the prior and posterior are both in the functional space, but here we use the Euclidean space as a simple yet accurate illustration. We set the target distribution of $T$ as $\mathcal{U}([0,1]^2)$, the uniform distribution on $[0,1]^2$, and the generated distributions $Q_n$ as uniform distributions on the graph of $y = \arccos(\cos(n\pi x))/\pi, x\in [0,1]$, for $n=1,2...$. The input noise $\xi$ could be sampled from $\mathcal{U}([0,1])$ and the generator function is $(\xi, \arccos(\cos(n\pi \xi))/\pi)$. We visualize the generated distributions $Q_n$ in Fig.~\ref{fig:zigzag}.

It is not hard to see that $Q_n$ converges to $T$ in the Wasserstein distance. However, for any $n$ and for any $x\in[0,1]$, the conditional distributions $Q_n(y|x)$ are Dirac delta distributions $\delta_{\arccos(\cos(n\pi x))/\pi}$, while the conditional distribution $T(y|x)$ is $\mathcal{U}([0,1])$. On the other hand, for any $y\in (0,1)$, while we can show that $Q_n(x|y)$ converges to $T(x|y)$, i.e., $\mathcal{U}([0,1])$, the corresponding distributions for $z$ are uniform discrete distributions with $n$ modes.

Such an example also indicates why in ~\ref{sec:comparison} the first generator architecture (Gen I), which is used in the main text, outperforms the second one (Gen II): the function represented by Gen I is more regularized, that is to say, some twisted representations of the prior distribution, as in the example constructed in this section, are avoided.
% In the above example the input noise dimension is less than the target distribution dimension, but even if the former is no less than the latter, it is still possible that some of the input noise dimensions are curled up so that the generated distributions still concentrate on low dimensional manifolds.

\section{Details of Learning Hyperparameters}
\label{sec:dnn_arch}

In all cases of Sec.~\ref{sec:results}, we draw 1,000 posterior samples of $\bm{\xi}$ to compute the posterior functions of interest, i.e., $M = 1,000$. The Adam optimizer is employed for the training of both GANs and DeepONet. For the training of GANs, the initial learning rate is $10^{-4}$, $\beta_1 = 0.5$,  $\beta_2 = 0.9$. For the training of DeepONets in Sec. \ref{sec:frac} and Sec. \ref{sec:2dflow}, the initial learning rate is set as  $10^{-4}$ and $10^{-3}$, respectively, $\beta_1 = 0.9$,  $\beta_2 = 0.999$. More details for the GANs as well as DeepONet (e.g., architecture, training steps) used in each case are presented in Table \ref{table:gan_arch} and Table \ref{table:deeponet_arch}, respectively.

\begin{table}[H]
\centering
{\footnotesize
\begin{tabular}{c|cc|cc|c|c}
\hline \hline
  & \multicolumn{2}{c|}{G ($\tilde{g}/\tilde{h}^*$)}  & \multicolumn{2}{c|}{D} & \multirow{2}{*}{$\bm{\xi}_D$} & \multirow{2}{*}{Training steps}  \\
  & width $\times$ depth  & Activation & width $\times$ depth & Activation &\\
  \hline
  
  {Sec. 3.1} & $64 \times 2 / 64 \times 2$   & tanh/tanh & $128 \times 3$ & Leaky ReLu &   10 &500,000 \\
  \hline
  {Sec. 3.2} & $64 \times 2 / 64 \times 2$   & tanh/tanh & $128 \times 3$ & Leaky ReLu &  40 (Forward), 60 (Inverse) & 500,000 \\
  \hline
  {Sec. 3.3} & $64 \times 2 / 64 \times 2$   & tanh/tanh & $128 \times 3$ & Leaky ReLu &   40 &500,000 \\
  \hline
  {Sec. 3.4} & $128 \times 2 / 128 \times 2$   & tanh/tanh & $512 \times 2$ & Leaky ReLu &  100 &500,000 \\
   \hline
  {Sec. 3.5} & $64 \times 3 / 64 \times 3$   & sin/tanh & $128 \times 3$ & Leaky ReLu & 20 & 500,000 \\
  \hline
  {Appendix B} & $64 \times 2 / 64 \times 2$   & tanh/tanh & $128 \times 3$ & Leaky ReLu & 40 & 500,000 \\
  \hline \hline
\end{tabular}
}
\caption{
Architecture and training steps of GANs in each case. The width and depth are for the hidden layers. $d_G$ is the dimension of $\bm{\xi}$ as well as the output dimension of $\tilde{g}$ and $\tilde{h}^*$ in $G$.
}
\label{table:gan_arch}
\end{table}

\begin{table}[H]
\centering
{\footnotesize
\begin{tabular}{c|cc|cc|c}
\hline \hline
  & \multicolumn{2}{c|}{Branch Net}  & \multicolumn{2}{c|}{Trunk Net} & \multirow{2}{*}{Training steps}  \\
  & width $\times$ depth  & Activation & width $\times$ depth & Activation &\\
  \hline
  
    {Sec. 3.3} & $64 \times 2 $   & tanh & $64 \times 3$ & tanh &  100,000 \\
  \hline 
  {Sec. 3.4} & $256 \times 2 $   & tanh  & $128 \times 2$ & tanh &  200,000 \\
  \hline \hline

\end{tabular}
}
\caption{
Architecture and training steps of DeepONets in each case. 
}
\label{table:deeponet_arch}
\end{table}

As for the posterior estimation, we employ the No-U-Turn \cite{hoffman2014no}, which can adaptively set path lengths in the Hamiltonian Monte Carlo method in this study. In all cases, the initial step size is set as 1, the target acceptance rate is 0.6, and the number of burnin steps is 2,000.  Note that we use a relatively large burnin step here to ensure the convergence of  posterior estimation in \ref{sec:comparison}. For most test cases in our main text, 2,000 burnin steps are enough to obtain converged results, which generally takes 2-5 mins for posterior estimation (Intel (R) Xeon (R) CPU E5-2643 @ 3.3 GHz).

% \section{Antiderivative operator}
% To verify the present method, we test a case which is similar as the validation case in \cite{lu2019deeponet}, i.e., learning the antiderivative operator using DeepONet. In particular, the following equation is considered:
% \begin{align}
%     \frac{du}{dx} &= f, ~ x \in [-1, 1],\\
%     u(-1) &= 0,
% \end{align}
% The objective is to predict $u$ for $x \in [-1, 1]$ for any $f$. 

% To pretrain the DeepONets, where $f$ is a Gaussian process reads as
% \begin{align}\label{eq:f_onet}
%     f & \thicksim \mathcal{GP}(0, \mathcal{K}), ~ \mathcal{K} = \exp \left( - \frac{(x - x')^2}{2 l^2}\right),\\
%     x,&~ x' \in [-1, 1], ~l = 0.2. 
% \end{align}

% \begin{figure}[H]
%     \centering
%     \subfigure[]{\label{fig:fu_oneta}
%     \includegraphics[width = 0.3\textwidth]{./Figures/f_uq_onet.png}
%     \includegraphics[width = 0.3\textwidth]{./Figures/u_uq_onet.png}}
%     \subfigure[]{\label{fig:fu_onetb}
%     \includegraphics[width = 0.3\textwidth]{./Figures/f_fu_uq.png}
%     \includegraphics[width = 0.3\textwidth]{./Figures/u_fu_uq.png}}
%     \caption{
%     Learning antiderivative using DeepONets with uncertainties. 
%     (a) Predicted  $f$ and $u$ based on partial observations on $f$. 
%     (b) Predicted  $f$ and $u$ based on partial observations on $f$ and $u$.
%     Mean: predicted mean; 2 std: predicted two standard deviations. 
%     }
%     \label{fig:fu_onet}
% \end{figure}

\bibliographystyle{elsarticle-num}
\bibliography{refs}

\begin{thebibliography}{10}
\expandafter\ifx\csname url\endcsname\relax
  \def\url#1{\texttt{#1}}\fi
\expandafter\ifx\csname urlprefix\endcsname\relax\def\urlprefix{URL }\fi
\expandafter\ifx\csname href\endcsname\relax
  \def\href#1#2{#2} \def\path#1{#1}\fi

\bibitem{lecun2015deep}
Y.~LeCun, Y.~Bengio, G.~Hinton, Deep learning, Nature 521~(7553) (2015)
  436--444.

\bibitem{goodfellow2016deep}
I.~Goodfellow, Y.~Bengio, A.~Courville, Y.~Bengio, Deep learning, Vol.~1, MIT
  press Cambridge, 2016.

\bibitem{kendall2017uncertainties}
A.~Kendall, Y.~Gal, What uncertainties do we need in {B}ayesian deep learning
  for computer vision?, arXiv preprint arXiv:1703.04977 (2017).

\bibitem{rasmussen2003gaussian}
C.~E. Rasmussen, Gaussian processes in {M}achine {L}earning, in: Summer school
  on machine learning, Springer, 2003, pp. 63--71.

\bibitem{neal2012bayesian}
R.~M. Neal, Bayesian learning for neural networks, Vol. 118, Springer Science
  \& Business Media, 2012.

\bibitem{raissi2017machine}
M.~Raissi, P.~Perdikaris, G.~E. Karniadakis, Machine learning of linear
  differential equations using {G}aussian processes, Journal of Computational
  Physics 348 (2017) 683--693.

\bibitem{yang2021b}
L.~Yang, X.~Meng, G.~E. Karniadakis, {B-PINNs}: Bayesian physics-informed
  neural networks for forward and inverse {PDE} problems with noisy data,
  Journal of Computational Physics 425 (2021) 109913.

\bibitem{raissi2019physics}
M.~Raissi, P.~Perdikaris, G.~E. Karniadakis, Physics-informed neural networks:
  A deep learning framework for solving forward and inverse problems involving
  nonlinear partial differential equations, Journal of Computational Physics
  378 (2019) 686--707.

\bibitem{meng2020composite}
X.~Meng, G.~E. Karniadakis, A composite neural network that learns from
  multi-fidelity data: {A}pplication to function approximation and inverse
  {PDE} problems, Journal of Computational Physics 401 (2020) 109020.

\bibitem{meng2021multi}
X.~Meng, H.~Babaee, G.~E. Karniadakis, Multi-fidelity {B}ayesian neural
  networks: Algorithms and {A}pplications, Journal of Computational Physics
  (2021) 110361.

\bibitem{blundell2015weight}
C.~Blundell, J.~Cornebise, K.~Kavukcuoglu, D.~Wierstra, Weight uncertainty in
  neural network, in: International Conference on Machine Learning, PMLR, 2015,
  pp. 1613--1622.

\bibitem{chen2014stochastic}
T.~Chen, E.~Fox, C.~Guestrin, Stochastic gradient {H}amiltonian {M}onte
  {C}arlo, in: International conference on machine learning, PMLR, 2014, pp.
  1683--1691.

\bibitem{lee2017deep}
J.~Lee, Y.~Bahri, R.~Novak, S.~S. Schoenholz, J.~Pennington, J.~Sohl-Dickstein,
  Deep neural networks as {G}aussian processes, arXiv preprint arXiv:1711.00165
  (2017).

\bibitem{flam2017mapping}
D.~Flam-Shepherd, J.~Requeima, D.~Duvenaud, Mapping {G}aussian process priors
  to {B}ayesian neural networks, in: NIPS Bayesian deep learning workshop,
  2017.

\bibitem{tran2020all}
B.-H. Tran, S.~Rossi, D.~Milios, M.~Filippone, All you need is a good
  functional prior for {B}ayesian deep learning, arXiv preprint
  arXiv:2011.12829 (2020).

\bibitem{yang2020physics}
L.~Yang, D.~Zhang, G.~E. Karniadakis, Physics-informed generative adversarial
  networks for stochastic differential equations, SIAM Journal on Scientific
  Computing 42~(1) (2020) A292--A317.

\bibitem{callaham2019robust}
J.~L. Callaham, K.~Maeda, S.~L. Brunton, Robust flow reconstruction from
  limited measurements via sparse representation, Physical Review Fluids 4~(10)
  (2019) 103907.

\bibitem{finn2017model}
C.~Finn, P.~Abbeel, S.~Levine, Model-agnostic meta-learning for fast adaptation
  of deep networks, in: International Conference on Machine Learning, PMLR,
  2017, pp. 1126--1135.

\bibitem{hoffman2014no}
M.~D. Hoffman, A.~Gelman, The {No-U-Turn} sampler: adaptively setting path
  lengths in {Hamiltonian Monte Carlo}, Journal of Machine Learning Research
  15~(1) (2014) 1593--1623.

\bibitem{lu2021learning}
L.~Lu, P.~Jin, G.~Pang, Z.~Zhang, G.~E. Karniadakis, Learning nonlinear
  operators via {DeepONet} based on the universal approximation theorem of
  operators, Nature Machine Intelligence 3~(3) (2021) 218--229.

\bibitem{patel2020gan}
D.~V. Patel, A.~A. Oberai, {GAN}-based priors for quantifying uncertainty,
  arXiv preprint arXiv:2003.12597 (2020).

\bibitem{patelbayesian}
D.~V. Patel, D.~Ray, H.~Ramaswamy, A.~A. Oberai, Bayesian inference in
  physics-driven problems with adversarial priors.

\bibitem{goodfellow2014generative}
I.~J. Goodfellow, J.~Pouget-Abadie, M.~Mirza, B.~Xu, D.~Warde-Farley, S.~Ozair,
  A.~Courville, Y.~Bengio, Generative adversarial networks, arXiv preprint
  arXiv:1406.2661 (2014).

\bibitem{gulrajani2017improved}
I.~Gulrajani, F.~Ahmed, M.~Arjovsky, V.~Dumoulin, A.~Courville, Improved
  training of {W}asserstein {GAN}s, arXiv preprint arXiv:1704.00028 (2017).

\bibitem{cai2021deepm}
S.~Cai, Z.~Wang, L.~Lu, T.~A. Zaki, G.~E. Karniadakis, Deepm\&mnet: Inferring
  the electroconvection multiphysics fields based on operator approximation by
  neural networks, Journal of Computational Physics 436 (2021) 110296.

\bibitem{MaoKar2018SINUM}
Z.~Mao, G.~E. Karniadakis, A spectral method (of exponential convergence) for
  singular solutions of the diffusion equation with general two-sided
  fractional derivative, SIAM Journal on Numerical Analysis 56~(1) (2018)
  24--49.

\bibitem{zheng2020physics}
Q.~Zheng, L.~Zeng, G.~E. Karniadakis, Physics-informed semantic inpainting:
  Application to geostatistical modeling, Journal of Computational Physics 419
  (2020) 109676.

\bibitem{meerschaert2013hydraulic}
M.~M. Meerschaert, M.~Dogan, R.~L. Van~Dam, D.~W. Hyndman, D.~A. Benson,
  Hydraulic conductivity fields: {G}aussian or not?, Water Resources Research
  49~(8) (2013) 4730--4737.

\bibitem{kang2019coupled}
X.~Kang, X.~Shi, A.~Revil, Z.~Cao, L.~Li, T.~Lan, J.~Wu, Coupled
  hydrogeophysical inversion to identify non-{G}aussian hydraulic conductivity
  field by jointly assimilating geochemical and time-lapse geophysical data,
  Journal of Hydrology 578 (2019) 124092.

\bibitem{NDP_SCR}
{VIV Model Test of a Catenary Riser}, Tech. Rep. 512345.00.01, Norwegian Marine
  Technology Research Institute (2001).

\bibitem{hospedales2020meta}
T.~Hospedales, A.~Antoniou, P.~Micaelli, A.~Storkey, Meta-learning in neural
  networks: A survey, arXiv preprint arXiv:2004.05439 (2020).

\bibitem{lao2020tfp}
J.~Lao, C.~Suter, I.~Langmore, C.~Chimisov, A.~Saxena, P.~Sountsov, D.~Moore,
  R.~A. Saurous, M.~D. Hoffman, J.~V. Dillon, tfp. mcmc: Modern {Markov Chain
  Monte Carlo} tools built for modern hardware, arXiv preprint arXiv:2002.01184
  (2020).

\end{thebibliography}
% \bibliography{refs.bib}

\end{document}